\pdfoutput=1
\documentclass[11pt]{article}

\usepackage{arxiv}

\usepackage{times}
\usepackage{longtable}
\usepackage{algorithm}
\usepackage{algorithmic}
\usepackage{nicefrac}
\usepackage{booktabs}
\usepackage{footmisc}
\usepackage[shortlabels]{enumitem}
\usepackage{multirow}

\usepackage{array}
\usepackage{url}
\usepackage[strict]{changepage}
\usepackage{makecell}
\usepackage{titlesec}
\usepackage{setspace}
\usepackage{bm}
\usepackage{mdframed}
\usepackage{bbm}
\usepackage{threeparttable}
\usepackage{subfigure}

\usepackage[T1]{fontenc}

\usepackage{titletoc}
\usepackage{longtable}
\usepackage{pifont} 
\usepackage{wrapfig}

\usepackage[table]{xcolor}
\usepackage{makecell}

\usepackage{fancyhdr} 

\usepackage[most]{tcolorbox} 

\definecolor{darkpastelgreen}{rgb}{0.01, 0.75, 0.24}

\titleformat*{\subparagraph}{\itshape}

\newsavebox\actorsfigure

\usepackage{circledsteps}
\usepackage{fontawesome5}

\definecolor{LightGray}{RGB}{250,250,250}
\definecolor{Gray}{RGB}{240, 240, 240}

\NewDocumentCommand{\heng}
{ mO{} }{\textcolor{red}{\textsuperscript{\textit{Heng}}\textsf{\textbf{\small[#1]}}}}
\usepackage[numbers]{natbib}
\usepackage{amsfonts}
\usepackage[hidelinks]{hyperref}

\title{A Survey on Post-training of Large Language Models}

\author{
{\bfseries Guiyao Tie$^{1}$\footnotemark[2]}\quad
{\bfseries Zeli Zhao$^{1}$}\quad
{\bfseries Dingjie Song$^{2}$}\quad
{\bfseries Fuyang Wei$^{3}$}\quad
{\bfseries Rong Zhou$^{2}$}\quad
{\bfseries Yurou Dai$^{2}$}\quad\\
{\bfseries Wen Yin$^{1}$}\quad
{\bfseries Zhejian Yang$^{4}$}\quad
{\bfseries Jiangyue Yan$^{5}$}\quad
{\bfseries Yao Su$^{6}$}\quad
{\bfseries Zhenhan Dai$^{1}$}\quad
{\bfseries Yifeng Xie$^{1}$}\quad\\
{\bfseries Yihan Cao$^{7}$}\quad
{\bfseries Lichao Sun$^{2}$}\quad
{\bfseries Pan Zhou$^{1}$}\quad
{\bfseries Lifang He$^{2}$}\quad
{\bfseries Hechang Chen$^{4}$}\quad
{\bfseries Yu Zhang$^{5}$}\quad\\
{\bfseries Qingsong Wen$^{8}$}\quad
{\bfseries Tianming Liu$^{9}$}\quad
{\bfseries Neil Zhenqiang Gong$^{10}$}\quad
{\bfseries Jiliang Tang$^{11}$}\quad \\
{\bfseries Caiming Xiong$^{12}$}\quad 
{\bfseries Heng Ji$^{13}$}\quad
{\bfseries Philip S. Yu$^{14}$}\quad
{\bfseries Jianfeng Gao$^{15}$}\quad
\\
\\
{\bfseries $^{1}$Huazhong University of Science and Technology}\quad
{\bfseries $^{2}$Lehigh University}\quad\\
{\bfseries $^{3}$The University of Hong Kong}\quad
{\bfseries $^{4}$Jilin University}\quad
{\bfseries $^{5}$Southern University of Science and Technology}\quad\\
{\bfseries $^{6}$Worcester Polytechnic Institute}\quad
{\bfseries $^{7}$LinkedIn Corporation}\quad\\
{\bfseries $^{8}$Squirrel Ai Learning}\quad
{\bfseries $^{9}$University of Georgia}\quad
{\bfseries $^{10}$Duke University}\quad\\
{\bfseries $^{11}$Michigan State University}\quad
{\bfseries $^{12}$Salesforce Research}\quad
{\bfseries $^{13}$University of Illinois Urbana-Champaign}\quad \\
{\bfseries $^{14}$University of Illinois at Chicago}\quad
{\bfseries $^{15}$Microsoft Research}\quad
}

\begin{document}
\maketitle
\renewcommand{\thefootnote}{\fnsymbol{footnote}}

\footnotetext[2]{Guiyao Tie is the current corresponding author: \href{mailto:tgy@hust.edu.cn}{tgy@hust.edu.cn}}
\footnotetext[3]{Latest Update: Mar., 2025.}


\begin{abstract}
The emergence of Large Language Models (LLMs) has fundamentally transformed natural language processing, making them indispensable across domains ranging from conversational systems to scientific exploration. However, their pre-trained architectures often reveal limitations in specialized contexts, including restricted reasoning capacities, ethical uncertainties, and suboptimal domain-specific performance. These challenges necessitate advanced post-training language models (PoLMs) to address these shortcomings, such as OpenAI-o1/o3 and DeepSeek-R1 (collectively known as Large Reasoning Models, or LRMs). This paper presents the first comprehensive survey of PoLMs, systematically tracing their evolution across five core paradigms: \textbf{Fine-tuning}, which enhances task-specific accuracy; \textbf{Alignment}, which ensures ethical coherence and alignment with human preferences; \textbf{Reasoning}, which advances multi-step inference despite challenges in reward design; \textbf{Efficiency}, which optimizes resource utilization amidst increasing complexity; and \textbf{Integration and Adaptation}, which extend capabilities across diverse modalities while addressing coherence issues. Charting progress from ChatGPT's foundational alignment strategies to DeepSeek-R1's innovative reasoning advancements, we illustrate how PoLMs leverage datasets to mitigate biases, deepen reasoning capabilities, and enhance domain adaptability. Our contributions include a pioneering synthesis of PoLM evolution, a structured taxonomy categorizing techniques and datasets, and a strategic agenda emphasizing the role of LRMs in improving reasoning proficiency and domain flexibility. As the first survey of its scope, this work consolidates recent PoLM advancements and establishes a rigorous intellectual framework for future research, fostering the development of LLMs that excel in precision, ethical robustness, and versatility across scientific and societal applications. Project Github: \textcolor{red}{ \underline{\href{https://github.com/Mr-Tieguigui/LLM-Post-Training}{https://github.com/Mr-Tieguigui/LLM-Post-Training}}.}
\end{abstract}





\newpage
\tableofcontents
\newpage
\section{Introduction}\label{Section 1}

\definecolor{quoteLine}{RGB}{0,0,0} 
\definecolor{quoteText}{RGB}{0,0,0} 
\definecolor{authorText}{RGB}{0,0,0} 
\definecolor{hidden-draw}{RGB}{106,142,189} 
\definecolor{hidden-blue}{RGB}{194,230,247} 
\definecolor{hidden-orange}{RGB}{217, 230, 252}

\begin{center}
\begin{tcolorbox}[
    colframe=white, 
    colback=white, 
    coltext=quoteText, 
    boxsep=5pt, 
    arc=0pt, 
    boxrule=0pt, 
]

\textit{It is generally agreed upon that authentic intelligence equips us with reasoning capabilities, enables us to test hypotheses, and prepares for future eventualities.}
\vspace{6pt}

{\color{quoteLine}\hrule} 
\vspace{3pt}

\begin{flushright}
\textbf{Jean Khalfa} -- \textsc{What is intelligence? (1994)}
\end{flushright}

\end{tcolorbox}
\end{center}

Language models (LMs)~\citep{radford2018improving, devlin2018bert} represent sophisticated computational frameworks designed to model and generate human language. These models have revolutionized the field of natural language processing (NLP)~\citep{Collobert2011NaturalLP} by enabling machines to understand, generate, and interact with human language in a manner that closely mimics human cognition. Unlike humans, who acquire language skills naturally through interaction and exposure to contextual environments, machines must undergo extensive, data-driven training to develop similar capabilities~\citep{GHOSH2025101851}. This presents a significant research challenge, as enabling machines to comprehend and generate human language, while engaging in natural, contextually appropriate dialogue, requires not only vast computational resources but also refined methodologies for model development~\citep{chang2024survey, zhao2023survey}.

The emergence of Large Language Models (LLMs) such as GPT-3~\citep{brown2020language}, InstructGPT~\citep{zhang2023instructfingpt}, and GPT-4~\citep{achiam2023gpt} has marked a transformative phase in the evolution of LMs. These models, distinguished by their extensive parameterization and advanced learning capabilities, are designed to capture complex linguistic structures, contextual relationships, and nuanced patterns within vast datasets. This enables LLMs not only to predict subsequent words but also to generate coherent, contextually relevant text across a wide range of tasks, including translation, question answering, and summarization. The development of LLMs has sparked significant academic interest~\citep{chang2024survey, zhao2023survey, zhou2023comprehensive}, which can be divided into two main stages: \textit{pre-training} and \textit{post-training}.

\noindent\textbf{Pre-training.}~~The concept of pre-training originates from transfer learning in computer vision (CV) tasks~\citep{zhou2023comprehensive}. Its primary goal is to develop a general model using extensive datasets, which facilitates easy fine-tuning for various downstream applications. A significant advantage of pre-training is its ability to utilize any unlabeled text corpus, thereby providing an abundant source of training data. However, early static pre-training methods, such as Neural Network Language Models (NNLM)~\citep{bengio2000neural} and Word2vec~\citep{mikolov2013efficient}, struggled to accommodate different textual semantic environments, prompting the development of dynamic pre-training techniques like BERT~\citep{devlin2018bert} and XLNet~\citep{yang2019xlnet}. BERT effectively addressed the limitations of static methods by leveraging the transformer architecture and employing self-attention mechanisms on large-scale unlabeled datasets. This study established the “pre-training and fine-tuning” learning paradigm, inspiring numerous subsequent studies that introduced diverse architectures, including GPT-2~\citep{radford2019language} and BART~\citep{lewis2019bart}.

\noindent\textbf{Post-training.}~~Post-training refers to the techniques and methodologies employed after a model has undergone pre-training, aiming to refine and adapt the model for specific tasks or user requirements. Following the release of GPT-3~\citep{brown2020language}, with its 175 billion parameters, the field of post-training experienced a significant surge in interest and innovation. Various approaches emerged to enhance model performance, including fine-tuning~\citep{ziegler2020finetuninglanguagemodelshuman, wei2021finetuned}, which adjusts model parameters using labeled datasets or specific task data; alignment strategies~\citep{peng2023instruction, su2023pandagpt, zeng2023evaluating}, which optimize models to better align with user preferences; knowledge adaptation techniques~\citep{dong2022survey, rubin2021learning}, which enable models to incorporate domain-specific knowledge; and reasoning improvements~\citep{yao2024tree, besta2024graph}, which enhance a model's ability to make logical inferences and decisions. 
Collectively known as Post-training Language Models (PoLMs), these techniques have led to the development of models such as GPT-4~\citep{achiam2023gpt}, LLaMA-3~\citep{dubey2024llama}, Gemini-2.0~\citep{gemini2.0}, and Claude-3.5~\citep{TheC3}, marking substantial progress in LLM capabilities. However, post-trained models often struggle to adapt to new tasks without retraining or significant parameter adjustments, making PTM development an area of active research.

\begin{figure}[ht]
\centering
\includegraphics[width=1\linewidth]{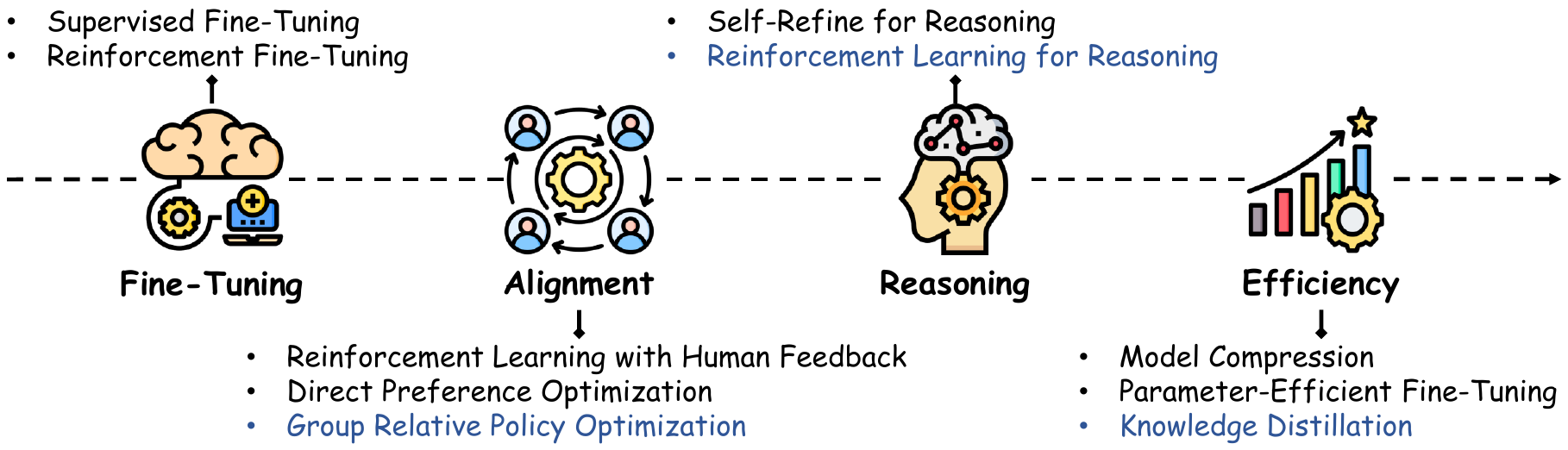}
\caption{The evolution of post-training techniques for Large Language Models, delineating the progression from initial methodologies to advanced approaches, with emphasis on DeepSeek model contributions (highlighted in blue).}
\label{fig:technique evolution}
\end{figure}

As highlighted, pre-trained language models (PLMs) primarily aim to provide general knowledge and capabilities, while PoLMs focus on adapting these models to specific tasks and requirements. A notable example of this adaptation is the latest LLM, DeepSeek-R1~\citep{DeepSeekAI2025DeepSeekR1IR}, which illustrates the evolution of PoLMs in enhancing reasoning abilities, aligning with user preferences, and improving adaptability across various domains~\citep{he2024ptqd}. Furthermore, the increasing availability of open-sourced LLMs (e.g., LLaMA~\citep{li2023llama}, Gemma~\citep{team2024gemma}, and Nemotron~\citep{adler2024nemotron}) and domain-specific large datasets (e.g., PromptSource~\citep{bach2022promptsource} and Flan~\citep{longpre2023flan}) is driving a trend among academic researchers and industry practitioners to develop PoLMs. This trend underscores the growing recognition of the importance of tailored adaptations in the field of PoLMs.

In the existing literature, PLMs have been widely discussed and surveyed~\citep{zhou2023comprehensive,liu2023pre,han2021pre,qiu2020pre}, while PoLMs are seldom reviewed systematically. To advance these techniques, it is essential to thoroughly examine the existing body of research to identify key challenges, gaps, and opportunities for further refinement. This survey aims to fill this gap by providing a structured framework for the evolving research in post-training. As shown in \textbf{Fig. \ref{fig:technique evolution}}, it explores multiple stages of post-training, with a particular focus on those employed \textbf{from ChatGPT to DeepSeek}. These techniques encompass a wide range of methodologies, including fine-tuning, LLM alignment, reasoning enhancement, and efficiency improvements. The blue section of the figure specifically highlights the set of post-training methods applied by DeepSeek, emphasizing the innovative strategies that have contributed to its success in adapting to user preferences and domain-specific needs.

\subsection{Major Contributions}
This paper represents the first comprehensive survey on PoLMs, providing a thorough, structured exploration of the latest advancements in the field. While previous surveys have typically focused on specific aspects of LLM development, such as preference alignment~\citep{wang2024comprehensive}, parameter-efficient fine-tuning~\citep{Han2024ParameterEfficientFF}, and foundational techniques of LLMs~\citep{Xiao2025FoundationsOL}, they have largely concentrated on narrow subtopics. In contrast, this survey takes a holistic approach, providing a complete review of the core techniques commonly employed during post-training and systematically categorizing them. Additionally, we investigate the datasets and real-world applications integral to these methods, as illustrated in \textbf{Fig. \ref{fig:oganization}}, and identify open challenges and promising directions for future research. The main contributions of this survey are as follows:

\begin{figure}[ht]
\centering
\includegraphics[width=1\linewidth]{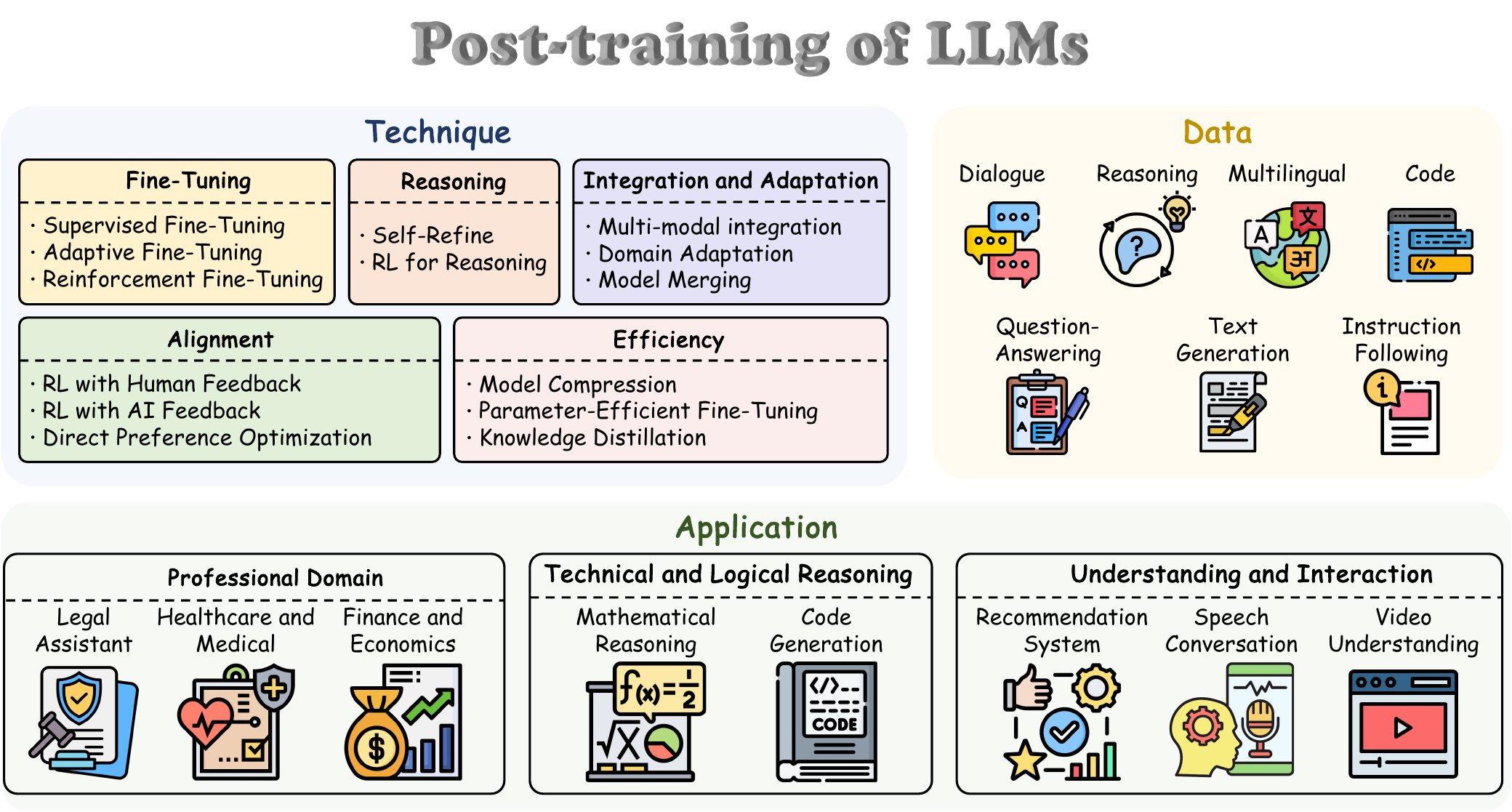}
\caption{Structural overview of post-training techniques surveyed in this study, illustrating the organization of methodologies, datasets, and applications.}
\label{fig:oganization}
\end{figure}

\begin{itemize}
    \item \textbf{Comprehensive Historical Synthesis.} We provide the first in-depth synthesis of PoLMs, tracing their evolution from ChatGPT’s initial reinforcement learning from human feedback (RLHF) to DeepSeek-R1’s innovative cold-start RL approach. This synthesis covers key techniques (i.e., Fine-tuning, Alignment, Reasoning, Efficiency, and Integration and Adaptation), analyzing their development and associated challenges, such as computational complexity and ethical considerations. By presenting this progression as a cohesive narrative, enriched with essential references, we provide researchers with a comprehensive overview of the evolution of post-training in recent years, serving as a foundational resource for the field.

    \item \textbf{Structured Taxonomy and Framework.} We introduce a structured taxonomy, depicted in \textbf{Fig. \ref{fig:oganization}}, classifying post-training methods into five distinct categories and organizing datasets into seven types while framing applications across professional, technical, and interactive domains. This framework clarifies the interrelationships and practical implications of these methods, offering a systematic perspective on their development. By providing well-defined categories and analytical insights, we improve accessibility and comprehension for both novices and experts, establishing a comprehensive guide for navigating the complexities of post-training research.

    \item \textbf{Future Directions.} We highlight emerging trends, particularly the rise of Large Reasoning Models (LRMs) such as o1~\citep{jaech2024openai} and DeepSeek-R1~\citep{DeepSeekAI2025DeepSeekR1IR}, which harness large-scale reinforcement learning to push the boundaries of reasoning. We emphasize that ongoin advancements are crucial for further enhancing reasoning capabilities and domain adaptability. Our analysis identifies key challenges, including scalability constraints, ethical alignment risks, and multimodal integration obstacles. We propose research avenues such as adaptive RL frameworks and fairness-aware optimization. These directions aim to propel post-training forward, ensuring LLMs achieve heightened precision and trustworthiness to meet future demands.
\end{itemize}

\subsection{Organization}

This survey is systematically organized to comprehensively explore Post-training Language Models (PoLMs), spanning their historical evolution, methodologies, datasets, applications, and future trajectories. Section \ref{Section 2} provides a historical overview of PoLMs. Section \ref{Section 3} examines Fine-tuning, including Supervised Fine-Tuning (SFT) in Section \ref{Section 3.1} and Reinforcement Fine-Tuning (RFT) in Section \ref{Section 3.3}. Section \ref{Section 4} addresses Alignment, covering Reinforcement Learning from Human Feedback (RLHF) in Section \ref{Section 4.1}, Reinforcement Learning from AI Feedback (RLAIF) in Section \ref{Section 4.2}, and Direct Preference Optimization (DPO) in Section \ref{Section 4.3}. Section \ref{Section 5} focuses on Reasoning, with Self-Refinement Methods in Section \ref{Section 5.1} and Reinforcement Learning for Reasoning in Section \ref{Section 5.2}. Section \ref{Section 6} surveys Efficiency-enhancing methods, including Model Compression in Section \ref{Section 6.1}, Parameter-Efficient Fine-Tuning (PEFT) in Section \ref{Section 6.2}, and Knowledge Distillation in Section \ref{Section 6.3}. Section \ref{Section 7} investigates Integration and Adaptation, addressing multi-modal approaches, domain adaptation, and model merging. Section \ref{Section 8} reviews datasets used in post-training. Section \ref{Section 9} explores LLM applications. Section \ref{Section 10} evaluates open problems and future directions. Finally, Section \ref{Section 11} concludes with a summary and research outlook.

\section{Overview}\label{Section 2}

\subsection{History of PoLMs}
\begin{figure}[h]
\centering
\includegraphics[width=1\linewidth]{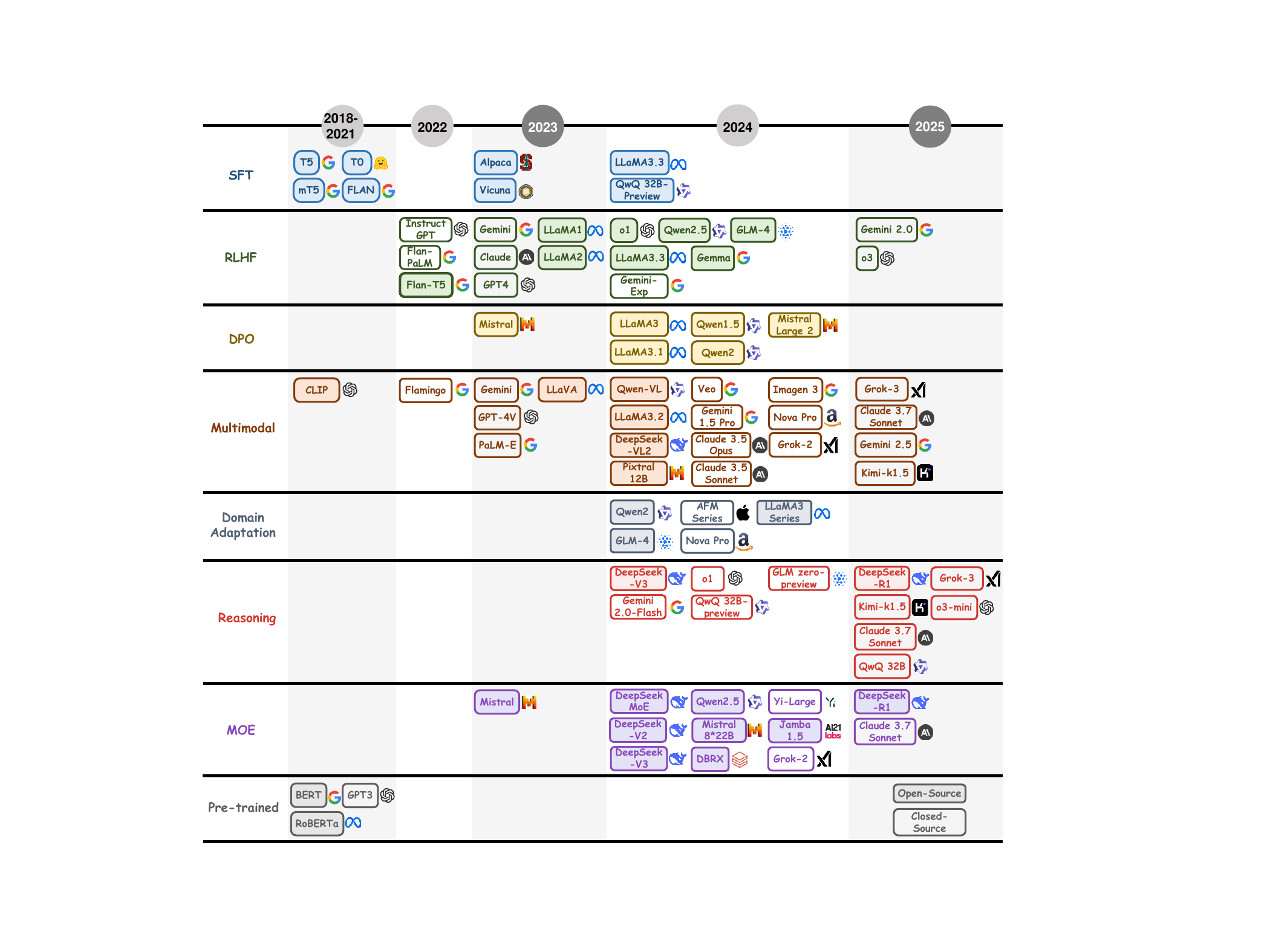}
\caption{Timeline of post-training technique development for Large Language Models (2018–2025), delineating key milestones in their historical progression.}
\label{fig:history}
\end{figure}

The advancement of LLMs constitutes a pivotal chapter in natural language processing (NLP), with post-training methods serving as critical catalysts in their evolution from generalized pre-trained architectures to specialized, task-adaptive systems. This section delineates the historical trajectory of Post-training Language Models (PoLMs), tracing their development from foundational pre-training milestones exemplified by BERT~\citep{devlin2018bert} and GPT~\citep{radford2018improving} to the sophisticated post-training paradigms embodied in contemporary models such as o1~\citep{jaech2024openai} and DeepSeek-R1~\citep{DeepSeekAI2025DeepSeekR1IR}. Illustrated in \textbf{Fig. \ref{fig:history}}, this progression reflects a shift from establishing broad linguistic competence to enhancing task-specific adaptation, ethical alignment, reasoning sophistication, and multi-modal integration, marking a transformative journey in LLM capabilities.

The inception of modern PoLMs history aligns with the pre-training revolution in \textbf{2018}, heralded by the releases of BERT~\citep{devlin2018bert} and GPT~\citep{radford2018improving}, which redefined NLP benchmarks. BERT’s bidirectional autoencoding framework, leveraging transformer architecture and self-attention, excelled in capturing contextual interdependencies for tasks like question answering, while GPT’s autoregressive design prioritized generative coherence, setting a precedent for text generation. These models established the "pre-training and fine-tuning" paradigm, with subsequent refinements in \textbf{2019} via T5~\citep{Raffel2019ExploringTL}, which unified diverse tasks under a text-to-text framework, fostering multi-task learning and laying a robust foundation for post-training advancements.

The landscape of PoLMs began to evolve substantially from \textbf{2020} onward, driven by a growing need to adapt pre-trained models efficiently to diverse tasks with limited data. Early innovations like prefix-tuning~\citep{Li2021PrefixTuningOC} and prompt-tuning~\citep{lester2021power} introduced lightweight adaptation strategies, enabling multi-task flexibility by modifying model inputs rather than retraining entire architectures, thus conserving computational resources while broadening applicability. This period also saw a pivotal shift toward user-centric optimization with the advent of Reinforcement Learning from Human Feedback (RLHF) in \textbf{2021}~\citep{Ouyang2022TrainingLM}, a technique that leveraged human evaluations to align model outputs with subjective preferences, enhancing practical utility in conversational settings. By \textbf{2022}, RLHF matured with the adoption of Proximal Policy Optimization (PPO)~\citep{Schulman2017ProximalPO}, refining alignment stability and mitigating overfitting to noisy feedback. The release of ChatGPT in late 2022~\citep{achiam2023gpt} crystallized these advancements, showcasing RLHF’s transformative potential in creating responsive, user-aligned LLMs and catalyzing a surge in PoLMs research. Concurrently, Chain-of-Thought (CoT) prompting~\citep{Wei2022ChainOT} emerged as a reasoning enhancement strategy, encouraging models to articulate intermediate steps in complex tasks, thereby improving transparency and accuracy, particularly in logical inference and problem-solving domains.

Between \textbf{2022} and \textbf{2024}, PoLMs diversified to address domain specificity, ethical robustness, and multi-modal integration, reflecting an increasingly nuanced approach to LLM refinement. Domain adaptation techniques, such as Retrieval-Augmented Generation (RAG)~\citep{lewis2020retrieval}, emerged to integrate external knowledge bases, enabling contextually enriched outputs for specialized fields without necessitating full retraining—a critical advancement for professional applications requiring up-to-date information. Ethical alignment efforts intensified, with Direct Preference Optimization (DPO)~\citep{rafailov2024directpreferenceoptimizationlanguage} in \textbf{2023} streamlining RLHF by directly optimizing model outputs against human preferences, bypassing intermediate reward modeling to enhance efficiency and robustness. Simultaneously, the pursuit of multi-modal capabilities gained traction, with models like PaLM-E~\citep{driess2023palm} and Flamingo~\citep{alayrac2022flamingo} pioneering vision-language integration, followed by BLIP-2~\citep{li2023blip} and LLaVA~\citep{Liu2023VisualIT}, which extended these efforts into broader domains like medical imaging. Efficiency innovations paralleled these developments, notably through Mixture of Experts (MoE) architectures; Google’s Switch-C Transformer~\citep{fedus2022switch} in 2022 introduced sparse activation of 1.6 trillion parameters across 2048 experts, while Mixtral~\citep{jiang2024mixtral} refined this paradigm, balancing scalability and performance. Reasoning enhancements during this period, such as self-play~\citep{wu2024self} and Monte Carlo Tree Search (MCTS) integration with CoT~\citep{wiechowski2021MonteCT}, further bolstered LLMs’ decision-making capabilities by simulating iterative reasoning pathways, laying the groundwork for advanced inference-focused models.

A significant architectural advancement unfolded with the rise of Mixture of Experts (MoE) models, which diverge from traditional dense architectures by dynamically activating selective parameter subsets, thereby optimizing computational efficiency while accommodating expansive parameter scales. This paradigm was pioneered by Google’s Switch-C Transformer~\citep{fedus2022switch} in \textbf{2022}, featuring 1.6 trillion parameters distributed across 2048 experts, a groundbreaking approach that balanced resource demands with performance gains. Subsequent iterations, such as Mixtral~\citep{jiang2024mixtral} and DeepSeek V2.5~\citep{Shao2024DeepSeekV2AS}—the latter leveraging 236 billion total parameters with 21 billion active across 160 experts—further refined this framework, achieving state-of-the-art results on LMSYS benchmarks and demonstrating that sparse MoE architectures can rival dense models in both scalability and efficacy. These developments underscored a shift toward efficiency-focused PoLMs, enabling LLMs to handle complex tasks with reduced computational overhead, a critical step in broadening their practical applicability. By \textbf{2025}, DeepSeek-R1~\citep{DeepSeekAI2025DeepSeekR1IR} emerged as a landmark in PoLMs innovation, departing from conventional Supervised Fine-Tuning (SFT) reliance to embrace Chain-of-Thought (CoT) reasoning and exploratory RL strategies. Exemplified by DeepSeek-R1-Zero, which integrates self-verification, reflection, and extended CoT generation, this model validates RL-driven reasoning incentives within an open research paradigm, introducing distillation techniques~\citep{DeepSeekAI2025DeepSeekR1IR} to transfer sophisticated reasoning patterns from larger to smaller architectures. This approach not only yields superior performance compared to standalone RL training but also heralds a scalable, reasoning-centric paradigm for LLMs, poised to address the persistent challenges of computational efficiency and task adaptability in post-training methodologies.

\subsection{Formula Foundations of PoLMs}

\subsubsection{Principle of Policy Optimization}
\label{Principle of Policy Optimization}

The Proximal Policy Optimization (PPO) algorithm~\citep{Schulman2017ProximalPO} is a key reinforcement learning technique, particularly useful in settings such as Reinforcement Learning with Human Feedback (RLHF)~\citep{Ouyang2022TrainingLM}, where maintaining stability and efficiency is paramount. PPO achieves these objectives by constraining the size of policy updates, ensuring that changes to the model's behavior are gradual and controlled, thus preventing catastrophic shifts in performance. This is especially important when fine-tuning large-scale language models, where drastic policy updates could lead to undesirable or unpredictable behavior.

\noindent\textbf{Definition.}~~In the context of PPO, the state \(s_t \in \mathcal{S}\) represents the environment at time \(t\), which includes all relevant information the model needs to make a decision. The action \(a_t \in \mathcal{A}(s_t)\) denotes the choice the model makes given the state \(s_t\). This action is part of a sequence of decisions made by the model. Upon executing an action, the agent receives a reward \(r_t \in \mathbb{R}\), which serves as feedback from the environment, signaling the success or failure of the action taken.
The advantage function \(A^\pi(s, a)\) measures how advantageous it is to take action \(a\) in state \(s\) under the current policy \(\pi\), compared to the expected value of actions in that state. It is formally defined as the difference between the action-value function \(Q^\pi(s, a)\) and the state-value function \(V^\pi(s)\), which are defined as:
\begin{align}
\label{advantage function}
    A^\pi(s, a) &= Q^\pi(s, a) - V^\pi(s),
\end{align}
where \(Q^\pi(s, a)\) represents the expected cumulative reward obtained by taking action \(a\) in state \(s\) and following policy \(\pi\), and \(V^\pi(s)\) is the expected cumulative reward starting from state \(s\) and following policy \(\pi\). Both functions account for future rewards, discounted by a factor \(\gamma\).

\noindent\textbf{Policy Update.}~~The PPO algorithm optimizes the policy \(\pi_\theta\) by making incremental updates based on the advantage function. The policy update is performed using the clipped objective function:
\begin{equation}
\label{eq:policy_update}
L^{C L I P}(\theta) = \hat{\mathbb{E}}_{t}\left[\min \left(r_{t}(\theta) \hat{A}_{t}, \operatorname{clip}\left(r_{t}(\theta), 1-\epsilon, 1+\epsilon\right) \hat{A}_{t}\right)\right],
\end{equation}
where \(r_{t}(\theta)\) represents the ratio of the probability of taking action \(a_t\) under the current policy \(\pi_\theta\) to that under the old policy \(\pi_{\theta_{\text{old}}}\). This ratio is defined as:
\[
r_{t}(\theta) = \frac{\pi_{\theta}(a_t | s_t)}{\pi_{\theta_{\text{old}}}(a_t | s_t)}.
\]
The term \(\hat{A}_{t}\) is the estimated advantage at timestep \(t\), and the clipping function \(\operatorname{clip}(r_{t}(\theta), 1-\epsilon, 1+\epsilon)\) restricts the policy update to a safe range, controlled by the hyperparameter \(\epsilon\). This clipping mechanism ensures that updates do not diverge too much from the previous policy, thus maintaining stability during training.

\noindent\textbf{Value Function Update.}~~The value function \(V_\phi\) estimates the expected cumulative reward from a given state \(s_t\) under the policy \(\pi_\theta\). To ensure that the value function provides accurate estimates, it is optimized by minimizing the mean squared error between the predicted value and the actual reward:
\begin{equation}
\label{eq:valuefunction_update}
    \phi_{k+1} = \arg \min _{\phi} \mathbb{E}_{s_t \sim \pi_{\theta_k}} \left[\left(V_{\phi}(s_t) - R(s_t)\right)^2\right],
\end{equation}
where \(R(s_t)\) is the actual cumulative reward obtained from state \(s_t\), and \(V_\phi(s_t)\) is the estimated value under the current policy. The goal is to adjust the parameters \(\phi\) to minimize the discrepancy between predicted and actual rewards, improving the accuracy of the value function.

\subsubsection{Principle of RLHF}
Reinforcement Learning with Human Feedback (RLHF) is a crucial method for aligning models with human preferences by utilizing human-generated feedback in the learning process. This approach incorporates a reward function that explicitly captures human input, enabling the model to better adapt to user preferences and real-world applications.

\noindent\textbf{Definition.}~~In RLHF, a language model \(\rho\) generates a probability distribution over sequences of tokens in the vocabulary \(\Sigma\). The model \(\rho\) produces a sequence of tokens \(x_0, x_1, \dots, x_{n-1}\) from the input space \(X = \Sigma^{\leq m}\), where each token is conditionally dependent on previous tokens. The model's output is defined by the following conditional probability distribution:
\begin{equation}
\rho\left(x_0 \cdots x_{n-1}\right) = \prod_{0 \leq k < n} \rho\left(x_k \mid x_0 \cdots x_{k-1}\right).
\label{eq:ppo-policy-update}
\end{equation}
The model \(\rho\) is trained on a task defined by an input space \(X\), a data distribution \(D\) over \(X\), and an output space \(Y = \Sigma^{\leq n}\). For example, in text summarization, as shown in~\citep{ziegler2020finetuninglanguagemodelshuman}, a GPT-2 model~\citep{radford2019language} is trained using RLHF, where the task involves predicting text summaries based on a dataset such as CNN/DailyMail~\citep{hermann2015teaching} and TL;DR~\citep{volske2017tl}.

\noindent\textbf{Objective Function.}~~The policy \(\pi\) is a language model that shares the same structure as the original model \(\rho\). Initially, the policy \(\pi\) is set equal to \(\rho\). The objective is to maximize the expected reward \(R(x, y)\) for input-output pairs \((x, y)\) by optimizing the policy. The reward function \(R(x, y): X \times Y \to \mathbb{R}\) assigns a scalar value to each input-output pair, and the optimal policy \(\pi^*\) is obtained by solving the following maximization problem:
\begin{equation}
\pi^* = \max_{\pi} \mathbb{E}[R] = \mathbb{E}_{x \sim \mathcal{D}, y \sim \pi(\cdot \mid x)}[R(x, y)].
\label{eq:RLHF_objective}
\end{equation}
This objective function represents a standard RL problem, where the model learns to maximize the expected reward through interaction with the environment, guided by human feedback.

\subsubsection{Principle of DPO}
Direct Preference Optimization (DPO) builds upon RLHF by directly optimizing the model's outputs based on human preferences, which are often expressed in the form of pairwise comparisons. DPO eliminates the need for traditional reward functions, focusing instead on optimizing model behavior by maximizing preference-based rewards.

\noindent\textbf{Objective Function.}~~We begin with the same RL objective as in previous methods~\citep{peng2019advantageweightedregressionsimplescalable,go2023aligninglanguagemodelspreferences,jaques2020humancentricdialogtrainingoffline}, under a general reward function \(r\). The optimal solution to the KL-constrained reward maximization objective is given by:
\begin{equation}
    \pi_r(y \mid x) = \frac{1}{Z(x)} \pi_{\mathrm{ref}}(y \mid x) \exp \left(\frac{1}{\beta} r(x, y)\right),
\end{equation}
where \(Z(x)\) is the partition function that ensures the output is normalized across all possible actions. Even when utilizing a maximum likelihood estimate \(r_{\phi}\) of the true reward \(r^*\), the partition function \(Z(x)\) can be approximated, simplifying the optimization process. This formulation allows for more efficient preference optimization by directly adjusting the policy based on human feedback.

\noindent\textbf{Preference Model.}~~Using the Bradley-Terry model, which models preferences between two outputs \(y_1\) and \(y_2\), the optimal policy \(\pi^*\) satisfies the following preference model:
\begin{equation}
    p^*(y_1 \succ y_2 \mid x) = \frac{1}{1 + \exp \left(\beta \log \frac{\pi^*(y_2 \mid x)}{\pi_{\text{ref}}(y_2 \mid x)} - \beta \log \frac{\pi^*(y_1 \mid x)}{\pi_{\text{ref}}(y_1 \mid x)} \right)},
\end{equation}
where \(p^*(y_1 \succ y_2 \mid x)\) represents the probability that the human prefers output \(y_1\) over \(y_2\) given the input \(x\). This approach effectively incorporates human preferences into the model's optimization process.

\subsubsection{Principle of GRPO}
The Group Relative Policy Optimization (GRPO) algorithm is a variant of the Proximal Policy Optimization (PPO) algorithm in reinforcement learning, first introduced in DeepSeek's previous work, \textit{DeepSeekMath: Pushing the Limits of Mathematical Reasoning in Open Language Models} \cite{shao2024deepseekmath}. GRPO omits the critic model, instead estimating the baseline using group scores, which significantly reduces training resource consumption compared to PPO.

\noindent\textbf{Definition.}~~The most significant difference between the GRPO and PPO algorithms lies in the method used to calculate the advantage function.
As we can see from Equation \ref{advantage function} in Section \ref{Principle of Policy Optimization}, the value of the advantage function $A^\pi(s, a)$ in PPO is derived from the difference between the Q-value and the V-value.

\noindent\textbf{Objective Function.}~~Specifically, for each question $q$, GRPO samples a group of outputs $\{o_1, o_2, \dots , o_G\}$ from the old policy $\pi_{\theta_{old}}$
and then optimizes the policy model by maximizing the following objective:
\begin{equation}
\begin{split}
\mathcal{J}_{GRPO}(\theta) &= \mathbb{E}[q \sim P(Q), \{o_i\}^G_{i=1} \sim \pi_{\theta_{\text{old}}}(O|q)] \\
&\frac{1}{G} \sum_{i=1}^{G} \frac{1}{|o_i|} \sum_{t=1}^{|o_i|} \biggl\{
\min \left[
\frac{\pi_{\theta}(o_{i,t}|q, o_{i,<t})}{\pi_{\theta_{\text{old}}}(o_{i,t}|q, o_{i,<t})}
\hat{A}_{i,t},
\text{clip}\left(
\frac{\pi_{\theta}(o_{i,t}|q, o_{i,<t})}{\pi_{\theta_{\text{old}}}(o_{i,t}|q, o_{i,<t})},
1-\epsilon, 1+\epsilon \right) \hat{A}_{i,t}
\right] \\
& - \beta D_{KL}[\pi_{\theta} \parallel \pi_{\text{ref}}] \biggl\},
\end{split}
\end{equation}
where $\epsilon$ and $\beta$ are hyper-parameters, and $\hat{A}_{i,t}$ is the advantage calculated based on relative
rewards of the outputs inside each group only, which will be detailed in the subsection \ref{Section 5.2}.

\section{PoLMs for Fine-Tuning}\label{Section 3}
Fine-tuning constitutes a cornerstone of adapting pre-trained Large Language Models (LLMs) to specialized tasks, refining their capabilities through targeted parameter adjustments. This process leverages labeled or task-specific datasets to optimize performance, bridging the gap between general-purpose pre-training and domain-specific requirements. This chapter explores three principal fine-tuning paradigms: \textbf{Supervised Fine-Tuning} (\S \ref{Section 3.1}), which employs annotated datasets to enhance task-specific accuracy; \textbf{Adaptive Fine-Tuning} (\S \ref{Section 3.2}), which customizes model behavior via instruction tuning and prompt-based methods; and \textbf{Reinforcement Fine-Tuning} (\S \ref{Section 3.3}), which integrates reinforcement learning to iteratively refine outputs based on reward signals, fostering continuous improvement through dynamic interaction.

\subsection{Supervised Fine-Tuning}\label{Section 3.1}
Supervised Fine-Tuning (SFT)~\citep{Ouyang2022TrainingLM} adapts pre-trained LLMs to specific tasks by leveraging task-specific labeled datasets. Distinct from instruction tuning, which relies on directive prompts, SFT directly adjusts model parameters using annotated data, yielding models that are both precise and contextually attuned while preserving broad generalization capabilities.
SFT bridges the divide between the expansive linguistic knowledge encoded during pre-training and the nuanced demands of targeted applications~\citep{han2021pre}. Pre-trained LLMs, through exposure to vast corpora, acquire generalized language patterns, reducing reliance on extensive domain-specific data for fine-tuning. Model selection is pivotal: smaller models like T5~\citep{Raffel2019ExploringTL} excel in resource-constrained settings with limited datasets, whereas larger models, such as GPT-4~\citep{achiam2023gpt}, leverage their superior capacity to excel in complex, data-rich tasks.

\begin{table}[h]
\caption{Summary of Pre-trained Large Language Model Releases by Various Organizations (2018–2025). This table details key models from Meta, DeepSeek, OpenAI, and other entities, including their parameter sizes, training data scales (where reported), open-source status, and release timelines. Open-source status is denoted by \faCheckCircle{} for models publicly accessible to the research community and \faTimesCircle[regular]{} for closed-source proprietary models.}
\centering
\label{tbl:pre-training model list}
\resizebox{\textwidth}{!}{
\begin{tabular}{cccccr}
\toprule
\textbf{Company}  &\textbf{Model}  & \textbf{Size (B)}  & \textbf{Data Scale}  & \textbf{Open Resource} & \textbf{Release Time} \\
\midrule
\rowcolor{Gray} & \textbf{LLaMA}~\citep{Touvron2023LLaMAOA}     &7B,13B,30B,65B & 1.4T tokens   &\faCheckCircle  &Feb-2023\\
\rowcolor{LightGray} & \textbf{LLaMA2}~\citep{Touvron2023Llama2O}     &7B,13B,70B & 2T tokens   &\faCheckCircle  &Jul-2023\\
\rowcolor{Gray} & \textbf{LLaMA3}~\citep{dubey2024llama}     &8B,70B,405B & 15T tokens   &\faCheckCircle  &Apr-2024\\
\multirow{-5}{*}{\begin{tabular}[c]{@{}c@{}}\textbf{Meta}\end{tabular}}   \\
\noalign{\vskip-1.20em}
\midrule
\rowcolor{LightGray}& \textbf{DeepSeek-V1}~\citep{Bi2024DeepSeekLS}  & 7B,67B    & 2T tokens &\faCheckCircle  & Dec-2023\\
\rowcolor{Gray}& \textbf{DeepSeek-V2}~\citep{Shao2024DeepSeekV2AS}  & 236B    & 8.1T tokens &\faCheckCircle  & May-2024\\
\rowcolor{LightGray}& \textbf{DeepSeek-V3}~\citep{DeepSeekAI2024DeepSeekV3TR} & 671B    & 14.8T tokens &\faCheckCircle  & Dec-2024 \\
\rowcolor{Gray}& \textbf{DeepSeek-R1}~\citep{DeepSeekAI2025DeepSeekR1IR}  & 671B    & - &\faCheckCircle  & Jan-2025\\
\multirow{-6}{*}{\begin{tabular}[c]{@{}c@{}}\textbf{DeepSeek}\end{tabular}}   \\
\noalign{\vskip-1.20em}
\midrule
\rowcolor{LightGray}& \textbf{Qwen1.5}~\citep{Bai2023QwenTR}  & 0.5B,1.8B,4B,7B,14B,72B    & 3T tokens&\faCheckCircle  & Sep-2023\\
\rowcolor{Gray}& \textbf{Qwen2}~\citep{Yang2024Qwen2TR}  & 0.5B,1.5B,7B,57B,72B    & 7T tokens &\faCheckCircle & Jul-2024 \\
\rowcolor{LightGray}& \textbf{Qwen2.5}~\citep{Yang2024Qwen25TR}  & 0.5B,1.5B,3B,7B,14B,72B    & 18T tokens &\faCheckCircle  & Dec-2024\\

\multirow{-5}{*}{\begin{tabular}[c]{@{}c@{}}\textbf{Qwen}\end{tabular}}   \\
\noalign{\vskip-1.20em}
\midrule
\rowcolor{Gray}&\textbf{Mistral-7B}~\citep{jiang2023mistral}  & 7B   & - &\faCheckCircle & Sep-2023\\
\rowcolor{LightGray}&\textbf{Mistral-8X7B}~\citep{jiang2024mixtral}  & 4.7B    & - &\faCheckCircle & Dec-2023\\
\rowcolor{Gray}&\textbf{Mistral-8X22B}~\citep{jiang2024mixtral}  & 14B   & - &\faCheckCircle  & Apr-2024\\
\rowcolor{LightGray}&\textbf{Mistral-Large2}~\citep{Mistral-Large2}  & 12.3B    & - &\faCheckCircle & Jul-2024\\
\multirow{-6}{*}{\begin{tabular}[c]{@{}c@{}}\textbf{Mistral}\end{tabular}}   \\
\noalign{\vskip-1.20em}
\midrule
\rowcolor{Gray}&\textbf{Claude2}~\citep{claude2}  & 2B    & - &\faTimesCircle[regular]  & Jul-2023\\
\rowcolor{LightGray}&\textbf{Claude3}~\citep{TheC3}  & -   & - &\faTimesCircle[regular]  & Mar-2024\\
\rowcolor{Gray}&\textbf{Claude3.5}~\citep{AhtropicClaude}  & 175B    & - &\faTimesCircle[regular]  & Oct-2024\\
\multirow{-5}{*}{\begin{tabular}[c]{@{}c@{}}\textbf{Anthropic}\end{tabular}}   \\
\noalign{\vskip-1.20em}
\midrule
\rowcolor{LightGray}&\textbf{Gemini1.0}~\citep{team2023gemini} & -   & - &\faTimesCircle[regular]  & Dec-2023 \\
\rowcolor{Gray}&\textbf{Gemini1.5}~\citep{team2024gemini15}  & -    & - &\faTimesCircle[regular]  & Mar-2024\\
\rowcolor{LightGray}&\textbf{Gemini2.0}~\citep{gemini2.0}  & -    & - &\faTimesCircle[regular] & Dec-2024\\
\multirow{-5}{*}{\begin{tabular}[c]{@{}c@{}}\textbf{Google}\end{tabular}}   \\
\noalign{\vskip-1.20em}
\midrule
\rowcolor{Gray}&\textbf{GPT-1}~\citep{radford2018improving}  & 0.01B    & - &\faTimesCircle[regular]  & Jun-2018\\
\rowcolor{LightGray} &\textbf{GPT-2}~\citep{radford2019language} & 0.15B    & - &\faTimesCircle[regular]  & Feb-2019 \\
\rowcolor{Gray}&\textbf{GPT-3}~\citep{brown2020language}  & 17.5B    & - &\faTimesCircle[regular] & Jul-2020\\
\rowcolor{LightGray} &\textbf{InstructGPT}~\citep{Ouyang2022TrainingLM}  & -    & - &\faTimesCircle[regular]  & Mar-2022\\
\rowcolor{Gray} &\textbf{GPT-4}~\citep{achiam2023gpt}  & 180B   & - &\faTimesCircle[regular]  & Mar-2023\\
\rowcolor{LightGray}&\textbf{GPT-4o}~\citep{openai2024gpt4o}  & 200B    & - &\faTimesCircle[regular] & Oct-2024\\
\rowcolor{Gray}&\textbf{o1}~\citep{jaech2024openai}  & 300B    & - &\faTimesCircle[regular]  & Dec-2024\\
\rowcolor{LightGray} &\textbf{o3-mini}~\citep{openaio3}  & -    & - &\faTimesCircle[regular]  & Jan-2025\\
\multirow{-10}{*}{\begin{tabular}[c]{@{}c@{}}\textbf{OpenAI}\end{tabular}}   \\
\noalign{\vskip-1.20em}
\midrule
\rowcolor{Gray}&\textbf{GLM-4}\citep{Zeng2024ChatGLMAF}  & 130B   & - &\faCheckCircle  & Jun-2024\\
\multirow{-3}{*}{\begin{tabular}[c]{@{}c@{}}\textbf{ZhiPuAI}\end{tabular}}   \\
\noalign{\vskip-1.20em}
\midrule
\rowcolor{LightGray}&\textbf{DBRX}~\citep{Databricks/DBRX}  & 132B    & - &\faCheckCircle  & Mar-2024\\
\multirow{-3}{*}{\begin{tabular}[c]{@{}c@{}}\textbf{Databricks}\end{tabular}}   \\
\noalign{\vskip-1.20em}
\midrule
\rowcolor{Gray}&\textbf{Yi-Large}~\citep{young2024yi}  & -    & - &\faCheckCircle & Mar-2024\\
\multirow{-3}{*}{\begin{tabular}[c]{@{}c@{}}\textbf{01.AI}\end{tabular}}   \\
\noalign{\vskip-1.20em}
\midrule
\rowcolor{LightGray}&\textbf{Jamba1.5 Large}~\citep{Lenz2024Jamba15HT}  & 94B    & - &\faCheckCircle  & Aug-2024\\
\multirow{-3}{*}{\begin{tabular}[c]{@{}c@{}}\textbf{AI21 Labs}\end{tabular}}   \\
\noalign{\vskip-1.20em}
\midrule
\rowcolor{Gray}&\textbf{Nova Pro}~\citep{Intelligence2024}  & -    & - &\faTimesCircle[regular] & Dec-2024\\
\multirow{-3}{*}{\begin{tabular}[c]{@{}c@{}}\textbf{Amazon}\end{tabular}}   \\
\noalign{\vskip-1.20em}
\midrule
\rowcolor{LightGray}&\textbf{Kimi-k1.5}~\citep{Team2025KimiKS}  & -    & - &\faTimesCircle[regular]  & Jan-2025\\
\multirow{-3}{*}{\begin{tabular}[c]{@{}c@{}}\textbf{MoonshotAI}\end{tabular}}   \\
\noalign{\vskip-1.20em}
\bottomrule
\end{tabular}
}
\end{table}

\subsubsection{Dataset Preparation for SFT}
Crafting a high-quality SFT dataset is a multi-faceted process critical to fine-tuning success.

\noindent\textbf{SFT Dataset Construction.}~~An SFT dataset is typically structured as \( \mathcal{D} = \{(I_k, X_k)\}_{k=1}^{N} \), where \( I_k \) is an instruction and \( X_k \) is its corresponding instance. This pairing enables the LLM to discern task-specific patterns and generate relevant outputs. Methods like Self-Instruct~\citep{Wang2022SelfInstructAL} enrich diversity by synthesizing novel instruction-output pairs, with duplicates filtered using metrics such as ROUGE-L~\citep{Ganesan2015ROUGE2U} to maintain variety.

\noindent\textbf{SFT Dataset Screening.}~~Screening ensures that only high-quality instruction–instance pairs remain in the final dataset. A screening function \(r(\cdot)\) is used to evaluate the quality of each pair \((I_k, X_k)\), yielding a curated subset \(\mathcal{D}'\):
\begin{equation}
\mathcal{D}' = \bigl\{\,(I_k, X_k) \in \mathcal{D} \mid r(I_k, X_k) \ge \tau \bigr\},
\end{equation}
where \(\tau\) is a user-defined quality threshold. For example, the \textit{Instruction Following Difficulty} (IFD) metric~\citep{Li2023FromQT} quantifies how effectively a given instruction guides the model toward generating the expected response. The IFD function is expressed as:
\begin{equation}
    r_{\theta}(Q, A) 
    = \frac{\sum_{i=1}^{N} \log P\bigl(w_i^A \mid Q,\; w_1^A,\; \ldots,\; w_{i-1}^A;\; \theta\bigr)}
           {\sum_{i=1}^{N} \log P\bigl(w_i^A \mid w_1^A,\; \ldots,\; w_{i-1}^A;\; \theta\bigr)},
\end{equation}
where \(Q\) denotes the instruction, \(A\) is the expected response, and \(\theta\) represents the model's learnable parameters. This metric compares the likelihood of generating a response both with and without the instruction, thereby providing a normalized measure of how effectively the instruction facilitates response generation. Instruction–instance pairs that do not meet the selected IFD threshold are excluded, resulting in a refined dataset \(\mathcal{D}'\).

\noindent\textbf{SFT Dataset Evaluation.}~~Evaluating an SFT dataset involves selecting a high-quality subset, \(\mathcal{D}_{\text{eval}}\), to serve as a benchmark for model performance. This subset can be sampled from the curated dataset \(\mathcal{D}'\) or derived from an independent portion to ensure impartiality. Traditional methods for SFT evaluation, such as Few-Shot GPT~\citep{brown2020language} and Fine-tuning strategies~\citep{Du2023MoDSMD}, are resource-intensive, whereas Instruction Mining~\citep{Cao2023InstructionMI} offers a more efficient alternative. Instruction Mining uses linear quality rules and a set of metrics to measure dataset quality, such as response length and average reward model scores~\citep{Touvron2023LLaMAOA}, to evaluate correlations between these metrics and overall dataset quality.

\begin{figure}[h]
\centering
\includegraphics[width=0.85\linewidth]{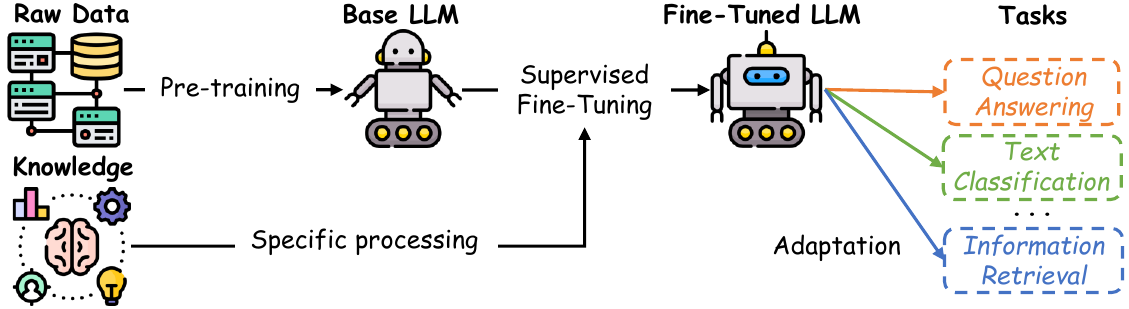}
\caption{Process of Supervised Fine-Tuning.}
\label{fig:SFT}
\end{figure}

\subsubsection{Process of SFT}
As depicted in \textbf{Fig. \ref{fig:SFT}}, once the dataset is prepared, the fine-tuning process begins with a pre-trained LLM, typically obtained via unsupervised or self-supervised pre-training on large-scale raw datasets. The objective of this pre-training phase is to acquire general feature representations applicable across various tasks~\citep{han2021pre}. Subsequently, during the fine-tuning phase, the model's parameters are adjusted using task-specific annotated data, which aligns the model with the requirements of a given application. The objective function commonly used in this phase is the cross-entropy loss. For a classification task with \(N\) samples and \(C\) categories, it can be expressed as:
\begin{equation}
    L_{\text{fine-tune}}(\theta) 
    = -\frac{1}{N} \sum_{i=1}^{N} \sum_{j=1}^{C} 
      y_{ij} \log P\bigl(y_j \mid x_i; \theta\bigr),
\end{equation}
where \(y_{ij}\) is the true label of sample \(i\) in category \(j\), and \(P\bigl(y_j \mid x_i; \theta\bigr)\) represents the model’s predicted probability of sample \(i\) belonging to category \(j\). Minimizing this loss function drives the model toward better alignment with the ground-truth labels, improving performance on the target task.

A prominent example is the BERT model~\citep{devlin2018bert}, which undergoes extensive pre-training on broad linguistic corpora such as BooksCorpus and Wikipedia. During the fine-tuning phase, these broad representations are refined using task-specific data (e.g., the IMDB dataset~\citep{maas-etal-2011-learning} for sentiment analysis), allowing BERT to specialize in tasks such as sentiment classification and question answering.

\subsubsection{Full-Parameter Fine-Tuning}
Full-parameter fine-tuning refers to the process of adjusting all parameters of a pre-trained model, in contrast to parameter-efficient methods such as LoRA~\citep{Hu2021LoRALA} or Prefix-tuning~\citep{Li2021PrefixTuningOC}, which modify only a subset of parameters. Full-parameter tuning is often preferred for tasks requiring high precision, such as those in the medical and legal domains~\citep{Lv2023FullPF}, but it entails substantial computational overhead. For example, fine-tuning a 65-billion-parameter model may require over 100 GB of GPU memory, creating challenges for resource-constrained environments.
To mitigate such constraints, memory optimization techniques like LOMO~\citep{Lv2023FullPF} have been introduced, which reduce the memory footprint of gradient calculations and optimizer states. The model’s parameters are updated according to the rule:
\begin{equation}
    \theta_{t+1} 
    = \theta_{t} - \eta \nabla_{\theta} L(\theta_t),
\end{equation}
where \(\theta_t\) represents the model parameters at iteration \(t\), \(\eta\) is the learning rate, and \(\nabla_{\theta} L(\theta_t)\) denotes the gradient of the loss function. Memory optimization techniques, including Mixed Precision Training~\citep{Micikevicius2017MixedPT} and Activation Checkpointing~\citep{Chen2016TrainingDN}, help reduce memory demands, enabling large models to be fine-tuned on systems with limited hardware resources.

\noindent\textbf{GPT-3 to InstructGPT.}~~A notable example of full-parameter fine-tuning is the transition from GPT-3 to InstructGPT~\citep{Ouyang2022TrainingLM}, where the model’s entire parameter set was fine-tuned using a dataset designed for instruction-following tasks. This approach leads to optimal performance but is computationally expensive due to the need to update all parameters.

\subsection{Adaptive Fine-Tuning}\label{Section 3.2}
Adaptive fine-tuning modifies a pre-trained model’s behavior to better address user-specific needs and handle a broader range of tasks. This approach introduces additional cues to guide the model's output generation, offering a flexible framework for customizing the model’s responses. Notable methods in adaptive fine-tuning include instruction tuning and prompt-based tuning, both of which significantly enhance the adaptability of LLMs by introducing task-specific guidance.

\begin{figure}[ht]
\centering
\includegraphics[width=0.85\linewidth]{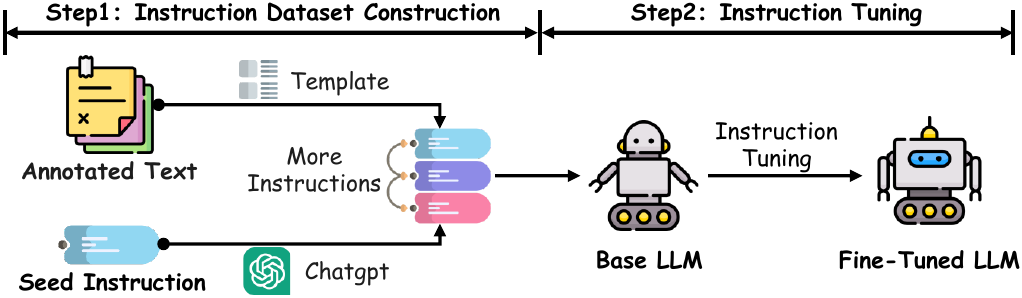}
\caption{Workflow of Instruction Fine-tuning, illustrating the general pipeline for Instruction Dataset Construction and Instrction Tuning in Large Language Models.}
\label{fig:Instruction tuning}
\end{figure}

\subsubsection{Instruction Tuning}
Instruction Tuning~\citep{Zhang2023InstructionTF} is a technique that refines a base LLM by fine-tuning it on specially constructed instruction datasets. This method substantially boosts the model’s ability to generalize across a variety of tasks and domains, improving its flexibility and accuracy. As shown in \textbf{Fig. \ref{fig:Instruction tuning}}, the process begins by transforming existing NLP datasets (e.g., those for text classification, translation, and summarization) into natural language instructions that include task descriptions, input examples, expected outputs, and illustrative demonstrations. Techniques like Self-Instruct~\citep{Wang2022SelfInstructAL} further enhance the diversity of these datasets by automatically generating additional instruction–output pairs, expanding the model’s exposure to a broader range of tasks. The fine-tuning procedure adapts the model's parameters to align with these task-specific instructions, resulting in an LLM that performs robustly across both familiar and previously unseen tasks. For instance, InstructGPT~\citep{Ouyang2022TrainingLM} and GPT-4~\citep{brown2020language} have shown significant improvements in instruction-following capabilities across a wide array of applications.

The effectiveness of Instruction Tuning largely depends on the quality and breadth of the instruction datasets. High-quality datasets should encompass a wide range of languages, domains, and task complexities to ensure that the model remains broadly applicable~\citep{Zhang2023InstructionTF}. Furthermore, the clarity and organization of instructions play a critical role in enabling the model to interpret and execute tasks effectively. Techniques such as integrating demonstration examples, including Chain-of-Thought prompting~\citep{Wei2022ChainOT}, can significantly improve performance on tasks requiring complex reasoning. Moreover, ensuring a balanced distribution of tasks during the fine-tuning phase is essential to avoid overfitting or diminishing model performance due to imbalanced task coverage. Techniques such as proportional task sampling or weighted loss functions are useful in addressing these issues, ensuring that each task contributes equitably to the fine-tuning procedure. Therefore, by constructing and managing instruction datasets meticulously, researchers can greatly enhance the generalization capabilities of fine-tuned LLMs, enabling them to excel across a wide range of tasks and domains~\citep{Honovich2022UnnaturalIT}.

\subsubsection{Prefix-Tuning}
\begin{figure}[h]
\centering
\includegraphics[width=0.95\linewidth]{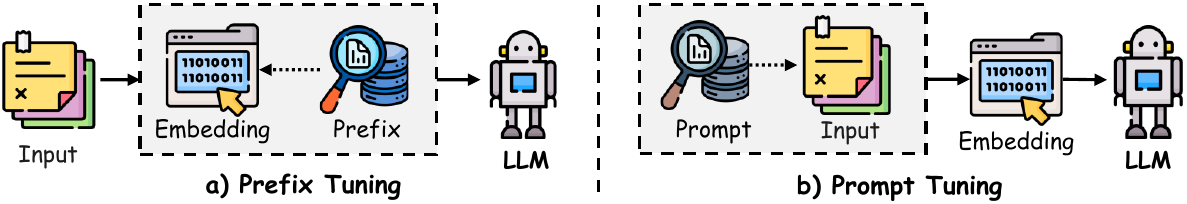}
\caption{Comparison of Prefix Tuning and Prompt Tuning, delineating their distinct approaches to parameter fine-tuning: a) Prefix Tuning and b) Prompt Tuning.}
\label{fig:PPT}
\end{figure}

Prefix-tuning~\citep{zhang2023towards1} is a parameter-efficient fine-tuning method that involves adding a sequence of trainable prefix tokens (continuous vectors) to each Transformer layer in the language model, while keeping the core model parameters fixed. As depicted in \textbf{Fig. \ref{fig:PPT} (a)}, these prefix vectors are task-specific and function as virtual token embeddings. To optimize the prefix vectors, a reparameterization trick is used, wherein a small multi-layer perceptron (MLP) function is learned to map a smaller matrix to the prefix parameters instead of directly optimizing the prefix vectors. This method has been shown to stabilize the training process. Once the prefix vectors are optimized, the mapping function is discarded, and only the derived prefix vectors are retained for enhancing task-specific performance.

By prepending a learned continuous prompt to the input sequence and utilizing layer-wise prompts, the model's behavior is steered toward task-specific outputs without requiring full-model fine-tuning. Since only the prefix parameters are adjusted, this results in a more parameter-efficient approach. Building upon this, P-Tuning v2~\citep{liu2021p} incorporates layer-wise prompt vectors into the Transformer architecture specifically for natural language understanding tasks. This approach also leverages multi-task learning to optimize shared prompts across tasks, enhancing model performance across different parameter scales~\citep{Li2021PrefixTuningOC}. The potential of prefix-tuning to facilitate rapid and efficient adaptation of large language models for specific tasks is evident, making it a compelling strategy for applications requiring flexibility and efficiency.

\subsubsection{Prompt-Tuning}
Prompt-tuning~\citep{lester2021power, liu2024gpt} is a method designed to adapt large language models efficiently by optimizing trainable vectors at the input layer rather than modifying the model's internal parameters. As shown in \textbf{Fig. \ref{fig:PPT} (b)}, this technique builds on discrete prompting methods~\citep{jiang2020can, shin2020autoprompt} by introducing soft prompt tokens, which can be structured either in an unrestricted format~\citep{lester2021power} or as a prefix~\citep{liu2024gpt}. These learned prompt embeddings are combined with the input text embeddings before being processed by the model, thereby guiding the model’s output while keeping the pre-trained weights frozen.
Two notable implementations of prompt-tuning are P-tuning~\citep{lester2021power}, which uses a flexible method to combine context, prompt, and target tokens, making it suitable for both understanding and generation tasks. This method enhances the learning of soft prompt representations through a bidirectional LSTM architecture. In contrast, standard prompt-tuning~\citep{liu2024gpt} employs a simpler design, wherein prefix prompts are prepended to the input, and only the prompt embeddings are updated during training based on task-specific supervision.

Research has shown that prompt-tuning can match the performance of full-parameter fine-tuning across many tasks, while requiring significantly fewer trainable parameters. However, its success is closely tied to the underlying language model's capacity, as prompt-tuning only modifies a small number of parameters at the input layer~\citep{lester2021power}. Building on these advancements, newer approaches such as P-Tuning v2~\citep{liu2021p} have demonstrated that prompt-tuning strategies can scale effectively across various model sizes, handling complex tasks previously thought to require full fine-tuning. These findings establish prompt-tuning as a highly efficient alternative to traditional fine-tuning, offering comparable performance with reduced computational and memory costs.

\subsection{Reinforcement Fine-Tuning}\label{Section 3.3}
Reinforcement Fine-Tuning (ReFT)~\citep{Luong2024ReFTRW} represents an advanced technique that integrates RL with SFT to enhance the model's ability to solve complex, dynamic problems. Unlike traditional SFT, which typically uses a single CoT annotation for each problem, ReFT enables the model to explore multiple valid reasoning paths, thereby improving its generalization capacity and problem-solving skills.
The ReFT process begins with the standard SFT phase, where the model is initially trained on labeled data to learn fundamental task-solving abilities through supervised annotations. Following this initial fine-tuning, the model undergoes further refinement using RL algorithms, such as Proximal Policy Optimization (PPO)~\citep{Schulman2017ProximalPO}. During the reinforcement phase, the model generates multiple CoT annotations for each problem, exploring different potential reasoning paths. These generated paths are evaluated by comparing the model’s predicted answers to the true answers, with rewards assigned for correct outputs and penalties for incorrect ones. This iterative process drives the model to adjust its policy, ultimately improving its reasoning strategy.

\begin{figure}[h]
\centering
\includegraphics[width=0.88\linewidth]{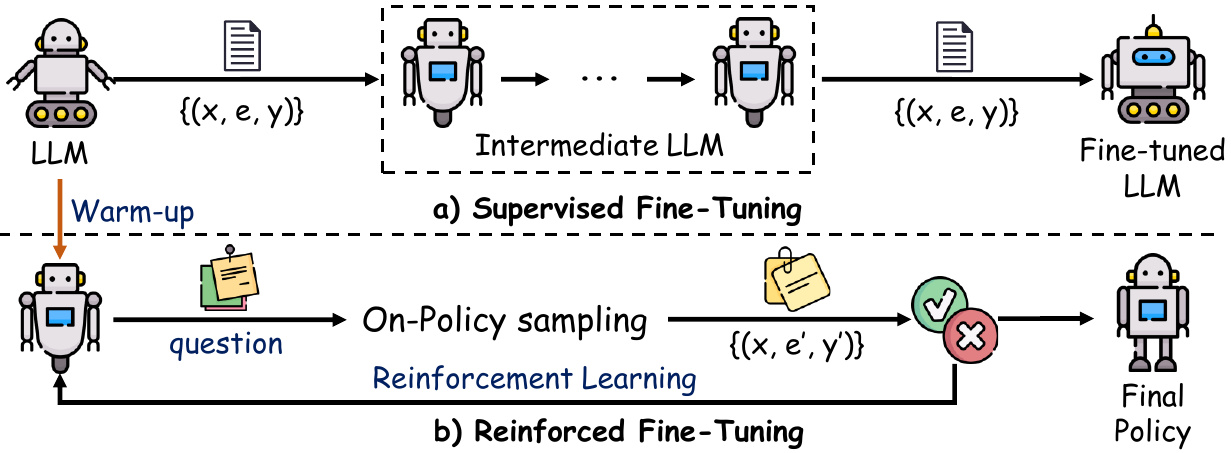}
\caption{Process of Reinforcement Fine-Tuning (ReFT), depicting the iterative Supervised Fine-Tuning (SFT) warm-up followed by RL training on identical datasets.}
\label{fig:ReFT}
\end{figure}

As shown in \textbf{Fig. \ref{fig:ReFT}}, the ReFT process is executed in two stages. The upper section represents the SFT phase, where the model iterates over the training data to learn the correct CoT annotation for each problem over several epochs. In the lower section, the ReFT phase is introduced: starting from the SFT-trained model, the model generates alternative CoT annotations (\(e'\)) based on its current policy and compares its predicted answers (\(y'\)) with the true answers (\(y\)). Positive rewards are given for correct answers, and negative rewards for incorrect answers, driving the model to improve its performance. These reward signals are then used to update the model's policy through reinforcement learning, enhancing its ability to generate accurate and diverse CoT annotations.

Recent studies have demonstrated that ReFT significantly outperforms traditional SFT approaches~\citep{Luong2024ReFTRW}. Moreover, the integration of inference-time strategies, such as majority voting and re-ranking, can further enhance performance, allowing the model to refine its outputs after training. Notably, ReFT achieves these improvements without requiring additional or augmented training data, learning solely from the existing dataset used during the SFT phase. This highlights the model’s superior generalization ability, as it learns more efficiently and effectively from the available data.

\section{PoLMs for Alignment}\label{Section 4}
Alignment in LLMs involves guiding model outputs to conform to human expectations and preferences, particularly in safety-critical or user-facing applications. This chapter discusses three major paradigms for achieving alignment: \textbf{Reinforcement Learning with Human Feedback} (\S\ref{Section 4.1}), which employs human-labeled data as a reward signal; \textbf{Reinforcement Learning with AI Feedback} (\S\ref{Section 4.2}), which leverages AI-generated feedback to address scalability issues; and \textbf{Direct Preference Optimization} (\S\ref{Section 4.3}), which learns directly from pairwise human preference data without requiring an explicit reward model. Each paradigm offers distinct advantages, challenges, and trade-offs in its pursuit of robust alignment. A concise comparison of these and related methods is summarized in \textbf{Table ~\ref{tab:alignment_comparison}}.

\begin{table*}[htbp]
\caption{Comparative Overview of Alignment Methods for Large Language Models (2022–2024). This table evaluates prominent alignment techniques across eight metrics: \textbf{RM1} (Explicit or Implicit Reward Model), \textbf{RM2} (Point Reward or Preference Probability Model), \textbf{RM3} (Response- or Token-level Reward), \textbf{RM4} (Positive or Negative Reward Model), \textbf{F} (Feedback Type: Human or AI), \textbf{RL1} (Reference Model or Reference Model-Free RL), \textbf{RL2} (On-policy or Off-policy RL), and \textbf{O} (Online/Iterative or Offline/Non-iterative Optimization).}
\label{tab:alignment_comparison}
\centering
\scriptsize
\renewcommand{\arraystretch}{1.2}
\resizebox{\textwidth}{!}{
\begin{tabular}{lccccccccr}
\toprule
\textbf{Methods} & \textbf{RM1} & \textbf{RM2} & \textbf{RM3} & \textbf{RM4} & \textbf{F} & \textbf{RL1} & \textbf{RL2} & \textbf{O} & \textbf{Release Time} \\
\midrule
\rowcolor{Gray} InstructGPT~\citep{Ouyang2022TrainingLM} & Explicit & Point & Response & Positive & Human & Reference & On & Offline & Mar-2022 \\
\rowcolor{LightGray} RLHF: Anthropic~\citep{bai2022training} & Explicit & Point & Response & Positive & Human & Reference & Off & Hybrid & Apr-2022 \\
\rowcolor{Gray} RLAIF-Anthropic~\citep{bai2022constitutional} & Explicit & Point & Response & Positive & AI & Reference & Off & Offline & Dec-2022 \\
\rowcolor{LightGray} RRHF~\citep{yuan2023rrhf} & -- & -- & -- & -- & Human & Free & Off & Offline & Apr-2023 \\
\rowcolor{Gray} DPO~\citep{rafailov2024directpreferenceoptimizationlanguage} & Implicit & Point & Response & Positive & Human & Reference & Off & Offline & May-2023 \\
\rowcolor{LightGray} PRO~\citep{song2024preference} & Explicit & Point & Response & Positive & Human & Free & Off & Offline & Jun-2023 \\
\rowcolor{Gray} RLAIF-Google~\citep{lee2023rlaif} & Explicit & Point & Response & Positive & AI & Reference & Off & Offline & Sep-2023 \\
\rowcolor{LightGray} CRINGE~\citep{xu2024thingscringeothersiterative} & Implicit & Point & Response & Positive & AI & Reference & Off & Online & Dec-2023 \\
\rowcolor{Gray} CPO~\citep{xu2024contrastive} & Implicit & Point & Response & Negative & Human & Reference & Off & Offline & Jan-2024 \\
\rowcolor{LightGray} Iterative DPO~\citep{yuan2024self} & Implicit & Point & Response & Positive & AI & Reference & Off & Online & Jan-2024 \\
\rowcolor{Gray} RLOO~\citep{ahmadian2024back} & Explicit & Point & Response & Positive & Human & Free & Off & Offline & Feb-2024 \\
\rowcolor{LightGray} LiPO~\citep{liu2024lipo} & Implicit & Point & Response & Positive & Human & Reference & Off & Offline & Feb-2024 \\
\rowcolor{Gray} GRPO~\citep{shao2024deepseekmathpushinglimitsmathematical} & Implicit & Point & Response & Positive & AI & Reference & Off & Online & Feb-2024 \\
\rowcolor{LightGray} NN~\citep{duan2024negating} & Implicit & Point & Response & Negative & Human & Reference & On & Offline & Mar-2024 \\
\rowcolor{Gray} R-DPO~\citep{park2024disentangling} & Implicit & Point & Response & Positive & Human & Reference & Off & Offline & Mar-2024 \\
\rowcolor{LightGray} DNO~\citep{rosset2024directnashoptimizationteaching} & -- & Preference & Response & Positive & Human & Reference & Hybrid & Offline & Apr-2024 \\
\rowcolor{Gray} DPO: from r to Q~\citep{rafailov2024r} & Implicit & Point & Token & Positive & Human & Reference & Off & Offline & Apr-2024 \\
\rowcolor{LightGray} TDPO~\citep{zeng2024token} & Implicit & Point & Token & Positive & Human & Reference & Off & Offline & Apr-2024 \\
\rowcolor{Gray} NPO~\citep{zhang2024negative} & Implicit & Point & Response & Negative & Human & Reference & Off & Offline & Apr-2024 \\
\rowcolor{LightGray} SIMPO~\citep{meng2024simpo} & -- & -- & -- & -- & Human & Free & Off & Offline & May-2024 \\
\rowcolor{Gray} UNA~\citep{wang2024unaunifyingalignmentsrlhfppo} & Implicit & Preference & Response & Positive and Negative & Human and AI & Free & Off & Offline & Sep-2024 \\
\rowcolor{LightGray} Aligner~\citep{ji2024alignerefficientalignmentlearning} & Implicit & Preference & Response & Positive and Negative & AI & free & Off & Offline & Nov-2024 \\
\bottomrule
\end{tabular}
}
\end{table*}

\subsection{Reinforcement Learning with Human Feedback}\label{Section 4.1}
Supervised Fine-Tuning (SFT)~\citep{Ouyang2022TrainingLM} has served as a foundational technique for guiding LLMs to follow human instructions. Nevertheless, the diversity and quality of annotated data in purely supervised scenarios can be uneven, and the ability of supervised models to capture more nuanced or adaptive human preferences is often limited. In response, reinforcement learning (RL)-based fine-tuning has been proposed to address these shortcomings. Among RL methods, Reinforcement Learning from Human Feedback (RLHF)~\citep{bai2022training} stands out as one of the earliest and most impactful RL-based post-training approaches for alignment.

\begin{figure}[t]
    \centering
    \includegraphics[width=1\linewidth]{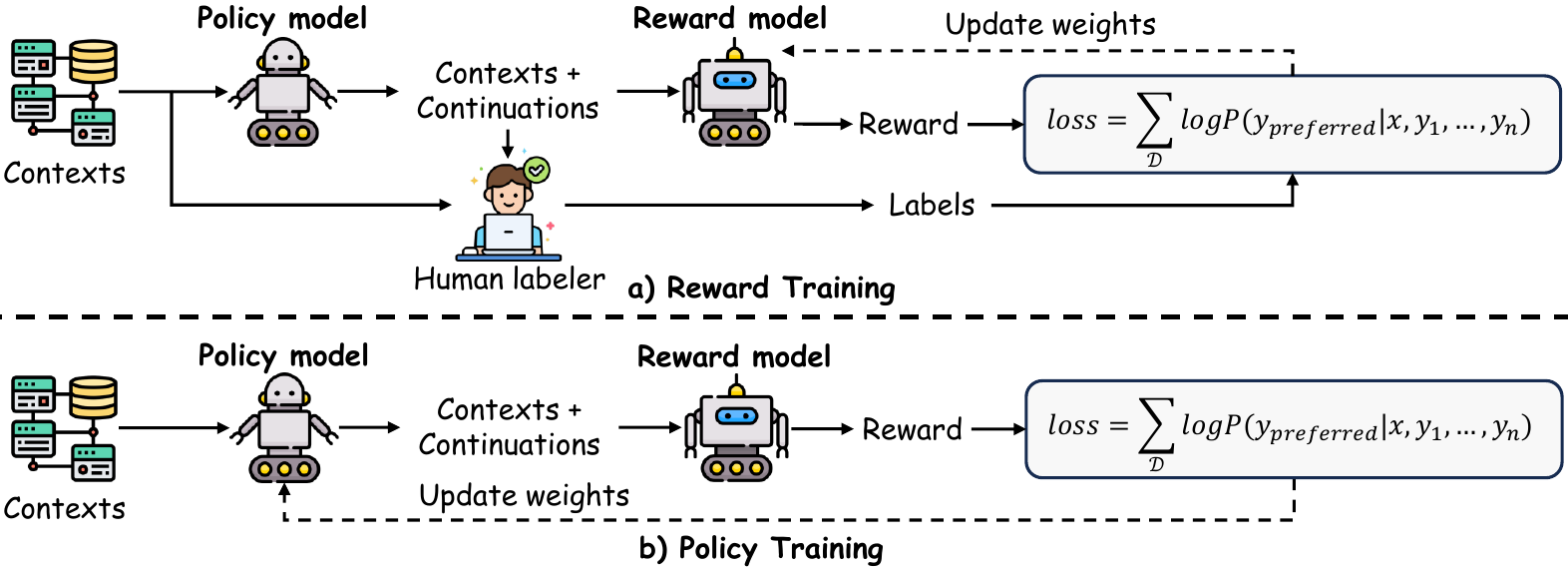}
    \caption{Workflow of Reinforcement Learning from Human Feedback (RLHF), delineating the overall training process for aligning Large Language Models with human preferences.}
    \label{fig:RLHF}
\end{figure}

As illustrated in \textbf{Fig.~\ref{fig:RLHF}}, RLHF first aggregates human feedback in the form of preference labels or reward signals, then uses that information to train a reward model. Guided by this reward model, the policy is iteratively adjusted to better match human preferences. Compared with SFT, RLHF incorporates continuous, preference-driven updates, leading to stronger alignment outcomes. Notably, modern LLMs such as GPT-4~\citep{achiam2023gpt}, Claude~\citep{TheC3}, and Gemini~\citep{team2023gemini} have benefited from these mechanisms, showcasing improvements in instruction-following, factual consistency, and user relevance. Below, we discuss the major components of RLHF, including feedback mechanisms, reward modeling, and policy-learning strategies.

\subsubsection{Feedback Mechanisms of RLHF}\label{Section 4.2.1}
Human feedback lies at the core of RLHF, informing the reward model about user preferences and guiding policy updates. This subsection adopts the taxonomy of~\citep{kaufmann2023survey} to categorize common forms of human feedback. \textbf{Table~\ref{tbl:feedback-classes-features}} presents these feedback types across dimensions such as granularity, involvement level, and explicitness. Each feedback modality contributes to distinct aspects of model optimization, offering varying levels of interpretability, scalability, and noise tolerance.

\begin{table*}[h]
    \centering
    \caption{Classification of Feedback Types in Post-training Methods for Large Language Models. This table provides an overview of common feedback classes and their defining attributes across six metrics: \textbf{Granularity} (scope: episode, segment, or step), \textbf{Involvement} (engagement: observed, active, or co-generative), \textbf{Arity} (instance count: single, multiple, or ternary), \textbf{Abstraction} (target: feature or instance), \textbf{Intent} (purpose: evaluative, descriptive, or literal), and \textbf{Explicitness} (directness: explicit or implicit).} 
    \label{tbl:feedback-classes-features}
    \renewcommand{\arraystretch}{1.1}
    \resizebox{\textwidth}{!}{
	\begin{tabular}{cccccccc}
	\toprule
    \textbf{Feedback} & \textbf{Method} & \textbf{Granularity} & \textbf{Involvement} & \textbf{Arity} & \textbf{Abstraction} & \textbf{Intent} & \textbf{Explicitness}\\
    \midrule
    \rowcolor{Gray}&\textbf{Critique}~\citep{xiao2020fresh} & - & Observed & Single  & - & Evaluative    & Explicit  \\
    \rowcolor{LightGray}&\textbf{Comparisons}~\citep{akrour2011preference}   & - & Observed & Multiple& - & Evaluative    & Explicit \\
    \rowcolor{Gray}&\textbf{Inter-Temporal}~\citep{cabi2019scaling}& Segment& Observed & Single  & - & Evaluative    & Explicit \\
    \rowcolor{LightGray}&\textbf{Proxy Rewards}~\citep{he2021assisted} & Episode& Observed & - & Feature& Descriptive   & Explicit \\
    \rowcolor{Gray}&\textbf{Social Behavior}~\citep{cui2021empathic}    & Segment& Observed & Single  & Instance    & Literal  & Implicit \\
    \rowcolor{LightGray}&\textbf{Improvements}~\citep{luo2023rlif}  & Episode& Co-generative & Single  & Instance    & -   & - \\
    \rowcolor{Gray}&\textbf{Natural Language}~\citep{ma2023eureka}   & - & Observed & Single  & - & Descriptive   & Explicit \\
    \multirow{-9}{*}{\begin{tabular}[c]{@{}c@{}}\textbf{Primary}\end{tabular}}   \\
    \noalign{\vskip-1.20em}
    \midrule
    \rowcolor{LightGray}&\textbf{E-Stops}~\citep{ghosal2023effect} & Episode& Observed & Single  & Instance    & Literal  & Implicit \\
    \rowcolor{Gray}&\textbf{Importance}~\citep{guan2021widening}    & - & Observed & - & - & Descriptive   & Explicit \\
    \multirow{-4}{*}{\begin{tabular}[c]{@{}c@{}}\textbf{Representation}\end{tabular}}   \\
    \noalign{\vskip-1.20em}
    \midrule
    \rowcolor{LightGray} &\textbf{Feature Traces}~\citep{bobu2022inducing}& Segment& Active   & Single  & Instance    & Descriptive   & Explicit \\
    \rowcolor{Gray}&\textbf{Similarity Queries}~\citep{bobu2023sirl} & - & Observed & Ternary& - & Descriptive   & Explicit \\
    \multirow{-4}{*}{\begin{tabular}[c]{@{}c@{}}\textbf{Supplementary}\end{tabular}}   \\
    \noalign{\vskip-1.20em}
    
    \bottomrule
    \end{tabular}
    }
\end{table*}

\noindent\textbf{Primary Feedback.}~~ This category comprises feedback types that most directly shape reward models in RLHF. For example, Critique~\citep{xiao2020fresh} focuses on explicit human assessments of agent behavior, often refined via binary or multi-label annotations to mitigate noise. Comparisons~\citep{akrour2011preference} allow evaluators to compare multiple outputs or trajectories; while larger choice sets can offer richer signals, they may also lead to causal confusion. Inter-Temporal Feedback~\citep{cabi2019scaling} refines trajectory assessment by providing judgments at different time steps, whereas Proxy Rewards~\citep{he2021assisted} incorporate approximate reward functions that direct the model toward a user-defined goal. Social Behavior~\citep{cui2021empathic} harnesses implicit cues (e.g., facial expressions) to align agent objectives with user sentiment. Improvements~\citep{luo2023rlif} emphasize real-time human interventions for incremental policy refinement. Finally, Natural Language Feedback~\citep{ma2023eureka} leverages textual information to convey preferences and suggestions for improvement.

\noindent\textbf{Supplementary Feedback.}~~In addition to primary feedback, two classes further strengthen the reward modeling process. Emergency stops (e-stops)~\citep{ghosal2023effect} allow humans to intervene in an agent's behavior by halting its trajectory without suggesting alternatives. This feedback is characterized by implicit involvement and a singular focus on preventing undesirable behavior. In contrast, importance labels~\citep{guan2021widening} indicate the significance of specific observations for achieving objectives, providing explicit feedback that does not directly alter behavior. This feedback varies by context and serves as supplementary input, reinforcing the overall learning process for the reward model. 

\noindent\textbf{Representation-Specific Feedback.}~~Certain feedback types primarily enhance representation learning rather than directly shaping the reward function. Feature Traces~\citep{bobu2022inducing} prompt human operators to demonstrate monotonic changes in a given feature, thus enabling dynamic expansion of feature sets. Similarity Queries~\citep{bobu2023sirl} compare triplets of trajectories, guiding representation learning via pairwise distances in trajectory space. By leveraging these representation-specific feedback forms, RLHF can achieve more robust generalization to new tasks and contexts.

\subsubsection{Reward Model of RLHF}\label{Section 4.2.2}
The true reward function \(r(x, y)\) is often unknown, making it necessary to construct a learnable reward model \(r_{\theta}(x, y)\) based on human-provided preferences. This model predicts the degree to which a candidate output \(y\) aligns with human expectations for a given input \(x\). To obtain the training data for \(r_{\theta}(x, y)\), human evaluators compare or label output pairs according to their relative suitability, and the model is typically trained using a cross-entropy loss on these comparisons. To discourage the policy \(\pi\) from straying too far from the initial model \(\rho\), a penalty term controlled by the hyperparameter \(\beta\) is introduced into the reward function:
\begin{equation}
    r_{\theta}(x, y)=r(x, y)-\beta \log \frac{\pi(y \mid x)}{\rho(y \mid x)},
\end{equation}
where \(\pi(y \mid x)\) is the probability that the fine-tuned policy \(\pi\) produces output \(y\) given input \(x\), and \(\rho(y \mid x)\) is the corresponding probability under the original model \(\rho\). This term ensures that, while \(\pi\) adapts to human feedback, it remains constrained by prior knowledge captured in \(\rho\).

Evaluating the reward function \(r_{\theta}(x, y)\) is critical, as it directly influences learning effectiveness and policy performance. Accurately assessing this function helps identify suitable reward structures for aligning model outputs with human preferences. However, in safety-sensitive domains, standard rollout methods~\citep{fu2017learning,brown2019extrapolating} and off-policy evaluations~\citep{le2019batch,irpan2019off} may be infeasible because of risks related to online interactions, biases, and the need for ground-truth rewards. To address these challenges, two prominent approaches are commonly adopted:

\noindent\textbf{Distance Functions.}~~Recent research has focused on reward evaluation distance functions that account for potential transformations, such as potential shaping. For example, EPIC~\citep{gleave2020quantifying} measures reward function equivalences under various transformations, while DARD~\citep{wulfe2022dynamics} refines canonicalization to ensure that evaluations remain grounded in feasible transitions. EPIC-like distances~\citep{jenner2022general} generalize EPIC’s methodology by allowing variability in canonicalization, normalization, and metric functions, and STARC~\citep{skalse2023starc} preserves EPIC’s theoretical properties while offering additional flexibility.

\noindent\textbf{Visual and Human Inspection.}~~Other methods rely on interpretability and curated datasets to gauge the validity of learned reward functions. PRFI~\citep{jenner2022preprocessing} uses a preprocessing step to simplify reward functions while retaining equivalence, thus enhancing their transparency. Meanwhile, CONVEXDA and REWARDFUSION~\citep{shen2023trickle} propose datasets designed to test how consistently reward models respond to semantic variations in prompts. Together, these techniques contribute to more reliable evaluations of reward functions, reinforcing the alignment of Large Language Models with human preferences.

\subsubsection{Policy Learning of RLHF}\label{Section 4.2.3}
\begin{figure}[h]
    \centering
    \includegraphics[width=0.8\linewidth]{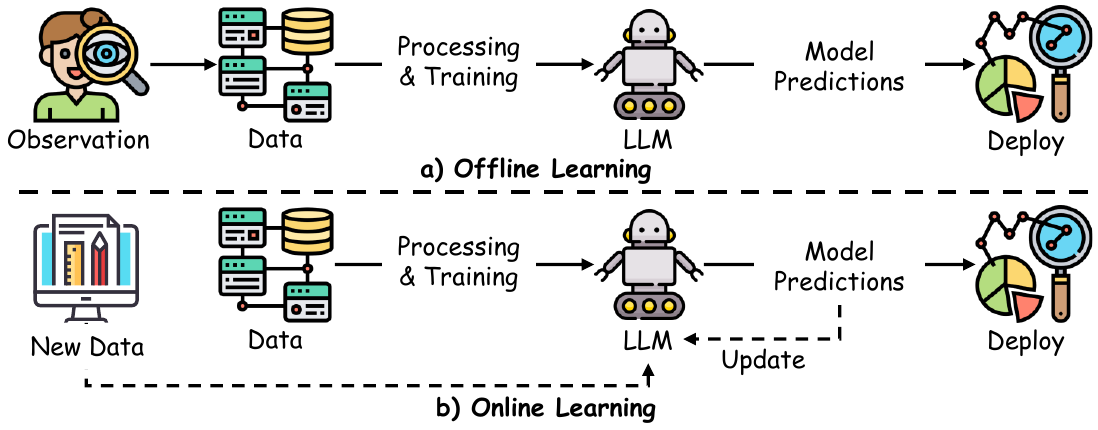}
    \caption{Comparison of Online and Offline RLHF, illustrating continuous feedback collection during policy execution in online RLHF versus pre-collected trajectory utilization in offline RLHF.}
    \label{fig:online and offline}
\end{figure}

Policy learning for RLHF, as shown in \textbf{Fig. \ref{fig:online and offline}}, involves optimizing policies through human feedback in both online and offline settings.

\noindent\textbf{Online Learning.}~~In online RLHF, systems gather real-time human preferences on newly generated model trajectories. Algorithms like DPS~\citep{novoseller2020dueling} use Bayesian updates to manage the dueling process, while PPS and PEPS~\citep{xu2020preference} integrate dynamic programming and bandit ideas to refine policy behavior. In LPbRL~\citep{saha2023dueling}, feature embeddings capture evolving reward structures, and PbOP~\citep{chen2022human} integrates least-squares estimates for both transition dynamics and preference signals. More recently, PARL~\citep{chakraborty2023parl} targets data-collection efficiency by treating feedback acquisition as an integral part of policy optimization.

\noindent\textbf{Offline Learning.}~~In offline RLHF, previously gathered preference-labeled trajectories are used to learn or refine a policy. For instance, \citep{zhu2023principled} study pessimistic maximum likelihood estimation for policy learning with pairwise comparison data, establishing bounds on performance. Extensions like FREEHAND~\citep{zhan2023provable} and DCPPO~\citep{li2023reinforcement} generalize to unknown preference models, exploring the interplay between offline data coverage and policy generalization. Moreover, \citep{zhu2024iterative} address overfitting in Boltzmann models for pairwise comparisons, while DCPPO~\citep{li2023reinforcement} further studies the dynamic discrete choice model for improved feedback efficiency.

\noindent\textbf{Blending Online and Offline Learning.}~~Hybrid methods combine offline pretraining with online preference aggregation, capitalizing on pre-collected data while still incorporating real-time updates. PFERL~\citep{kong2022provably} adopts a two-phase approach to minimize human queries, whereas PERL~\citep{jin2020provably} explores optimistic least-squares strategies for active exploration. Dueling RL~\citep{saha2023dueling} and its extensions (e.g., REGIME in PRPRL~\citep{zhan2023provable}) reduce human labeling requirements by carefully partitioning data acquisition from feedback collection, thus optimizing trade-offs among sample efficiency, annotation cost, and policy performance.

\subsection{Reinforcement Learning with AI Feedback}\label{Section 4.2}
Reinforcement Learning with AI Feedback (RLAIF) extends the RLHF paradigm by employing LLMs to generate feedback signals. This approach can complement or replace human feedback, providing more scalable, lower-cost preference data in tasks where human annotations are scarce, costly, or inconsistent.

\subsubsection{RLAIF vs RLHF}
A major challenge in applying RLHF at scale lies in its reliance on human-generated preference labels, which necessitate considerable resources for gathering, curating, and labeling data. The process of annotating data is both time-intensive and costly, and human evaluators may introduce inconsistencies, thereby complicating large-scale, consistent labeling across all model outputs. These constraints significantly limit the scalability and efficiency of RLHF.
To address these challenges, RLAIF was proposed by \citep{bai2022constitutional}, which combines human feedback with AI-generated feedback for training models via reinforcement learning. By leveraging LLMs as the source of feedback, RLAIF reduces reliance on human annotators, offering a viable alternative to traditional RLHF. This approach enables continuous feedback generation, significantly enhancing scalability while preserving the flexibility of human-guided model optimization.

\begin{figure}[h]
    \centering
    \includegraphics[width=0.95\linewidth]{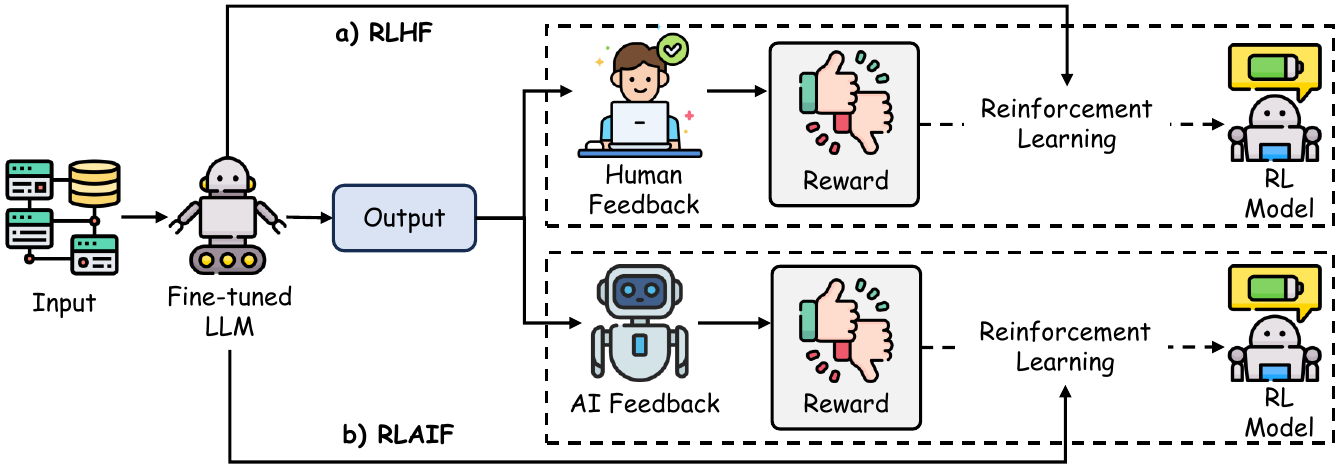}
    \caption{Comparison of RLHF and RLAIF Approaches, delineating their distinct methodologies for preference alignment in Large Language Models.}
    \label{fig:RLHF vs RLAIF}
\end{figure}

As depicted in \textbf{Fig.~\ref{fig:RLHF vs RLAIF}}, the key distinction between RLHF and RLAIF lies in the source of feedback: RLHF relies on human-generated preferences, while RLAIF uses AI-generated feedback to guide policy updates. Empirical studies, such as those by \citep{leerlaif}, have demonstrated that RLAIF can achieve performance comparable to or even superior to RLHF, as evaluated by human raters. Notably, RLAIF not only surpasses traditional supervised fine-tuning baselines but does so with an LLM preference labeler of the same scale as the policy model, underscoring the approach's efficiency.

\subsubsection{RLAIF Training Pipeline}
The RLAIF training pipeline follows several key stages wherein AI-generated feedback is utilized to iteratively refine the model's behavior. The pipeline facilitates the alignment of LLM outputs with human expectations in a manner that scales across various tasks, as detailed by \citep{lee2023rlaif}. The stages are as follows:

\noindent\textbf{AI Feedback Collection.}~~In this phase, the AI system generates feedback based on predefined criteria, which may include task-specific metrics, correctness of responses, or appropriateness of the model’s outputs. Unlike human feedback, which requires interpretation and manual annotation, AI feedback can be consistently generated across a broad range of model outputs. This characteristic enables AI feedback to be continuously provided, scaling the feedback loop significantly.

\noindent\textbf{Reward Model Training.}~~The AI-generated feedback is subsequently used to train or refine a reward model. This model maps input-output pairs to corresponding rewards, aligning the model's output with the desired outcomes as dictated by the feedback. While traditional RLHF relies on direct human feedback to evaluate outputs, RLAIF utilizes AI-generated labels, which, although potentially introducing issues related to consistency and bias, offer advantages in scalability and independence from human resources.

\noindent\textbf{Policy Update.}~~The final stage involves updating the model’s policy based on the reward model trained in the previous step. Reinforcement learning algorithms are employed to adjust the model's parameters, optimizing the policy to maximize cumulative reward across a variety of tasks. This process is iterative, with the reward model guiding the model’s outputs towards higher alignment with the intended objectives.

The principal advantage of RLAIF lies in its ability to scale the feedback loop without requiring continual human intervention. By substituting human feedback with AI-generated feedback, RLAIF facilitates the continuous improvement of LLMs across multiple tasks, alleviating the bottleneck posed by human labeling efforts.

\subsection{Direct Preference Optimization}\label{Section 4.3}
As previously discussed, RLHF~\citep{Ouyang2022TrainingLM} typically consists of three stages: Supervised Fine-Tuning~\citep{wei2021finetuned, Wang2022SelfInstructAL}, Reward Modeling, and Reinforcement Learning (usually implemented via Proximal Policy Optimization, PPO)~\citep{Schulman2017ProximalPO}. Despite its effectiveness, RLHF can be complex and prone to instability, particularly in the stages where a reward model is fitted and then used to fine-tune a large language model. The difficulty lies in creating a reward model that accurately reflects human preferences and in the challenge of fine-tuning the language model to optimize this estimated reward while staying close to the original model.
To address these issues, Direct Preference Optimization (DPO)~\citep{rafailov2024directpreferenceoptimizationlanguage} has been introduced as a more stable and computationally efficient alternative. DPO simplifies the reward optimization process by directly linking the reward function to the optimal policy. It treats the reward maximization problem as a single-stage policy training problem based on human preference data, thus avoiding the complexities of reward model fitting and the dependencies of the Bradley-Terry model~\citep{bt}.

\subsubsection{Foundation of DPO}
\label{Foundation of DPO}
RLHF involves training a reward model (RM) and fine-tuning a language model (LM) via reinforcement learning. DPO simplifies this process by training the LM directly with human preference data, implicitly capturing the reward model within the policy itself.

\noindent\textbf{KL-Regularized Reward Maximization Objective.}  
DPO begins with the well-established KL-regularized reward maximization framework, as shown in the following objective:
\begin{equation}
\label{objective}
    \pi^{*} 
    \;=\; 
    \arg\max_{\pi}
    \;
    \mathbb{E}_{x \,\sim\, \mathcal{D},\, y \,\sim\, \pi(\cdot \mid x)} 
    \Bigl[
        r(x, y) 
        \;-\; 
        \beta \,\mathrm{KL}\!\Bigl(\pi(\cdot \mid x)\,\bigl\|\,\pi_{\mathrm{ref}}(\cdot \mid x)\Bigr)
    \Bigr],
\end{equation}
where \(r(x, y)\) represents the reward function, \(\beta > 0\) is a coefficient controlling the degree of proximity to the reference policy \(\pi_{\mathrm{ref}}\), and \(\mathrm{KL}(\cdot\|\cdot)\) denotes the Kullback-Leibler divergence. Here, \(x \sim \mathcal{D}\) represents the input drawn from the data distribution, and \(y \sim \pi(\cdot \mid x)\) denotes the output sampled from the policy.

\noindent\textbf{Deriving the Optimal Policy.}  
Under appropriate assumptions, the solution to Eq.~\eqref{objective} is derived in the form of a Boltzmann distribution~\citep{peng2019advantageweightedregressionsimplescalable,go2023aligninglanguagemodelspreferences,jaques2020humancentricdialogtrainingoffline}:

\begin{equation}
\label{optimal_obj}
    \pi^{*}(y \mid x)
    \;=\;
    \frac{1}{Z(x)}\,
    \pi_{\mathrm{ref}}(y \mid x)
    \exp\!\Bigl(\tfrac{1}{\beta}\,r(x, y)\Bigr),
\end{equation}
where the partition function
\begin{equation}
    Z(x) 
    \;=\; 
    \sum_{y}\,
    \pi_{\mathrm{ref}}(y \mid x)
    \exp\!\Bigl(\tfrac{1}{\beta}\,r(x, y)\Bigr)
\end{equation}
acts as a normalization term ensuring that \(\pi^{*}\) remains a valid probability distribution (i.e., that its probabilities sum to 1).

\noindent\textbf{Reparameterizing the Reward.}  
Taking the natural logarithm of both sides of Eq.~\eqref{optimal_obj}, we can relate the reward \(r(x, y)\) to the optimal policy \(\pi^{*}\). This yields:

\begin{equation}
\label{reparameterization}
    r^{*}(x, y)
    \;=\;
    \beta
    \Bigl[
        \log \pi^{*}(y \mid x)
        \;-\;
        \log \pi_{\mathrm{ref}}(y \mid x)
    \Bigr]
    \;+\;
    \beta \,\log Z(x),
\end{equation}
where \(\beta \log Z(x)\) is a constant that does not affect pairwise comparisons of rewards. If the optimal policy \(\pi^{*}\) is known, the true reward \(r^{*}(x, y)\) can be determined up to this constant.

\noindent\textbf{Bradley–Terry Preferences.}  
Under the Bradley-Terry model~\citep{bt}, human preferences between two outputs \(y_1\) and \(y_2\) are governed by the difference in their reward values. The probability of preferring \(y_1\) over \(y_2\) is given by
\begin{equation}
\label{reward}
    p^{*}\bigl(y_{1} \succ y_{2} \mid x\bigr)
    \;=\;
    \frac{
        \exp\!\bigl(r^{*}(x, y_{1})\bigr)
    }{
        \exp\!\bigl(r^{*}(x, y_{1})\bigr)
        \;+\;
        \exp\!\bigl(r^{*}(x, y_{2})\bigr)
    }.
\end{equation}
Substituting Eq.~\eqref{reparameterization} into Eq.~\eqref{reward}, we obtain the final preference model:
\begin{equation}
\label{final_preference_model}
    p^{*}\bigl(y_{1} \succ y_{2} \mid x\bigr)
    \;=\;
    \frac{
        1
    }{
        1
        \;+\;
        \exp\!\Bigl(
            \beta
            \Bigl[
                \log \tfrac{\pi^{*}(y_{2} \mid x)}{\pi_{\mathrm{ref}}(y_{2} \mid x)}
                \;-\;
                \log \tfrac{\pi^{*}(y_{1} \mid x)}{\pi_{\mathrm{ref}}(y_{1} \mid x)}
            \Bigr]
        \Bigr)
    }.
\end{equation}

This expression links the pairwise human preference probability to the ratio of the optimal policy \(\pi^{*}\) and reference policy \(\pi_{\mathrm{ref}}\).

\noindent\textbf{Objective of DPO.}~~DPO sidesteps explicit reward modeling by learning a policy directly from preference data. Given a dataset of preference triplets \(\{(x, y_{w}, y_{l})\}\), where \(y_{w}\) is the preferred output and \(y_{l}\) is the less preferred output for a prompt \(x\), DPO maximizes the likelihood of observed preferences. Formally, DPO adopts the following objective:

\begin{equation}
    \mathcal{L}_{\mathrm{DPO}}\bigl(\pi_{\theta}; \pi_{\mathrm{ref}}\bigr)
    \;=\;
    -\,
    \mathbb{E}_{(x,\,y_{w},\,y_{l}) \,\sim\, \mathcal{D}}
    \biggl[
        \log \sigma\!\Bigl(
            \beta
            \bigl[\log \tfrac{\pi_{\theta}(y_{w} \mid x)}{\pi_{\mathrm{ref}}(y_{w} \mid x)}\bigr]
            \;-\;
            \beta
            \bigl[\log \tfrac{\pi_{\theta}(y_{l} \mid x)}{\pi_{\mathrm{ref}}(y_{l} \mid x)}\bigr]
        \Bigr)
    \biggr],
\end{equation}
where \(\sigma(\cdot)\) is the logistic sigmoid function, and \(\beta \log \tfrac{\pi_{\theta}(y \mid x)}{\pi_{\mathrm{ref}}(y \mid x)}\) represents a reparameterized reward difference between \(\pi_{\theta}\) and the reference policy \(\pi_{\mathrm{ref}}\). By maximizing \(\mathcal{L}_{\mathrm{DPO}}\), the policy \(\pi_{\theta}\) aligns with human preferences without requiring a separate reward model. 
Because the DPO objective inherits a KL-regularized formulation from RLHF, it preserves essential theoretical guarantees—such as consistency under well-defined preference assumptions~\citep{bong2022generalizedresultsexistenceconsistency}—while unifying the training procedure into a single stage. Consequently, DPO facilitates a more direct path for aligning language models with human evaluations, reducing system complexity and enhancing training stability.

\subsubsection{Training Details of DPO}
\label{Training Details of DPO}
\begin{figure}[h]
    \centering
    \includegraphics[width=0.95\linewidth]{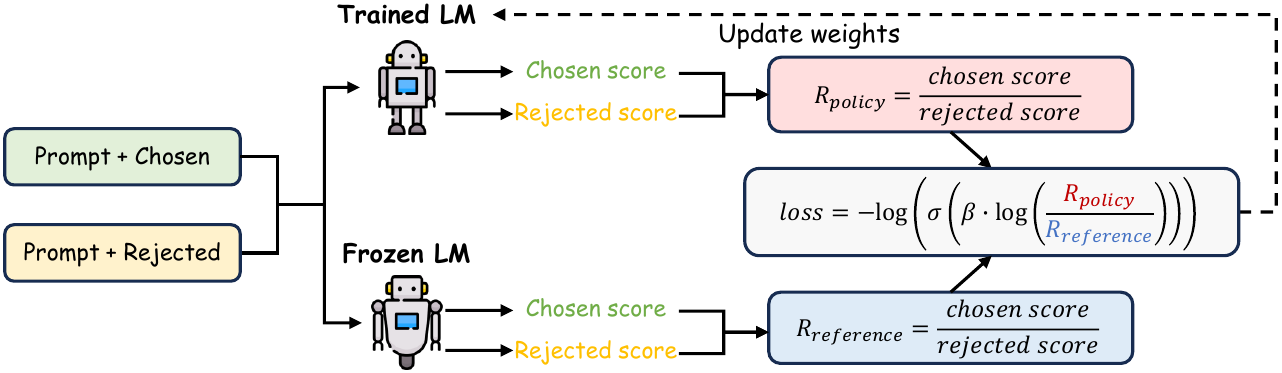}
    \caption{Workflow of Direct Preference Optimization (DPO), illustrating the training pipeline for optimizing Large Language Model outputs based on human preferences.}
    \label{fig:DPO}
\end{figure}

The DPO framework builds upon two core models: a reference policy \(\pi_{\mathrm{ref}}\) and a target policy \(\pi_{\mathrm{tar}}\). The reference policy, typically a pre-trained and supervised fine-tuned language model, remains fixed throughout training. By contrast, the target policy is initialized from \(\pi_{\mathrm{ref}}\) and iteratively updated using preference-based feedback, thereby improving alignment with human judgments. \textbf{Fig. ~\ref{fig:DPO}} depicts this overall pipeline.

\noindent\textbf{Data Collection and Preparation.}~~DPO relies on a curated preference dataset obtained by sampling multiple candidate responses from \(\pi_{\mathrm{ref}}\) for each prompt \(x\). Human annotators then compare or rank these responses based on coherence, relevance, and clarity, among other criteria. The resulting preference labels serve as the core training signals for optimizing \(\pi_{\mathrm{tar}}\).

\noindent\textbf{Training Procedure.}~~The target policy is refined through a series of gradient-based updates aimed at minimizing the loss \(L_{\mathrm{DPO}}\). Specifically, 
1) Generation: \(\pi_{\mathrm{ref}}\) produces candidate outputs for each prompt \(x\).  
2) Annotation: human annotators compare the generated outputs, determining their relative preference.  
3) Optimization: using these pairwise preferences, \(\pi_{\mathrm{tar}}\) is iteratively updated to better emulate human-favored outputs.  
Throughout this process, \(\pi_{\mathrm{ref}}\) remains unchanged, providing a stable baseline against which to measure improvements.

\noindent\textbf{Practical Considerations.}~~Selecting a robust reference policy is often critical to initializing DPO effectively. SFT typically yields a well-performing baseline for \(\pi_{\mathrm{ref}}\), ensuring that subsequent preference-driven updates can focus on refinement rather than fundamental skill acquisition. Additionally, preference data must be sufficiently diverse to capture variations in user expectations, thereby promoting model adaptability and preventing overfitting to narrowly defined tasks.

\subsubsection{Variants of DPO}\label{Variants of DPO}
Multiple variants of DPO have emerged to address specific alignment challenges and optimize different aspects of text generation. \textbf{Table~\ref{tab:alignment_comparison}} contains an overview of these methods, which range from token-level generation optimizations to controlling verbosity and handling listwise or negative preferences.

\noindent{\textbf{DPO for Optimizing Generation.}}~~Token-level and iterative DPO strategies facilitate finer-grained or continuous alignment with human preferences.
Reformulated as a bandit problem, token-level DPO~\citep{rafailov2024r} adopts a Markov Decision Process (MDP) defined by \((S, A, f, r, \rho_0)\). This approach mitigates challenges such as excessive KL divergence for dispreferred tokens. TDPO~\citep{zeng2024token} applies sequential forward KL divergence instead of reverse KL, improving both alignment and diversity preservation in text generation.
Iterative DPO~\citep{yuan2024self} adopts a multi-round approach to continuously refine outputs through repeated preference evaluations, often performed by the model itself. Pairwise Cringe Optimization (PCO)~\citep{xu2024thingscringeothersiterative} extends binary feedback to a pairwise setting, using a soft margin to balance exploration and exploitation. Step-wise DPO~\citep{kim2024sdpodontusedata} partitions the preference dataset and applies iterative updates, using the updated policy from each round as the baseline for the next.

\noindent{\textbf{Controllable and Flexible DPO.}}~~Some DPO variants aim to manage verbosity and reduce the need for a fixed reference policy. R-DPO~\citep{park2024disentangling} penalizes output length through a regularization term in the objective function, addressing overly verbose or redundant responses. SimPO~\citep{meng2024simpo} eliminates the requirement for a reference policy by normalizing response length and streamlining the loss function to handle both desirable and undesirable outputs.
RLOO~\citep{ahmadian2024back} leverages the REINFORCE algorithm without training a value model, substantially reducing computational overhead. It treats the entire response as a single action and learns from sparse rewards, simplifying implementation compared to traditional PPO-based methods.

\noindent{\textbf{Listwise DPO.}}~~Rather than limiting preference data to pairwise comparisons, Listwise DPO approaches optimize over sets of outputs. Listwise Preference Optimization (LiPO)~\citep{liu2024lipo} applies Learning-to-Rank techniques directly on ranked lists of candidate responses, improving efficiency relative to repeated pairwise comparisons. RRHF~\citep{yuan2023rrhf} incorporates preference alignment into SFT, eliminating the need for a separate reference model. PRO~\citep{song2024preference} breaks down listwise preferences into simpler binary tasks, simplifying alignment during SFT.

\noindent{\textbf{Negative DPO.}}~~Certain tasks require learning from undesired or harmful outputs: Negating Negatives (NN)~\citep{duan2024negating} discards positive responses and maximizes divergence from less preferred outputs.  Negative Preference Optimization (NPO)~\citep{zhang2024negative} employs gradient ascent on negative preferences, effectively reducing harmful outputs and mitigating catastrophic collapse.

\section{PoLMs for Reasoning}\label{Section 5}
Reasoning constitutes a central pillar for enabling LLMs to tackle tasks involving multi-step logic, intricate inference, and complex decision-making. This chapter examines two core techniques for enhancing model reasoning capabilities: \textbf{Self-Refine for Reasoning} (\S \ref{Section 5.1}), which guides the model to autonomously detect and remedy errors in its own reasoning steps; and \textbf{Reinforcement Learning for Reasoning} (\S \ref{Section 5.2}), which employs reward-based optimization to improve the consistency and depth of the model’s chain-of-thought. These approaches collectively enable more robust handling of long-horizon decision-making, logical proofs, mathematical reasoning, and other challenging tasks.

\subsection{Self-Refine for Reasoning}\label{Section 5.1}
\begin{figure}[h]
\centering
\includegraphics[width=1.0\linewidth]{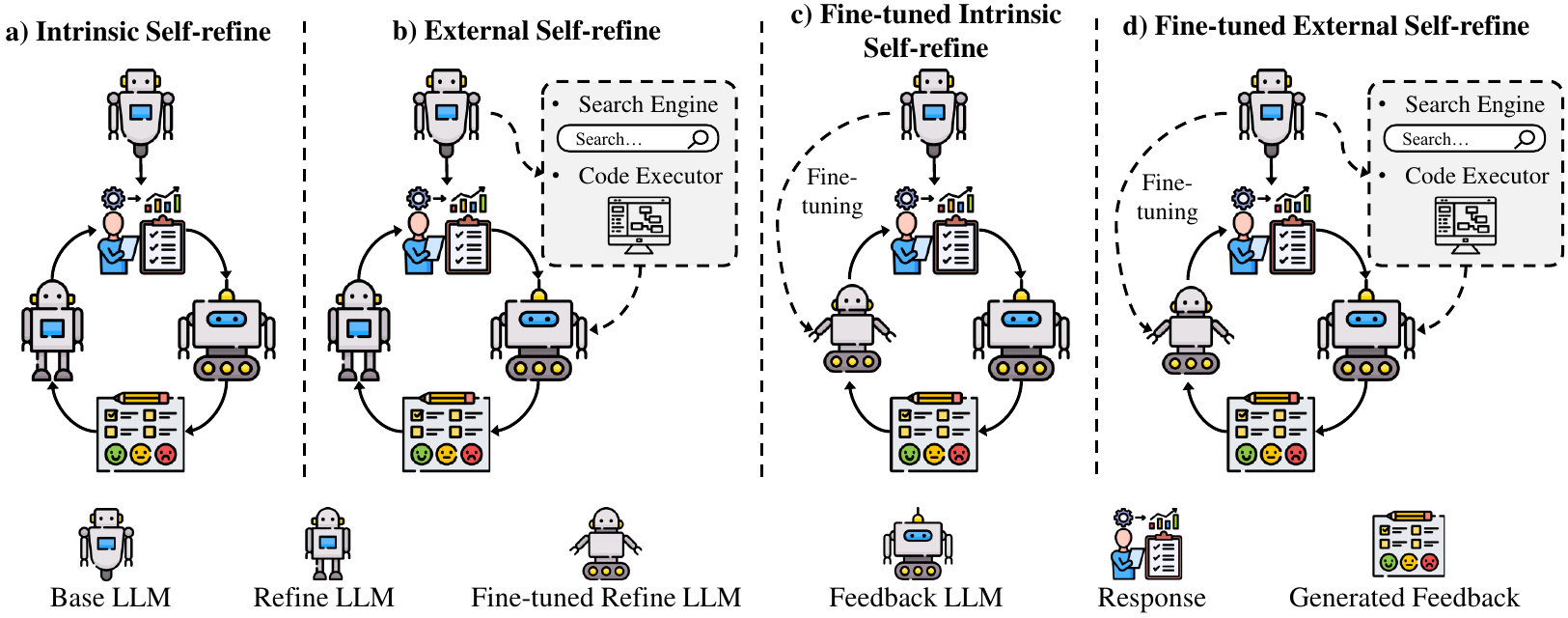}
\caption{Taxonomy of Self-Refine methods, delineating architectural variations for enhancing reasoning in Large Language Models.}
\label{fig:self-refine}
\end{figure}

Reasoning remains a core challenge in optimizing LLMs for tasks that demand intricate logical inference and context-dependent decision-making. In this context, self-refine emerges as a powerful mechanism to iteratively pinpoint and correct errors during or after text generation, substantially improving both reasoning depth and overall reliability. As shown in \textbf{Fig. \ref{fig:self-refine}}, self-refine methods can be divided into four categories: Intrinsic Self-refine, which relies on the model’s internal reasoning loops; External Self-refine, which incorporates external feedback resources; Fine-tuned Intrinsic Self-refine, which iteratively updates the model’s reasoning processes based on self-generated corrections; and Fine-tuned External Self-refine, which harnesses external signals and fine-tuning to refine reasoning in a more adaptive, long-term manner. \textbf{Table \ref{tab:self-refine}} further illustrates how each category fortifies LLM reasoning capacity across various tasks.

\begin{table*}[h]  
\centering
\caption{Overview of Self-Refine Methods in Large Language Models (2022–2025). This table summarizes prominent self-refinement techniques, detailing their primary LLMs, tasks, and release timelines across three metrics: \textbf{ET.} (External Tools: \faCheckCircle{} denotes usage, \faTimesCircle[regular]{} denotes absence), \textbf{FT.} (Fine-Tuning: \faCheckCircle{} indicates application, \faTimesCircle[regular]{} indicates non-application), and \textbf{SR.} (Self-Refine Type: IS for Intrinsic Self-refine, ES for External Self-refine, IF for Intrinsic Fine-tuning, EF for External Fine-tuning).}
\label{tab:self-refine}
\renewcommand{\arraystretch}{1.05}
\resizebox{\textwidth}{!}{
\begin{tabular}{lllcccr}  
\toprule  
\textbf{Methods} & \textbf{Main LLMs} & \textbf{Main Tasks} & \textbf{ET.} & \textbf{FT.} & \textbf{SR.} & \textbf{Release Time}\\ \midrule  
\rowcolor{Gray}\textbf{Self-Critique}~\citep{saunders2022self} & InstructGPT & Topic Summarization    & \faTimesCircle[regular] & \faCheckCircle   & IF & Jun-2022 \\  
\rowcolor{LightGray}\textbf{CodeRL}~\citep{le2022coderl} & GPT-3.5 & Program Synthesis & \faCheckCircle & \faCheckCircle  & EF & Nov-2022 \\  
\rowcolor{Gray}\textbf{CAI Revisions}~\citep{bai2022constitutional} & 52B (no details) & Detoxification      & \faTimesCircle[regular] & \faTimesCircle[regular] & IS & Dec-2022 \\  
\rowcolor{LightGray}\textbf{Baldur}~\citep{first2023baldur} & Minerva 8B, 62B & Proof Generation        & \faCheckCircle & \faCheckCircle     & EF & Mar-2023 \\  
\rowcolor{Gray}\textbf{Self-Refine}~\citep{madaan2024self} & GPT-3.5, GPT-4 & Math, Coding, Dialogue & \faTimesCircle[regular] & \faTimesCircle[regular] & IS & May-2023 \\  
\rowcolor{LightGray}\textbf{RARR}~\citep{gao2022rarr} & Palm 540B & NQ, SQA, QReCC             & \faCheckCircle                 & \faTimesCircle[regular] & ES & May-2023 \\  
\rowcolor{Gray}\textbf{SelfEvolve}~\citep{jiang2023selfevolve} & InstructGPT, GPT-4 & DS-1000, HumanEval& \faCheckCircle       & \faTimesCircle[regular] & ES & Jun-2023 \\  
\rowcolor{LightGray}\textbf{RL4F}~\citep{akyurek2023rl4f} & GPT-3 & Action Plan, Topic & \faTimesCircle[regular] & \faCheckCircle    & IF & Jul-2023 \\  
\rowcolor{Gray}\textbf{Self-Edit}~\citep{zhang2023self} & CodeGen, GPT-3.5 & Code Generation            & \faCheckCircle & \faCheckCircle     & EF & Jul-2023 \\  
\rowcolor{LightGray}\textbf{CoVe}~\citep{dhuliawala2023chain} & PaLM-540B & Multiple Answers       & \faTimesCircle[regular] & \faTimesCircle[regular] & IS & Sep-2023 \\  
\rowcolor{Gray}\textbf{FLARE}~\citep{Bharadwaj2023FLAREFL} & GPT-3.5 & StrategyQA, ASQA           & \faCheckCircle                & \faTimesCircle[regular] & ES & Oct-2023 \\  
\rowcolor{LightGray}\textbf{Logic-LM}~\citep{pan2023logic} & GPT-3.5, GPT-4 & PrOntoQA, Logic Reasoning  & \faCheckCircle    & \faTimesCircle[regular] & ES & Oct-2023 \\  
\rowcolor{Gray}\textbf{Reflexion}~\citep{shinn2024reflexion} & GPT-4 & Games, Coding, HotpotQA    &\faCheckCircle & \faTimesCircle[regular] & ES & Oct-2023 \\  
\rowcolor{LightGray}\textbf{Self-Debug}~\citep{chen2023teaching} & GPT-3.5, GPT-4 & Text-to-Code               & \faCheckCircle           & \faTimesCircle[regular] & ES & Oct-2023 \\  
\rowcolor{Gray}\textbf{SelFee}~\citep{ye2023selfee} & LLaMA-7B, 13B & MT-Bench               & \faTimesCircle[regular] & \faCheckCircle & IF & 2023 \\  
\rowcolor{LightGray}\textbf{RCI}~\citep{Yang2024RandomCI} & GPT-3.5-Turbo & Computer Taks, CSQA    & \faTimesCircle[regular] & \faTimesCircle[regular] & IS & 2023 \\  
\rowcolor{Gray}\textbf{REFINER}~\citep{paul2023refiner} & GPT-3.5 & Math, Logic, Moral Stories & \faTimesCircle[regular] & \faCheckCircle  & IF & Feb-2024 \\  
\rowcolor{LightGray}\textbf{CRITIC}~\citep{gou2023critic} & GPT-3, LLaMA2-70B & GSM8k, SVAMP, HotpotQA     & \faCheckCircle    & \faTimesCircle[regular] & ES & Feb-2024 \\
\rowcolor{Gray}\textbf{ProMiSe}~\citep{ramji2024self} & FLAN-T5, LLaMA2-13B & MultiDoc2Dia, QuAC  & \faCheckCircle & \faTimesCircle[regular]   & ES & Feb-2024 \\
\rowcolor{LightGray}\textbf{PREFER}~\citep{zhang2024prefer} & InstructGPT, GPT-4 & NLI, NLC  & \faTimesCircle[regular] & \faTimesCircle[regular]   & IS & Mar-2024 \\  
\rowcolor{Gray}\textbf{Volcano}~\citep{lee2023volcano} & GPT-3.5 & Visual Reasoning       & \faTimesCircle[regular] & \faCheckCircle  & IF & Apr-2024 \\  
\rowcolor{LightGray}\textbf{CYCLE}~\citep{ding2024cycle} & CodeGen, StarCoder & HumanEval, MBPP-S, APPS  & \faTimesCircle[regular] & \faCheckCircle   & IF & Apr-2024 \\ 
\rowcolor{Gray}\textbf{SRIT}~\citep{ranaldi2024self} & LLaMA2-7B, GPT-3.5 & SIQA, PIQA, CSQA, OBQA   & \faTimesCircle[regular] & \faTimesCircle[regular]   & IS & May-2024 \\
\rowcolor{LightGray}\textbf{MCTSr}~\citep{zhang2024accessing} & LLaMA3-8B, GPT-4 & GSM8K, GSM Hard, MATH  & \faTimesCircle[regular] & \faTimesCircle[regular]   & IS & Jun-2024 \\  
\rowcolor{Gray}\textbf{Self-Contrast}~\citep{zhang2024self} & GPT-3.5, L-70B & GSM8K, SVAMP, CommonMT  & \faTimesCircle[regular] & \faTimesCircle[regular]   & IS & Jun-2024 \\
\rowcolor{LightGray}\textbf{TEaR}~\citep{feng2024improving} & GPT-3.5, Claude-2 & WMT22, WMT23  & \faTimesCircle[regular] & \faTimesCircle[regular]   & IS & Jun 2024 \\
\rowcolor{Gray}\textbf{LLMRefine}~\citep{xu2024llmrefine} &  PaLM (Bison) & MQM, ASQA, Summ  & \faTimesCircle[regular] & \faCheckCircle   & IF & Jun-2024 \\
\rowcolor{LightGray}\textbf{Self-Bias}~\citep{xu2024pride} & GPT-4, DeepSeek & Flores-200, MQM & \faCheckCircle & \faTimesCircle[regular]   & ES & Aug-2024\\
\rowcolor{Gray}\textbf{Exp-Refiner}~\citep{quan2024verification} & GPT-3.5, GPT-4 & e-SNLI, QASC, WorldTree  & \faTimesCircle[regular] & \faTimesCircle[regular] & IS & Oct-2024 \\
\rowcolor{LightGray}\textbf{Self-corrective}~\citep{Abdelaal2025GLLMSG} & GPT-2 & GSM8k, SVAMP, Detoxification  & \faTimesCircle[regular] & \faCheckCircle   & IF & Jan-2025 \\
\bottomrule  
\end{tabular}
}
\end{table*}

\noindent\textbf{Intrinsic Self-Refine.}~~Intrinsic self-refine methods focus on empowering the model itself to detect and fix errors internally without resorting to outside tools. For instance, RCI Prompting~\citep{kim2024language} only triggers corrections when a contradiction or error is identified, avoiding overreactions to minor uncertainties. CAI Revisions~\citep{bai2022constitutional} corrects undesirable outputs (e.g., offensive text) while teaching the model to self-moderate its responses. Similarly, Self-Refine~\citep{madaan2024self} leverages the transition from lower-quality prompts to high-fidelity instructions, refining the intermediate logic to boost consistency. CoVe~\citep{dhuliawala2023chain} addresses multi-answer questions by dividing them into subtasks, each verified individually to ensure precision and consistency across the entire reasoning chain.
Weak-to-Strong Generalization (W2SG) approaches leverage advanced algorithms to enable strong student models to learn effectively from noisy demonstrations produced by less capable teacher models~\citep{burns2023weak}. 
This framework has seen several key developments and applications across different domains. Recent research has enhanced W2SG through various innovations.  
For instance, ensemble learning techniques have been successfully applied to improve the robustness and effectiveness of W2SG methods~\citep{sang2024improving}. 
\citep{zheng2024weak} adopt weak-to-strong extrapolation to enhance LLMs alignment.

\noindent\textbf{External Self-Refine.}~~These methods involve extrinsic feedback sources or computational tools to guide and correct the model’s reasoning. CRITIC~\citep{gou2023critic} systematically checks step-by-step outputs, enhancing the reliability of complex reasoning tasks. Reflexion~\citep{shinn2024reflexion} and Self-Debug~\citep{chen2023teaching} compare generated answers to reference solutions or few-shot exemplars, respectively, iteratively refining the logic. Techniques like FLARE~\citep{Bharadwaj2023FLAREFL} and Logic-LM~\citep{pan2023logic} incorporate references from external documents or symbolic solvers, thereby minimizing logical missteps. RARR~\citep{gao2022rarr} and SelfEvolve~\citep{jiang2023selfevolve} show that verifying intermediate states (e.g., compiler messages or relevant knowledge sources) is a powerful way to prune erroneous paths early and refine the model toward a correct solution.~\cite{rlhf2024b} proposes Iterative Preference Learning from Human Feedback, which includes an iterative version of the Direct Preference Optimization (DPO) algorithm for online settings, and a multi-step rejection sampling strategy for offline scenarios. PIT~\cite{selfimprovement2024a} implicitly learns the improvement goal from human preference data. 

\noindent\textbf{Fine-Tuned Intrinsic Self-Refine.}~~By fine-tuning the base model specifically for internal revision, these approaches systematically strengthen the LLM’s self-correction loops. Self-Critique~\citep{saunders2022self} aims to improve summarization via self-review, while SelFee~\citep{ye2023selfee} uses iterative feedback loops to ensure higher levels of logical consistency. Volcano~\citep{lee2023volcano} reduces multimodal hallucinations by fine-tuning a dedicated corrector module within the LLM’s architecture, and RL4F~\citep{akyurek2023rl4f} harnesses RL-based critique loops to raise performance by an average of 10\% on benchmarks requiring in-depth reasoning. REFINER~\citep{paul2023refiner} similarly concentrates on intermediate reasoning paths without changing the model’s original generation process, demonstrating that consistent improvements can be achieved by training the model to carefully re-examine its partial outputs. Additionally, the concept of easy-to-hard generalization has emerged as a promising variant of W2SG, where models are initially trained on easily verifiable examples before tackling more complex tasks~\citep{hase2024unreasonable}. One notable implementation of this approach involves training a strong reward model on human-verifiable examples, which then guide the supervision of more capable models on challenging tasks~\citep{sun2024easy}. In addition, the effectiveness of W2SG extends beyond LLMs, with successful applications demonstrated in computer vision tasks as well~\citep{guo2024vision}.

\noindent\textbf{Fine-Tuned External Self-Refine.}~~In scenarios where long-term improvements are crucial, the model’s parameters are updated via external feedback mechanisms. For example, Self-Edit~\citep{zhang2023self} regenerates code outputs based on execution results, leading to iterative improvements in correctness. Baldur~\citep{first2023baldur} strengthens theorem proving by adding or modifying context, while CodeRL~\citep{le2022coderl} employs test-based critics to verify functional accuracy in program synthesis tasks. Together, these techniques demonstrate that combining external resources with targeted fine-tuning fosters reliable, stepwise advancements in the model’s overall reasoning performance.

\subsection{Reinforcement Learning for Reasoning}\label{Section 5.2}
In Subsection \ref{Section 5.1}, we explored self-refine methods, a widely used approach to improve LLM reasoning through local tuning and optimization. This technique is typically applied to single-step tasks or output refinement, such as text generation and question answering, offering quick inference gains. However, it struggles with complex, long-term reasoning tasks requiring multi-step logic.
The release of OpenAI's o1 series~\citep{jaech2024openai} highlights reinforcement learning (RL) as a powerful alternative, training LLMs for advanced reasoning by refining long internal CoT through reward-based feedback. This significantly boosts performance in complex tasks like mathematical proofs and strategic planning. The o1 success has spurred research into large-scale RL, with models like QwQ-32B-Preview~\citep{qwq32} excelling in mathematics and programming, and DeepSeek-R1~\citep{DeepSeekAI2025DeepSeekR1IR} matching o1's capabilities. This subsection examines RL’s role in enhancing reasoning, focusing on DeepSeek-R1 and DeepSeek-R1-Zero, the leading open-source models.

\subsubsection{Formulating Reasoning as an MDP}
Reasoning within LLMs can be elegantly modeled as a sequential decision-making process, wherein the model iteratively constructs a series of intermediate steps \( a_1, a_2, \dots, a_T \) in response to an input query \( x \) to optimize the likelihood of arriving at a correct final answer. This conceptualization transforms reasoning into a structured framework amenable to reinforcement learning (RL), specifically through the lens of a Markov Decision Process (MDP), denoted as \( \mathcal{M} = (\mathcal{S}, \mathcal{A}, P, R, \gamma) \). The MDP encapsulates the dynamic interplay of states, actions, transitions, rewards, and temporal discounting, providing a robust mathematical foundation for training LLMs to navigate complex inference tasks. By framing reasoning as a sequence of deliberate choices, this approach enables the model to systematically explore and refine its logical pathways, drawing parallels to decision-making in domains like game playing or robotics, yet adapted to the unique challenges of linguistic and conceptual reasoning. The ultimate objective is to derive an optimal policy \( \pi^*(a_t | s_t) \) that maximizes the expected cumulative reward, expressed as \( J(\theta) = \mathbb{E}_{\pi_\theta} \left[ \sum_{t=1}^{T} \gamma^t R(s_t, a_t) \right] \), leveraging RL techniques such as Proximal Policy Optimization (PPO)~\citep{Schulman2017ProximalPO} or Advantage Actor-Critic (A2C)~\citep{mnih2016asynchronous} to iteratively enhance reasoning capabilities based on environmental feedback.

\noindent\textbf{State Space.}~~The state space \( \mathcal{S} \) forms the backbone of this MDP, with each state \( s_t \in \mathcal{S} \) representing the current reasoning trajectory at timestep \( t \), a rich composite of linguistic and structural elements critical to the inference process. Specifically, \( s_t \) encompasses the initial query \( x \), the sequence of prior reasoning steps \( \{a_1, \dots, a_{t-1}\} \), and an internal memory representation that encodes logical dependencies and intermediate conclusions, such as partial solutions or inferred relationships. This state evolves dynamically as reasoning unfolds, mirroring the progression of thought by integrating both the explicit path articulated through generated steps and latent knowledge distilled from contextual understanding. For instance, in a mathematical proof, \( s_t \) might include the problem statement, previously derived equations, and a memory of applicable theorems, enabling the model to maintain coherence across steps. This multifaceted state representation ensures that the LLM can adaptively track its reasoning context, a prerequisite for tackling tasks requiring sustained logical continuity, such as multi-step problem-solving or narrative coherence in text generation.
    
\noindent\textbf{Action Space.}~~The action space \( \mathcal{A} \) defines the range of possible decisions at each step, where an action \( a_t \in \mathcal{A} \) corresponds to the selection of the next reasoning move, offering a versatile toolkit for advancing the inference process. These actions may include generating a token or phrase in natural language to articulate a reasoning segment, applying a predefined logical or mathematical transformation (e.g., algebraic simplification), selecting a relevant theorem or rule from a knowledge base to extend the reasoning chain, or halting the process upon reaching a conclusive answer. The action space’s nature varies by task: it may be discrete, as in choosing from a finite set of logical rules in formal proofs, or continuous, as in producing free-form text in open-ended reasoning scenarios, reflecting the LLM’s generative flexibility. This duality allows the model to navigate both structured domains, like symbolic logic, and unstructured ones, like commonsense reasoning, adapting its strategy to the task’s demands while maintaining a coherent trajectory towards the solution.
    
\noindent\textbf{Transition Function.}~~Transition dynamics, encapsulated by the function \( P(s_{t+1} | s_t, a_t) \), govern how the state evolves with each action, delineating the progression of the reasoning trajectory within the MDP framework. In contrast to traditional RL environments where stochasticity arises from external variables (e.g., environmental noise), reasoning transitions in LLMs are predominantly deterministic, driven by the model’s autoregressive outputs or structured inference rules, such as applying a deductive step in a proof. However, uncertainties emerge from inherent model limitations—such as imperfect knowledge, ambiguous intermediate states, or probabilistic sampling in text generation—introducing variability that RL must address. For autoregressive LLMs, transitions follow a predictable sequence generation process, yet the potential for error accumulation or divergent interpretations necessitates a robust design to ensure reliability. This deterministic-yet-uncertain dynamic underscores the need for adaptive policies that can stabilize reasoning across diverse contexts, from precise mathematical derivations to nuanced narrative constructions.

\noindent\textbf{Reward Function.}~~The reward function \( R(s_t, a_t) \) serves as the evaluative core of the MDP, providing critical feedback on the quality of each reasoning step to guide the model’s learning process. Unlike conventional RL tasks with explicit rewards (e.g., points in a game), reasoning rewards must be carefully engineered to balance sparsity and density, reflecting the task’s complexity and goals. Sparse rewards, such as assigning a value only upon reaching a correct final answer, offer simplicity but may delay learning in multi-step scenarios, while dense rewards, which assess step-wise correctness, logical validity, or alignment with human preferences, provide granular guidance, as elaborated in \S \ref{rm for reasoning}. This flexibility allows the reward function to adapt to diverse reasoning demands—whether rewarding the application of a valid inference rule in a proof or the coherence of a narrative segment—ensuring that the model receives meaningful signals to refine its strategy across both immediate and extended inference horizons.

\noindent\textbf{Discount Factor.}~~$\gamma$: A scalar $\gamma \in [0,1]$ that determines the trade-off between immediate and future rewards. A higher $\gamma$ encourages multi-step reasoning optimization, promoting deep inference chains rather than short-term heuristics.
Given this MDP formulation, the objective is to learn an optimal reasoning policy $\pi^*(a_t | s_t)$ that maximizes the expected cumulative reward:
\begin{equation}
J(\theta) = \mathbb{E}_{\pi_\theta} \left[ \sum_{t=1}^{T} \gamma^t R(s_t, a_t) \right].
\end{equation}
This framework enables the application of reinforcement learning techniques such as such as Proximal Policy Optimization (PPO) \cite{Schulman2017ProximalPO} or Advantage Actor-Critic (A2C) \cite{mnih2016asynchronous} to refine LLM reasoning capabilities by iteratively adjusting the policy $\pi_\theta$ based on feedback from the reasoning environment.

\subsubsection{Reward Design for Reasoning}\label{rm for reasoning}
Unlike traditional RL tasks with clear rewards like game scores, reasoning in LLMs demands structured reward designs reflecting correctness, efficiency, and informativeness. Common approaches include binary correctness rewards, assigning \( r_T = 1 \) for a correct final answer and \( r_T = 0 \) otherwise, which is simple but introduces high variance due to sparse feedback; step-wise accuracy rewards, offering incremental feedback based on metrics like inference rule validity or intermediate step consistency to guide multi-step reasoning; self-consistency rewards, measuring stability across multiple reasoning paths and assigning higher rewards for agreement to enhance robustness; and preference-based rewards, derived from RLHF or RLAIF, where a model \( r_\phi(s_t, a_t) \) trained on human or AI feedback evaluates reasoning quality, providing nuanced guidance for complex tasks.

\subsubsection{Large-Scale RL on Base Model}
\label{sec:reasoning-oriented RL}
\begin{figure}[h]
    \centering
    \includegraphics[width=0.85\linewidth]{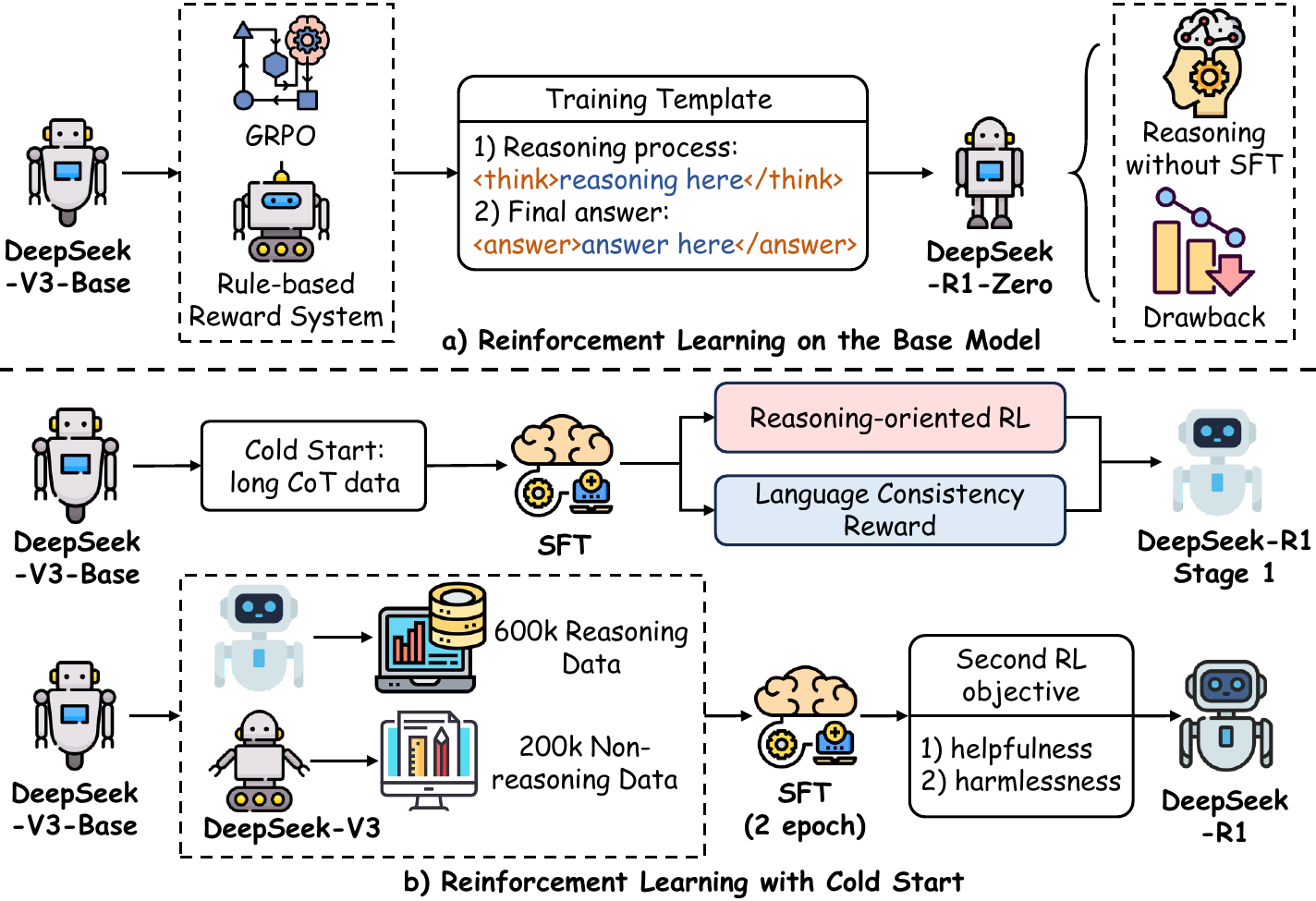}
    \caption{Workflow of Reinforcement Learning for Reasoning in DeepSeek-R1, illustrating the process for optimizing reasoning capabilities in Large Language Models.}
    \label{fig:reasoning RL}
\end{figure}

Large-scale Reinforcement Learning has emerged as a transformative post-training paradigm for enhancing the reasoning capabilities of LLMs, shifting the focus from traditional SFT to dynamic, self-evolving optimization strategies. This approach leverages extensive computational frameworks and iterative reward-based feedback to refine base models directly, bypassing the need for pre-annotated datasets and enabling autonomous development of complex inference skills. By integrating large-scale RL, LLMs can address intricate multi-step reasoning tasks (e.g., mathematical problem-solving, logical deduction, and strategic planning), where conventional SFT often falls short due to its reliance on static, human-curated data~\citep{Ouyang2022TrainingLM}. The DeepSeek-R1 model exemplifies this paradigm, employing advanced RL techniques to achieve state-of-the-art reasoning performance while optimizing resource efficiency, as illustrated in \textbf{Fig. \ref{fig:reasoning RL}}. This subsection delineates the key methodologies underpinning DeepSeek-R1’s success, including novel optimization algorithms, adaptive exploration, and trajectory management, which collectively redefine the potential of RL-driven reasoning in LLMs.

\noindent\textbf{Group Relative Policy Optimization.}~~The DeepSeek-R1-Zero model leverages a sophisticated variant of Proximal Policy Optimization (PPO), termed Group Relative Policy Optimization (GRPO), to mitigate the substantial computational and resource demands inherent in traditional RL training for LLMs. Unlike standard PPO, which relies on extensive critic networks, GRPO employs a group-based baseline estimation to streamline the optimization process, significantly reducing training overhead while preserving the robustness of policy updates. This efficiency enables large-scale RL deployment on resource-constrained systems, facilitating iterative refinement of reasoning strategies across extended trajectories. By optimizing the policy within manageable computational bounds, GRPO positions DeepSeek-R1-Zero as a scalable solution for enhancing reasoning capabilities, as depicted in \textbf{Fig. \ref{fig:reasoning RL}}, making it a cornerstone of contemporary RL-driven inference research.

\noindent\textbf{DeepSeek-R1-Zero.}~~DeepSeek-R1-Zero exemplifies the transformative potential of large-scale RL to elevate LLM reasoning without the conventional reliance on SFT as an initial step, instead adopting a pure RL-driven self-evolution paradigm. This approach enables the model to autonomously develop sophisticated reasoning skills by iteratively refining its internal CoT through reward feedback, bypassing the need for pre-annotated datasets typically required in SFT. The result is a marked improvement in performance across complex, multi-step reasoning tasks (e.g., mathematical problem-solving and logical derivations) demonstrating RL’s capacity to unlock advanced inference capabilities from a base model. Positioned as one of the strongest open-source reasoning models, DeepSeek-R1-Zero’s success underscores the viability of cold-start RL strategies, offering a resource-efficient alternative to traditional training pipelines while achieving parity with state-of-the-art benchmarks.

\noindent\textbf{Stepwise Reward Modeling.}~~To guide reasoning across a trajectory \( \tau = (s_1, a_1, \dots, s_T, a_T) \), DeepSeek-R1 employs a stepwise reward model \( f_\theta \) that delivers granular feedback at each timestep, defined as \( r_t = f_\theta(s_t, a_t \mid \mathcal{D}_{\text{reasoning}}) \), where \( \mathcal{D}_{\text{reasoning}} \) comprises human-annotated CoT sequences with step-level correctness labels. This dense reward structure contrasts with sparse end-of-sequence rewards by providing immediate, actionable insights into the quality of individual reasoning steps, enabling the model to fine-tune its strategies with precision. By leveraging expertly curated data, the reward model ensures that feedback aligns with human reasoning standards, fostering consistency and accuracy across extended inference chains, a critical feature for tackling tasks requiring protracted logical synthesis.

\noindent\textbf{Adaptive Exploration.}~~DeepSeek-R1 enhances policy optimization through an adaptive exploration mechanism integrated into its objective:
\begin{align}
\mathcal{L}_{\text{PPO+}} &= \mathbb{E}_\tau \left[ \min\left( \frac{\pi_\phi(a|s)}{\pi_{\text{old}}(a|s)} A_t, \text{clip}\left(\frac{\pi_\phi(a|s)}{\pi_{\text{old}}(a|s)}, 1-\epsilon, 1+\epsilon\right) A_t \right) \right] \nonumber \\
&\quad + \lambda_t \mathcal{H}(\pi_\phi(\cdot|s)),
\end{align}
where the entropy term \( \mathcal{H} \) is modulated by an adaptive coefficient \( \lambda_t = \alpha \cdot \exp(-\beta \cdot \text{Var}(R(\tau_{1:t}))) \), dynamically adjusting based on reward variance across the trajectory. This approach balances exploration and exploitation, encouraging the model to explore diverse reasoning paths early in training while converging to optimal strategies as variance decreases, thereby enhancing both robustness and efficiency in reasoning refinement.

\noindent\textbf{Trajectory Pruning.}~~To optimize computational efficiency during reasoning, DeepSeek-R1 implements a dual-attention critic \( V_\psi(s_t) = \text{LocalAttn}(s_t) + \text{GlobalAttn}(s_{1:t}) \), which evaluates the value of each state by combining local step assessments with global trajectory context. Pruning occurs when \( V_\psi(s_t) < \gamma \cdot \max_{k \leq t} V_\psi(s_k) \), terminating low-value reasoning paths to focus resources on promising trajectories. This mechanism reduces wasteful exploration, accelerates convergence, and ensures that the model prioritizes high-quality reasoning sequences, contributing to its exceptional performance in complex inference tasks.

\subsubsection{RL for Reasoning with Cold Start}

DeepSeek-R1-Zero further advances RL’s application by adopting a cold-start approach, eschewing SFT and relying entirely on large-scale RL from an untrained base model. This self-evolutionary strategy refines reasoning through iterative feedback, generating robust CoT sequences without pre-annotated data dependencies. By training directly on reasoning tasks, DeepSeek-R1-Zero demonstrates RL’s versatility, achieving performance comparable to or exceeding models initialized with SFT, such as its DeepSeek-R1 counterpart. This approach not only reduces reliance on extensive labeled datasets but also showcases RL’s potential to autonomously develop complex reasoning capabilities, offering a scalable paradigm for future LLM development. Collectively, RL provides a promising framework for enhancing reasoning, with effective reward design, policy optimization (e.g., GRPO), and exploration strategies remaining critical. Future research could explore hybrid methods integrating imitation learning or self-supervised objectives to further refine these capabilities, solidifying RL’s role in advancing LLM inference.
\section{PoLMs for Efficiency}\label{Section 6}

Building on the post-training optimization techniques discussed in earlier chapters, post-training efficiency specifically targets the operational performance of LLMs after their initial pre-training. The principal goal is to optimize key deployment metrics (e.g., processing speed, memory usage, and resource consumption), thereby making LLMs more practical for real-world applications. Approaches to achieving post-training efficiency fall into three main categories: \textbf{Model Compression} (\S \ref{Section 6.1}), which reduces the overall computational footprint through techniques such as pruning and quantization; \textbf{Parameter-Efficient Fine-Tuning} (\S \ref{Section 6.2}), which updates only a fraction of a model’s parameters or employs specialized modules, thus minimizing retraining costs and accelerating adaptation to new tasks; and \textbf{Knowledge Distillation} (\S \ref{Section 6.3}), which transfers the knowledge from a larger, pre-trained model to a smaller model, enabling the smaller model to achieve comparable performance with reduced resource demands.

\subsection{Model Compression}\label{Section 6.1}
Model compression encompasses a set of techniques designed to reduce the size and computational demands of LLMs, which includes post-training quantization, parameter pruning, and low-rank approximation. 

\subsubsection{Post-training Quantization}
\begin{figure}[h]
    \centering
    \includegraphics[width=0.7\linewidth]{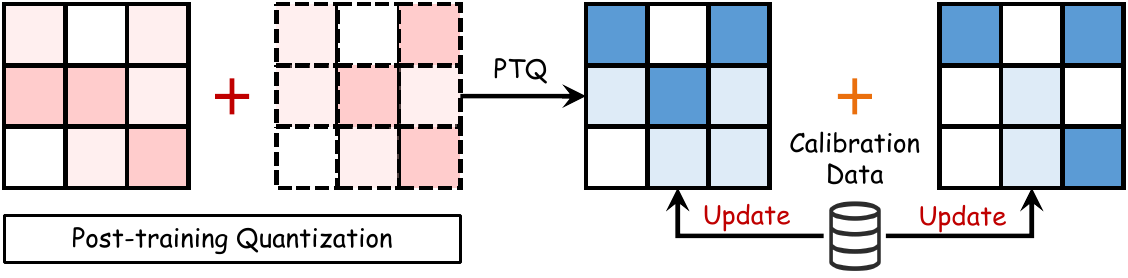}
    \caption{Illustrations of post-training quantization techniques for LLMs.}
    \label{fig:PTQ}
\end{figure}

A crucial compression method for LLMs is quantization, which converts high-precision data types $X^H$ (30-bit floating point) into lower-precision formats $X^L$ (8-bit integer)\cite{dettmers2024qlora}. This conversion is formulated as:
\begin{equation}
    X^L = \text{Round} (\frac{\text{absmax}(X^L)}{\text{absmax}(X^H)} X^H ) = \text{Round}(\mathcal{K} \cdot X^H),
\end{equation}
where $\mathcal{K}$ represents the quantization constant, and \text{absmax} refers to the absolute maximum of the elements. The function \text{Round} transforms floating-point numbers into integers. LLM quantization encompasses both post-training quantization (PTQ) and quantization-aware training (QAT). PTQ enables adjustments to model weights and activations after pre-training, using a small calibration dataset to optimize for both computational efficiency and performance as illustrated in \textbf{Fig. \ref{fig:PTQ}}. Additionally, \textbf{Table \ref{tab:quantization}} presents the performance metrics of several prominent quantization methods for LLMs.

\begin{table*}[h]
\centering
\caption{Overview of Quantization Methods for Large Language Models (2021–2025). This table summarizes representative quantization techniques, detailing their main LLMs, bit widths, perplexity differences, speedups, and release timelines across three metrics: \textbf{Bit Width} (bits for weights, activations, and KV cache), \textbf{Perplexity Difference} (performance variation on Wikitext-2 and C4 datasets), and \textbf{Speedup} (computational speed improvement relative to baseline models).}
\label{tab:quantization}
\renewcommand{\arraystretch}{1.1}
\resizebox{\textwidth}{!}{%
\begin{tabular}{cccccccccr}
\toprule
\multirow{2}{*}{\textbf{Type}} & \multirow{2}{*}{\textbf{Methods}} & \multirow{2}{*}{\textbf{Main LLMs}} & \multicolumn{3}{c}{\textbf{Bit Width}} & \multicolumn{2}{c}{\textbf{Perplexity Difference}} & \multirow{2}{*}{\textbf{Speedup}} & \multirow{2}{*}{\textbf{Release Time}} \\ \cmidrule(lr){4-8}
 &  &  & \textbf{W} & \textbf{A} & \textbf{KV} & \textbf{Wikitext-2} & \textbf{C4} & & \\ \midrule
\rowcolor{Gray} & \textbf{GPTQ}~\citep{frantar2022gptq} & OPT-175B & 3 & 16 & 16 & 0.34 & 0.23 & 3.24$\times$ & Mar-2023\\
\rowcolor{LightGray} & \textbf{SpQR}~\citep{dettmers2024qlora} & LLaMA-30B & 3.89 & 16 & 16 & 0.15 & 0.1 & 2.0$\times$ & Jun-2023\\
\rowcolor{Gray} & \textbf{QuIP}~\citep{chee2024quip} & LLaMA2-70B & 2 & 16 & 16 & 3.007 & 3.228 & - & 2023\\
\rowcolor{LightGray} & \textbf{AWQ}~\citep{lin2024awq} & LLaMA2-70B & 3 & 16 & 16 & 0.42 & - & 3.2$\times$ & 2024\\
\rowcolor{Gray} & \textbf{OWQ}~\citep{lee2024owq} & LLaMA-65B & 3.01 & 16 & 16 & 0.72 & - & - & Mar-2024\\
\rowcolor{LightGray} & \textbf{EasyQuant}~\citep{tang2024easyquant} & LLaMA-65B & 4 & 16 & 16 & 3.98 & 6.30 & - & Mar-2024\\
\rowcolor{Gray} & \textbf{Agile-Quant}~\citep{tang2024easyquant} & BLOOM-7.1B & 8 & 8 & 16 & 11.73 & 14.70 & 1.8x & Mar-2024\\
\rowcolor{LightGray} & \textbf{LUT-GEMM}~\citep{park2022lut} & LLaMA-65B & 3 & 16 & 16 & 0.14 & - & 2.04$\times$ & Apr-2024\\
\rowcolor{Gray} & \textbf{SqueezeLLM}~\citep{kim2023squeezellm} & LLaMA-13B & 3 & 16 & 16 & 0.51 & 0.67 & 2.4$\times$ & Jun-2024\\
\rowcolor{LightGray} & \textbf{DAQ}~\citep{luo2024daq} & LLaMA2-30B & 3 & 16 & 16 & 4.23 & - & - & Oct-2024\\
\rowcolor{Gray} & \textbf{MobileQuant}~\citep{tan2024mobilequant} & TinyLlaMA-1.1B & 4 & 16 & 16 & 15.6 & - & - & Oct-2024\\
\rowcolor{LightGray} & \textbf{GWQ}~\citep{shao2024gwq} & LLaMA2-13B & 4.63 & 16 & 16 & 4.88 & 6.47 & 1.2$\times$ & Dec-2024\\ 
\multirow{-14}{*}{\begin{tabular}[c]{@{}c@{}}\textbf{WOQ}\end{tabular}}   \\
\noalign{\vskip-1.20em}
\midrule

\rowcolor{Gray} & \textbf{QT}~\citep{angelidis2021extractive} & OPT-1.3B & 8 & 8 & 16 & 17.74 & - & - & Mar-2021\\
\rowcolor{LightGray} & \textbf{ZeroQuant}~\citep{yao2022zeroquant} & GPT-J-6B & 8 & 8 & 16 & 0.16 & - & 3.67$\times$ & 2022\\
\rowcolor{Gray} & \textbf{LLM.int8()}~\citep{dettmers2022gpt3} & OPT-13B & 8 & 8 & 16 & - & 0.00 & 1.22$\times$ & 2022\\
\rowcolor{LightGray} & \textbf{RPTQ}~\citep{yuan2023rptq} & OPT-175B & 4 & 4 & 16 & 2.26 & 2.15 & - & May-2023\\
\rowcolor{Gray} & \textbf{Olive}~\citep{guo2023olive} & BLOOM-7B & 4 & 4 & 16 & 2.11 & 2.24 & 4.5$\times$ & Jun-2023\\
\rowcolor{LightGray} & \textbf{ZeroQuant-FP}~\citep{wu2023zeroquant} & LLaMA-30B & 4 & 8 & 16 & 0.18 & 0.13 & - & Jul-2023\\
\rowcolor{Gray} & \textbf{SmoothQuant}~\citep{xiao2023smoothquant} & OPT-175B & 8 & 8 & 16 & 0.18 & - & 1.56$\times$ & Oct-2023\\
\rowcolor{LightGray} & \textbf{OS+}~\citep{wei2023outlier} & LLaMA-65B & 4 & 4 & 16 & 5.77 & - & - & Oct-2023\\
\rowcolor{Gray} & \textbf{OmniQuant}~\citep{shao2023omniquant} & LLaMA-7B & 4 & 6 & 16 & 0.41 & 0.55 & - & Mar-2024\\
\rowcolor{LightGray} & \textbf{RoLoRA}~\citep{huang2024rolora} & LLaMA3-8B & 4 & 16 & 16 & - & - & - & Sep-2024\\
\rowcolor{Gray} & \textbf{HotaQ}~\citep{shen2024hotaq} & LLaMA & 4 & 4 & 8 & - & - & 2.5$\times$ & Oct-2024\\
\rowcolor{LightGray} & \textbf{Q-DiT}~\citep{chen2024q} & LLaMA3-8B & 6 & 8 & 16 & - & - & - & Nov-2024\\ 
\multirow{-14}{*}{\begin{tabular}[c]{@{}c@{}}\textbf{WAQ}\end{tabular}}   \\
\noalign{\vskip-1.20em}
\midrule

\rowcolor{Gray} & \textbf{KVQuant}~\citep{hooper2024kvquant} & LLaMA-65B & 16 & 16 & 2 & 0.19 & 0.11 & 1.4$\times$ & Jan-2024\\
\rowcolor{LightGray} & \textbf{WKVQuant}~\citep{yue2024wkvquant} & LLaMA-13B & 4 & 16 & 4 & 0.12 & 0.14 & - & Feb-2024\\
\rowcolor{Gray} & \textbf{QAQ}~\citep{dong2024qaq} & LLaMA3-8B & 4 & 16 & 16 & - & - & 10$\times$ & Apr-2024\\
\rowcolor{LightGray}& \textbf{ZipCache}~\citep{he2024zipcache} & LLaMA3-8B & 4 & 16 & 16 & 0.38 & - & 4.98$\times$ & May-2024\\
\rowcolor{Gray} & \textbf{KIVI}~\citep{liu2024kivi} & LLaMA2-13B & 4 & 16 & 16 & - & - & 2.6$\times$ & Jul-2024\\
\rowcolor{LightGray} & \textbf{DL-QAT}~\citep{ke2024dl} & LLaMA2-7B & 4 & 16 & 16 & 6.3 & - & - & Nov-2024\\
\multirow{-8}{*}{\begin{tabular}[c]{@{}c@{}}\textbf{KVQ}\end{tabular}}   \\
\noalign{\vskip-1.20em}
\bottomrule
\end{tabular}
}
\end{table*}

\noindent\textbf{Weight-Only Quantization (WOQ).}~~WOQ focuses on compressing model weights to improve efficiency. GPTQ~\citep{frantar2023sparsegpt} applies layer-wise quantization using Optimal Brain Quantization (OBQ), reducing weights to 3 or 4 bits to lower memory usage and processing time. To push efficiency further, QuIP~\citep{chee2024quip} introduces incoherence processing for 2-bit quantization, offering an even more compact representation. Similarly, AWQ~\citep{lin2024awq} and OWQ~\citep{lee2024owq} address accuracy retention by maintaining high precision for particularly sensitive weights, thereby minimizing potential accuracy losses during inference. Finally, SpQR~\citep{dettmers2024qlora} combines sparse quantization with decoding, enabling efficient token-by-token inference while preserving model responsiveness.
    
\noindent\textbf{Weight-Activation Co-Quantization (WAQ).}~~WAQ integrates weights and activations to enhance efficiency. LLM.int8()~\citep{dettmers2022gpt3} addresses activation outliers using precise storage and quantizes to 8 bits while maintaining performance. SmoothQuant~\citep{xiao2023smoothquant} implements per-channel scaling, transferring the quantization difficulties from activations to weights for lossless results. Additionally, OS$+$~\citep{wei2023outlier} mitigates the impact of outliers through channel-wise shifting and scaling, thereby boosting efficiency. OmniQuant~\citep{shao2023omniquant} redirects the quantization hurdles from activations to weights and fine-tunes clipping thresholds for extreme values. To further enhance efficiency, RPTQ~\citep{li2024turning} groups similar channels to ensure uniformity in quantization parameters. 
    
\noindent\textbf{KV-Cache Quantization (KVQ).}~~KV-Cache Quantization addresses memory optimization challenges in LLMs, especially as input token counts increase. KVQuant~\citep{hooper2024kvquant} introduces tailored methods for efficient inference with large context lengths, maintaining performance with minimal loss. KIVI~\citep{liu2024kivi} optimizes memory savings by applying distinct quantization strategies for key and value caches, achieving 2-bit quantization without fine-tuning. WKVQuant~\citep{yue2024wkvquant} further refines this with a two-dimensional quantization strategy and cross-block regularization, delivering memory efficiency comparable to weight-activation quantization with nearly the same performance.

\subsubsection{Parameter Pruning}
\begin{figure}[h]
    \centering
    \includegraphics[width=0.55\linewidth]{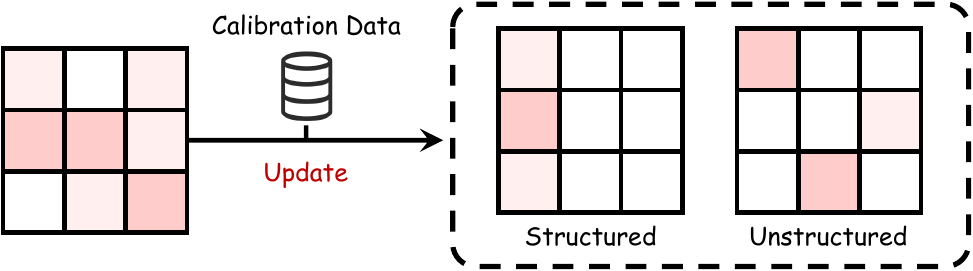}
    \caption{Illustrations of pruning parameters techniques for LLMs.}
    \label{fig:parameter pruning}
\end{figure}

Parameters Pruning~\citep{lecun1989optimal} is a crucial technique for improving the efficiency of LLMs by minimizing model size and complexity without sacrificing accuracy. As shown in \textbf{Fig. \ref{fig:parameter pruning}}, pruning can be divided into Unstructured Pruning, and Structured Pruning.

\noindent\textbf{Unstructured Pruning.}~~Unstructured pruning enhances the sparsity of LLMs by eliminating weights that are not critical. The approach known as SparseGPT~\citep{frantar2023sparsegpt} achieves as much as 60\% sparsity through one-shot pruning while maintaining minimal loss. The method Wanda~\citep{sun2023simple} performs pruning based on weight magnitudes and activations without requiring retraining. Meanwhile, SAMSP~\citep{shao2024one} leverages the sensitivity of the Hessian matrix for dynamic adjustments to sparsity, aimed at minimizing errors. DSnoT~\citep{du2024bitdistiller} improves performance by employing iterative pruning cycles. Finally, Flash-LLM~\citep{xia2023flash} retrieves sparse weights from global memory and reconstructs them densely in on-chip buffers to facilitate efficient computation.
    
\noindent\textbf{Structured Pruning.}~~This approach focuses on pruning entire parameter groups in LLMs to enhance hardware efficiency and simplify structures. For instance, LLM-runer~\citep{ma2023llm} assesses importance for LLaMA~\citep{Touvron2023LLaMAOA} and uses LoRA~\citep{Hu2021LoRALA} to recover accuracy post-pruning. FLAP~\citep{an2024fluctuation} optimizes compression without fine-tuning using a structured metric. Additionally, SliceGPT~\citep{ashkboos2024slicegpt} employs PCA for pruning while maintaining efficiency. Sheared LLaMA~\citep{xia2023sheared} refines model shape with regularization-based pruning. LoRAPrune~\citep{zhang2024loraprune} enhances efficiency with iterative structural pruning based on LoRA importance. Furthermore, Deja Vu~\citep{liu2023deja} reduces latency, maintaining accuracy by predicting key attention heads and MLP parameters, using contextual sparsity.

\noindent\textbf{Low-Rank Approximation.}~~Low-rank approximation serves to compress LLMs by approximating a weight matrix $W$ with smaller matrices $U$ and $V$, thereby achieving $W \approx UV^\top$. This methodology not only reduces the number of parameters but also enhances operational efficiency. For instance, TensorGPT~\citep{xu2023tensorgpt} employs Tensor-Train Decomposition (TTD) to develop a more efficient embedding format. LoSparse~\citep{li2023losparse} integrates low-rank approximation with pruning, specifically aimed at compressing coherent neuron components. FWSVD~\citep{hsu2022language} implements a weighted SVD approach, whereas ASVD~\citep{yuan2023asvd} provides a training-free SVD alternative, both targeting post-training efficiency. Lastly, SVD-LLM~\citep{wang2024svd} further improves compression by establishing a direct relationship between singular values and compression loss.

\subsection{Parameter-Efficient Fine-Tuning}\label{Section 6.2}
\begin{figure}[h]
    \centering
    \includegraphics[width=0.75\linewidth]{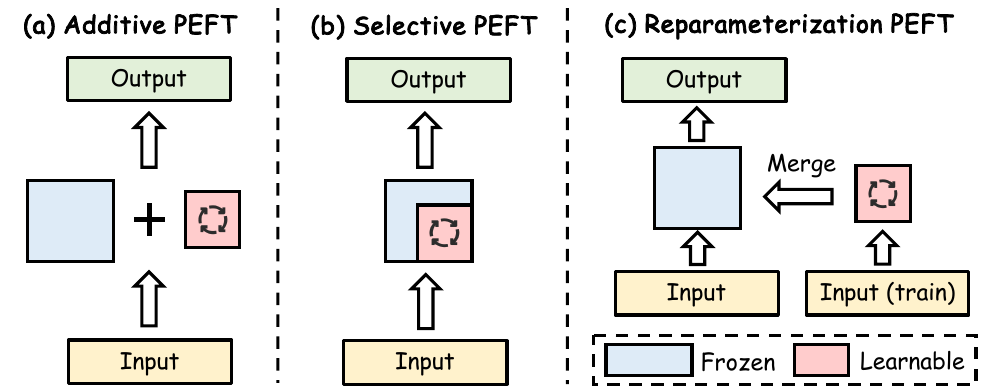}
    \caption{Illustrations of Parameter-Efficient Fine-Tuning (PEFT), illustrating approaches for resource-efficient adaptation in Large Language Models.}
    \label{fig:peft}
\end{figure}

The procedure for parameter-efficient fine-tuning (PEFT) consists of freezing the complete LLM backbone while only modifying a limited number of newly added parameters. As depicted in \textbf{Fig. \ref{fig:peft}}, PEFT methods are divided into four categories: additive PEFT, selective PEFT, reparameterized PEFT, and hybrid PEFT.

\subsubsection{Additive PEFT}
Additive PEFT incorporates new trainable modules to the LLM without changing the original parameters, allowing task-specific tuning while retaining the base model's knowledge, which is efficient for fine-tuning.

\noindent\textbf{Adapters.}~~Adapter integrates compact layers within transformer blocks, defined as:
\begin{equation}
    \text{Adapter}(x) = W_{\text{up}} \sigma(W_{\text{down}} x) + x,
\end{equation}
where an adapter layer includes a down-projection matrix $W_{\text{down}} \in \mathbb{R}^{r \times d}$, a non-linear activation $\sigma$, and an up-projection matrix $W_{\text{up}} \in \mathbb{R}^{d \times r}$. Here, $d$ is the hidden layer dimension, and $r$ is the bottleneck dimension, reducing complexity while maintaining performance. Building on this structure, Serial Adapter~\citep{houlsby2019parameter} introduced two modules per transformer block. AdapterFusion~\citep{pfeiffer2020adapterfusion} improved efficiency by placing adapters post $Add \& Norm$. Parallel Adapter (PA)~\citep{he2021towards} ran adapters parallel to sublayers, while CoDA~\citep{lei2023conditional} optimized by running adapters in parallel with sublayers. Unlike AdapterFusion, MerA~\citep{he2023mera} unified adapters using optimal transport techniques for weights and activations.

\noindent\textbf{Soft Prompt.}~~Soft prompt enhances model performance by adding adjustable vectors to the input sequence instead of optimizing discrete tokens~\citep{petrov2023prompting}. This approach is formalized as:
\begin{equation}
     X^{(l)} = [s_1^{(l)}, \dots, s_{N_S}^{(l)}, x_1^{(l)}, \dots, x_{N_X}^{(l)}],
\end{equation}
where $s_i^{(l)}$ denotes soft prompt tokens, and $x_i^{(l)}$ represents original input tokens. $N_S$ and $N_X$ are the counts of soft prompt and original input tokens, respectively. Prefix Tuning~\citep{li2021supervision} introduces learnable vectors between transformer layers, stabilized by reparameterization and refined by P-Tuning v2~\citep{liu2021p} and APT~\citep{zhang2023towards}. Meanwhile, Prompt Tuning~\citep{lester2021power} focuses on the initial embedding layer for large model optimization with low computational cost. Xprompt~\citep{ma2022xprompt} and IDPG~\citep{wu2022idpg} streamline prompt generation and insertion. Methods like SPoT~\citep{vu2021spot} and PTP~\citep{chen2023ptp} then address stability and convergence speed, while DePT~\citep{shi2023dept} and SMoP~\citep{choi2023smop} reduce computational demands through optimized prompt structures.

\noindent\textbf{Other Additive Methods.}~~In addition to earlier techniques, methods such as (IA)$^3$~\citep{liu2022few} and SSF~\citep{lian2022scaling} focus on post-training efficiency by introducing minimal but powerful adjustments to model parameters. The self-attention and FFN operations are mathematically defined as:
\begin{equation}
    SA(x) = \text{Softmax}\left(\frac{Q \cdot (l_k \odot K)^T}{\sqrt{d_{head}}}\right) \cdot (l_v \odot V),
\end{equation}

\begin{equation}
    FFN_{transformer}(x) = W_{up} \cdot (l_{ff} \odot \sigma(W_{down}x)),
\end{equation}
where $\odot$ represents the Hadamard product, the scale vectors $l_k$ and $l_v$ can be smoothly incorporated into the weight matrices of $A_Q$ and $A_W$. Additionally, IPA ~\citep{lu2023inference} aligns LLMs like GPT-4 with user-specific requirements. Moreover, it does not require changes to the underlying model and therefore maintains efficiency during the fine-tuning process.

\subsubsection{Selective PEFT}
Selective PEFT enhances efficiency by fine-tuning only a subset of parameters, as shown in \textbf{Fig. \ref{fig:peft}(b)}. This involves applying a binary mask $M = \{m_1, m_2, \dots, m_n\}$ to the parameters $\theta = \{\theta_1, \theta_2, \dots, \theta_n\}$, where each $m_i$ indicates if $\theta_i$ is selected for fine-tuning. The updated parameter set is expressed as:
\begin{equation}
    \theta'_i = \theta_i - \eta \cdot m_i \cdot \frac{\partial \mathcal{L}}{\partial \theta_i},
\end{equation}
where $\eta$ is the learning rate, and $\frac{\partial \mathcal{L}}{\partial \theta_i}$ is the gradient of the loss function. Only selected parameters (where $m_i = 1 $) are updated, reducing computational costs while maintaining effectiveness. Early approaches include Diff pruning~\citep{guo2020parameter}, which regularizes a learnable binary mask using a differentiable \( L_0 \)-norm, and FishMask~\citep{sung2021training}, which selects parameters based on fisher information for greater relevance. LT-SFT~\citep{ansell2021composable} applies the Lottery Ticket Hypothesis to identify impactful parameters. SAM~\citep{fu2023effectiveness} employs second-order approximations for selection, while Child-tuning~\citep{xu2021raise} selects parameters dynamically in a child network. Additionally, FAR~\citep{vucetic2022efficient} and BitFit~\citep{zaken2021bitfit} further exemplify selective PEFT by focusing on optimizing specific parameter groups.

\subsubsection{Reparameterized PEFT}
Reparameterized PEFT mainly employs a low-rank parameterization to improve efficiency, as illustrated in \textbf{Fig. \ref{fig:peft}(c)}. LoRA (Low Rank Adaptation)\cite{Hu2021LoRALA} introduces two trainable matrices, $W_{\text{up}} \in \mathbb{R}^{d \times r}$ and $W_{\text{down}} \in \mathbb{R}^{r \times k}$, modifying the output as:
\begin{equation}
    h_{\text{out}} = W_0 h_{\text{in}} + \alpha (W_{\text{up}} W_{\text{down}} h_{\text{in}}),
\end{equation}
where $\alpha$ is a scaling factor. This approach allows efficient adaptation to new tasks while preserving core knowledge. Building on LoRA, Intrinsic SAID~\citep{aghajanyan2020intrinsic} minimizes the fine-tuning parameter space, further reducing computational demands. Dynamic variants, including DyLoRA~\citep{valipour2022dylora} and AdaLoRA~\citep{zhang2023adalora}, adapt rank dynamically based on task-specific requirements, with AdaLoRA additionally incorporating SVD-based pruning for greater efficiency. SoRA~\citep{ding2023sparse} simplifies processes by removing orthogonality constraints, while Laplace-LoRA~\citep{yang2023bayesian} applies Bayesian calibration for fine-tuning. 
Compacter~\citep{karimi2021compacter} and VeRA~\citep{kopiczko2023vera} further reduce parameter complexity. Additionally, DoRA~\cite{liu2024dora} optimizes updates in the directional component and HiRA~\cite{huang2025hira} employs the Hadamard product for high-rank updates, thereby enhancing both efficiency and performance. To tackle multiple tasks and evolving domains, Terra~\cite{zhuang2024time} integrates a time-varying matrix, and ToRA~\cite{wang2025mtsam} utilizes a Tucker decomposition to further improve the LoRA structure.
In addition to structural design, PiSSA~\cite{meng2025pissa} and LoRA-GA~\cite{wang2025lora} optimize LoRA's initialization using SVD and gradient alignment. Meanwhile, LoRA+~\cite{hayou2024lora+}, LoRA-Pro~\cite{wang2024lora}, and CopRA~\cite{zhuang2024copra} further refine the gradient update strategies. Additionally, ComLoRA~\cite{huang2025colora} employs competitive learning to select the best-performing LoRA component.

\subsubsection{Hybrid PEFT}
Hybrid PEFT methods enhance post-training efficiency by integrating or optimizing various fine-tuning strategies. A prominent technique, UniPELT~\citep{mao2021unipelt}, merges LoRA, prefix tuning, and adapters within transformer blocks. This method dynamically activates components through a gating mechanism managed by feedforward networks (FFNs) that produce scalars $G \in [0, 1]$, ultimately optimizing parameter utilization. Another innovative method, MAM Adapter~\citep{he2021towards}, refines this technique by strategically positioning prefix tuning in self-attention layers and using scaled parallel adapters in feedforward layers. Furthermore, NAS-based approaches such as NOAH~\citep{zhang2022neural} and AUTOPEFT~\citep{zhou2024autopeft} improve post-training efficiency by identifying optimal PEFT configurations tailored to specific tasks. Additionally, HeadMap~\citep{wang2025headmap} identifies a series of attention heads (i.e., knowledge circuits) that play a key role in certain tasks using a greedy approach, and efficiently improves model performance by mapping the output of these attention heads back into the residual flow of LLM. Finally, LLM-Adapters~\citep{hu2023llm} provide a framework for integrating various PEFT techniques within LLMs, ensuring the most effective module placements to maintain efficiency across different model scales.

\subsection{Knowledge Distillation}\label{Section 6.3}

Knowledge Distillation (KD) constitutes a cornerstone technique in the post-training optimization of LLMs, enabling the transfer of knowledge from a large, pre-trained teacher model to a compact student model to enhance efficiency without sacrificing performance. Initially introduced in the context of model compression, KD has garnered substantial attention for its capacity to distill complex knowledge into resource-efficient architectures, enabling deployment in constrained environments such as edge devices and embedded systems. By leveraging the nuanced output distributions of a teacher model—richer than traditional hard labels—KD empowers the student to replicate not only class predictions but also the inter-class relationships and subtle patterns ingrained in the teacher’s representations. This process typically involves optimizing a composite loss function that balances supervised learning objectives with distillation-specific goals, substantially reducing computational and memory demands while preserving generalization capabilities.

The fundamental mechanism of KD hinges on minimizing a hybrid loss that integrates traditional classification loss with a distillation term. Formally, given a teacher model’s soft output probabilities \(\mathbf{p_t}\) and a student model’s predictions \(\mathbf{p_s}\), alongside true labels \(\mathbf{y}\) and student outputs \(\mathbf{y_s}\), the KD loss is expressed as:

\begin{equation}
    \mathcal{L}_{KD} = \alpha \mathcal{L}_{CE}(\mathbf{y}, \mathbf{y_s}) + (1 - \alpha) \mathcal{L}_{KL}(\mathbf{p_t}, \mathbf{p_s}),
\end{equation}
where \(\mathcal{L}_{CE}\) denotes the cross-entropy loss capturing alignment with ground truth, \(\mathcal{L}_{KL}\) represents the Kullback-Leibler divergence~\citep{kullback1951information} measuring divergence between teacher and student distributions, and \(\alpha \in [0, 1]\) is a hyperparameter modulating the trade-off between these objectives. The soft targets \(\mathbf{p_t}\), often tempered by a temperature parameter \(T\) (i.e., \(\mathbf{p_t} = \text{softmax}(\mathbf{z_t}/T)\), where \(\mathbf{z_t}\) are teacher logits), encode richer probabilistic information, enabling the student to emulate the teacher’s decision-making nuances beyond mere label accuracy.


\begin{table*}[h]
\centering
\renewcommand{\arraystretch}{1.25}
\caption{Summary of Knowledge Distillation Methods for Large Language Models (2020–2025). This table outlines key distillation techniques, detailing their skills, teacher and student models, objectives, and release timelines, categorized as \textbf{Black-box KD} (access limited to teacher outputs, typically from closed-source LLMs) and \textbf{White-box KD} (access to teacher parameters or distributions, usually from open-source LLMs). Metrics include IF (Instruction Following), CoT (Chain-of-Thought), ICL (In-context Learning), SFT (Supervised Fine-Tuning), D\&S (Divergence and Similarity), RL (Reinforcement Learning), TP (Think Pattern), NLU (Natural Language Understanding), and NLG (Natural Language Generation).}
\label{tab:kd}
\resizebox{\textwidth}{!}{%
\begin{tabular}{ccccccr}
\toprule
\textbf{Type} & \textbf{Methods} & \textbf{Skill} & \textbf{Teacher Model} & \textbf{Student Model} &\textbf{Objective} & \textbf{Release Time} \\
\midrule
\rowcolor{Gray} & SELF-INSTRUCT~\citep{wang2022self}& IF & T5-LM, GPT-3 & GPT-3 & SFT & Mar-2022\\
\rowcolor{LightGray} & MT-COT~\citep{li2022explanations} & CoT & GPT-3 & T5 & SFT &  Oct-2022\\
\rowcolor{Gray} & CoT Prompting~\citep{magister2022teaching}& CoT & GPT-3, PaLM & T5 & SFT & Dec-2022\\
\rowcolor{LightGray} & Fine-tune-CoT~\citep{ho2022large} & CoT & GPT-3 & T5, GPT-2, GPT-3 & SFT & Dec-2022\\
\rowcolor{Gray} & SOCRATIC CoT~\citep{shridhar2023distilling} & CoT & GPT-3 & GPT-2 & SFT & Dec-2022 \\
\rowcolor{LightGray} & ICL Distillation~\citep{huang2022context} & ICL & GPT-2, BERT & GPT-2, BERT & SFT & Dec-2022 \\
\rowcolor{Gray} & SSLM~\citep{fu2023specializing}& CoT & GPT-3.5 & T5 & SFT & Jan-2023\\
\rowcolor{LightGray} & LaMini-LM~\citep{wu2023lamini}& IF & ChatGPT & Various Models & SFT & Apr-2023\\ 

\rowcolor{Gray} &  SCOTT~\citep{wang2023scott}& CoT & GPT-neox & T5-3b & SFT & May-2023\\
\rowcolor{LightGray} & Distilling Step-by-Step~\citep{hsieh2023distilling} & CoT & PaLM & T5 & SFT  & May-2023\\
\rowcolor{Gray} & Lion~\citep{jiang2023lion} & IF & ChatGPT & LLaMA & - & May-2023\\
\rowcolor{LightGray} &  PaD~\citep{zhu2023pad} & CoT & GPT-3.5-turbo & CodeT5 & SFT & May-2023\\
\rowcolor{Gray} & AICD~\citep{liu2024learning}& ICL & GPT-3.5-turbo & GPT-J & SFT & Jun-2023\\
\rowcolor{LightGray} & KPTD~\citep{padmanabhan2023propagating} & NLU & GPT-3.5 & GPT-2, LLaMA & SFT & Jun-2023\\
\rowcolor{Gray} & DRA~\citep{wang2023democratizing} & CoT & ChatGPT & GPT-J & SFT & Oct-2023\\
\rowcolor{LightGray} & TDIG~\citep{li2024turning} & CoT & GPT-3.5-turbo, GPT-4 & LLaMA & SFT & Dec-2023\\
\rowcolor{Gray} & Selective Reflection-Tuning~\citep{li2024selective}& IF & ChatGPT & LLaMA & SFT & Feb-2024\\
\rowcolor{LightGray} & DEBATunE~\citep{li2024llmsspeakdiversepeople}& IF/TP & ChatGPT &  LLaMA & SFT + RL & Feb-2024\\
\rowcolor{Gray} & DeepSeek-R1~\citep{DeepSeekAI2025DeepSeekR1IR}& TP & DeepSeek-R1, DeepSeek-V3 &  LLaMA, Qwen & SFT & Jan-2025\\

\multirow{-21}{*}{\begin{tabular}[c]{@{}c@{}}\textbf{Black-box KD}\end{tabular}}   \\
\noalign{\vskip-1.20em}
\midrule

\rowcolor{Gray} & DynaBERT~\citep{hou2020dynabert} & NLU & BERT, RoBERTa & DynaBERT, DynaRoBERTa & SFT & Apr-2020\\
\rowcolor{LightGray} & TED~\citep{liang2023less} & NLU & GPT-2 & GPT-2 & D\&S + SFT & Oct-2022\\
\rowcolor{Gray} & GKD~\citep{agarwal2024policy}& NLG/NLU/IF & T5-XL & T5 & D\&S + RL & Jun-2023\\
\rowcolor{LightGray} & MINILLM~\citep{gu2023minillm} & IF & GPT-2, OPT, LLaMA & GPT-2, OPT, LLaMA & D\&S & Jun-2023\\
\rowcolor{Gray} & RICD~\citep{yang2024rlcdreinforcementlearningcontrastive}& IF & LLaMA &  LLaMA & SFT + RL & Jul-2023\\
\rowcolor{LightGray} & BabyLlama~\citep{timiryasov2023baby} & IF & GPT-2, LLaMA & LLaMA & D\&S & Aug-2023\\
\rowcolor{Gray} & MiniMoE~\citep{zhang2023towards} & NLU/TP & GPT-2, Pythia & GPT-GPT-2, Pythia & SFT & Nov-2023\\
\rowcolor{LightGray} & Genie~\citep{yehudai2024genie}& NLG& Falcon, LLaMA & FLAN, LLaMA & SFT & Jan-2024\\
\rowcolor{Gray} & Self-Rewarding~\citep{yuan2024self}& IF & LLaMA &  LLaMA & SFT + RL & Jan-2024\\
\rowcolor{LightGray} & DistiLLM~\citep{Yang2024RepresentationSF}& IF/NLG & GPT-2, OPT, OpenLLaMA &  GPT-2, OPT, OpenLLaMA & D\&S + RL & Feb-2024\\
\rowcolor{Gray} & AMMD~\citep{jia2024adversarialmomentmatchingdistillationlarge}& IF & OpenLLaMA &  OpenLLaMA & D\&S + RL & Jun-2024\\
\rowcolor{LightGray} & MultiLevelOT~\citep{cui2025multi} &  IF  & LLaMA2, Mistral, Qwen, LLaMA3 & OPT, Pythia, Bloomz, mT0 & D\&S + SFT  & Apr-2025\\

\multirow{-13}{*}{\begin{tabular}[c]{@{}c@{}}\textbf{White-box KD}\end{tabular}}   \\
\noalign{\vskip-1.20em}

\bottomrule
\end{tabular}
}
\end{table*}

KD is widely used in model compression for resource-limited settings and transfer learning, where a pre-trained teacher guides a task-specific student. Its effectiveness depends on factors such as teacher capacity, student architecture, and distillation loss design. Recent advances extend KD beyond output distillation, enabling more efficient and adaptable LLMs in post-training optimization.
KD methods can be broadly categorized into black-box KD and white-box KD, depending on the level of access to the teacher model’s internal parameters and intermediate representations. As shown in \textbf{Table~\ref{tab:kd}}, knowledge distillation methods can be broadly categorized into two types: Black-box KD and White-box KD. We provide a systematic summary of various knowledge distillation techniques in large language models (LLMs), along with their corresponding skills, teacher models, and student models.

\noindent\textbf{Black-box KD.}
Black-box KD refers to a scenario in which the student model learns solely from the teacher’s output logits, without access to its internal representations or architectural details. This approach, originally proposed by Hinton~\citep{hinton2015distillingknowledgeneuralnetwork}, aligns with the classical KD paradigm and is widely adopted due to its flexibility. A key advantage of black-box KD is that it treats the teacher model as an opaque function, enabling knowledge transfer even when the teacher is a proprietary or pre-trained model with restricted access.
In practice, large teacher LLMs (e.g., ChatGPT and GPT-4~\citep{achiam2023gpt}) are commonly employed to generate high-quality outputs. Meanwhile, smaller language models (SLMs), including GPT-2~\citep{radford2019language}, T5~\citep{raffel2023exploringlimitstransferlearning}, Flan-T5~\citep{chung2022scalinginstructionfinetunedlanguagemodels}, and CodeT5~\citep{wang2021codet5identifierawareunifiedpretrained}, serve as student models. These SLMs are optimized for efficiency while maintaining strong generalization capabilities, making them suitable for deployment in resource-constrained environments.


\noindent\textbf{White-box KD.}
White-box KD extends the traditional distillation paradigm by leveraging additional insights from the teacher's internal representations. This approach is beneficial when the architecture of the teacher model is known and accessible, allowing for richer forms of supervision.
Unlike black-box KD, which treats the teacher as an opaque function, white-box KD allows the student model to learn not only from the teacher's output logits but also from its intermediate activations, hidden layers, and potentially even attention weights~\citep{furlanello2018bornneuralnetworks}.



\noindent\textbf{DeepSeek-R1: Direct Distillation of Reasoning Patterns.}~~DeepSeek-R1 exemplifies the transformative potential of KD by distilling intricate reasoning patterns from large-scale models into compact architectures, significantly enhancing the reasoning capabilities of smaller LLMs without the computational burden of direct RL on such models. This approach, termed direct distillation, leverages a curated dataset of approximately 800k samples generated by a large teacher model, comprising 200k non-reasoning instances derived from DeepSeek-V3 and 600k reasoned instances produced by the DeepSeek-R1-Stage1 checkpoint. These samples form the foundation for SFT applied to open-source base models, such as Qwen and LLaMA mini-variants, enabling the student models to inherit sophisticated reasoning faculties typically reserved for their larger counterparts.

\begin{figure}[h]
    \centering
    \includegraphics[width=0.8\linewidth]{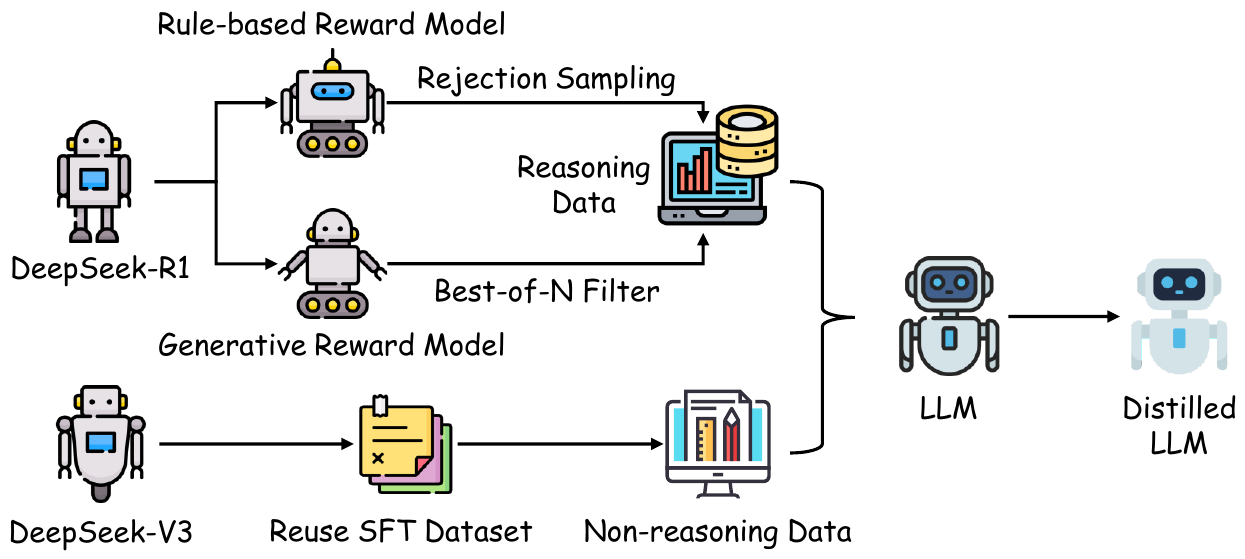}
    \caption{Workflow of knowledge distillation in DeepSeek-R1, illustrating the process of transferring reasoning patterns from large to compact models.}
    \label{fig:distill}
\end{figure}

The direct distillation process in DeepSeek-R1 unfolds in a structured pipeline, as depicted in \textbf{Fig.~\ref{fig:distill}}. Initially, the teacher model—pre-trained on extensive datasets—generates a diverse corpus encompassing both reasoning and non-reasoning outputs, capturing a spectrum of logical patterns and factual knowledge. The non-reasoning data (about 200k samples) provides a baseline of general knowledge, while the reasoning data (about 600k samples) encapsulates multi-step reasoning chains, refined through the teacher’s advanced capabilities. This dataset is then employed in an SFT phase, where the student model is trained to align its output distribution with that of the teacher, using reasoning data for direct fine-tuning of the smaller model to distill a compact inference model. Unlike traditional RL applied directly to small models, which may yield suboptimal reasoning due to limited capacity, DeepSeek-R1’s direct distillation circumvents such constraints by transferring pre-optimized reasoning behaviors, achieving superior performance with reduced resource demands.

A distinguishing feature of DeepSeek-R1’s KD methodology is its emphasis on preserving reasoning integrity across model scales. By integrating reasoned trajectories from DeepSeek-R1-Stage1—a checkpoint refined through large-scale RL—the student models not only replicate factual accuracy but also emulate complex inference processes, such as those required for mathematical problem-solving or logical deduction. This targeted transfer contrasts with conventional KD, which often prioritizes classification tasks, and underscores DeepSeek-R1’s innovation in reasoning-oriented distillation. Furthermore, the approach minimizes the need for extensive RL iterations on the student, leveraging the teacher’s pre-computed reasoning outputs to streamline training, thus enhancing efficiency and scalability. This methodology positions DeepSeek-R1 as a paradigm for distilling advanced reasoning into compact LLMs, offering a blueprint for future post-training optimization efforts.

\section{PoLMs for Integration and Adaptation}\label{Section 7}
Integration and adaptation techniques are pivotal for enhancing the versatility and efficacy of LLMs across diverse real-world applications. These methodologies enable LLMs to seamlessly process heterogeneous data types, adapt to specialized domains, and leverage multiple architectural strengths, thereby addressing complex, multifaceted challenges. This chapter delineates three principal strategies: \textbf{Multi-modal Integration} (\S \ref{Section 7.1}), which equips models to handle diverse data modalities such as text, images, and audio; \textbf{Domain Adaptation} (\S \ref{Section  7.2}), which refines models for specific industries or use cases; and \textbf{Model Merging} (\S \ref{Section 7.3}), which amalgamates capabilities from distinct models to optimize overall performance. Collectively, these approaches enhance LLMs’ adaptability, efficiency, and robustness, broadening their applicability across varied tasks and contexts.

\begin{figure}[h]
    \centering
    \includegraphics[width=0.6\linewidth]{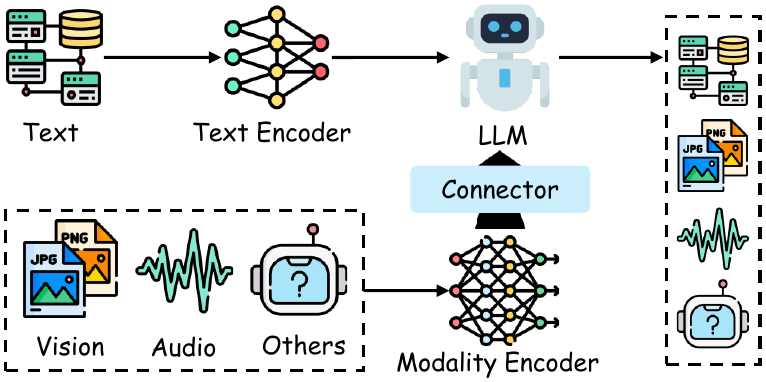}
    \caption{The architecture of a typical large multi-modal models.}
    \label{fig:Mutimodal_1}
\end{figure}

\subsection{Multi-Modal Integration}\label{Section 7.1}

Building upon the post-training optimization strategies elucidated in preceding chapters, this section examines advanced methodologies designed to augment LLMs and Large Multi-modal Models (LMMs) for effective processing of multi-modal data. While supervised fine-tuning enhances LLMs’ proficiency in task-specific contexts, its limitations in harnessing the full spectrum of multi-modal capabilities necessitate more sophisticated post-training approaches. These techniques enable LMMs to address complex, cross-modal tasks (e.g., generating web page code from visual inputs~\citep{zhu2023minigpt}, interpreting nuanced cultural artifacts like memes~\citep{yang2023mm}, and performing mathematical reasoning without reliance on optical character recognition~\citep{driess2023palm}), by integrating diverse data types into a unified framework. Typically, LMMs comprise a modal encoder, a pre-trained LLM backbone, and a modal connector~\citep{yin2023survey}, as depicted in \textbf{Fig. \ref{fig:Mutimodal_1}}. This architecture forms the foundation for post-training methods that refine each component, facilitating robust multi-modal integration and performance enhancement.

\subsubsection{Modal Connection}
\begin{figure}[h]
    \centering
    \includegraphics[width=0.8\linewidth]{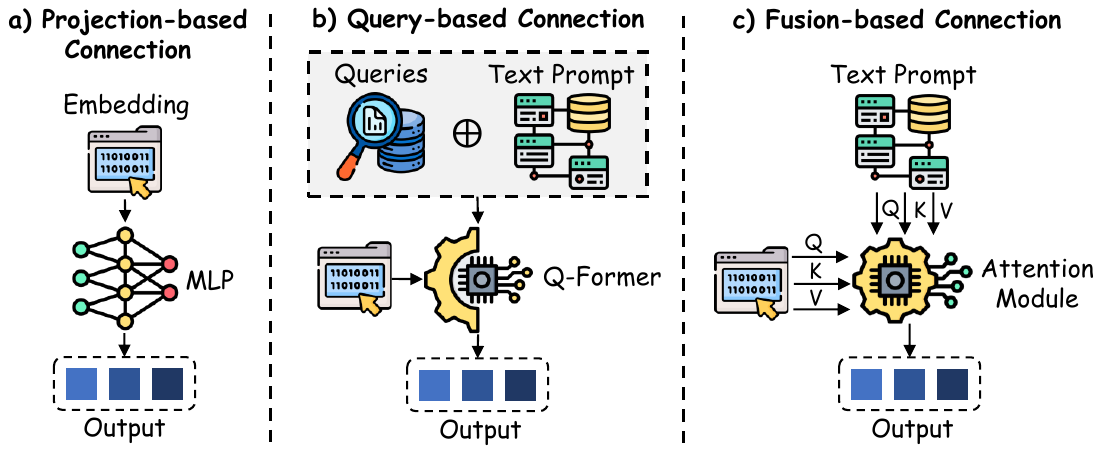}
    \caption{Taxonomy of modal connection methods in multi-modal integration, delineating projection-based, query-based, and fusion-based approaches.}
    \label{fig:mutimodal}
\end{figure}

Modal connection methods are pivotal in synthesizing multi-modal data into a coherent representational framework, categorized into three primary strategies: projection-based, query-based, and fusion-based approaches~\citep{yin2023survey}, as outlined in \textbf{Fig. \ref{fig:mutimodal}}. 

\noindent\textbf{Projection-based Modal Connection.}~~Projection-based methods transform diverse modal inputs into a unified text embedding space, aligning their features with the linguistic dimensions of LLMs for seamless integration. LLaMA-Adapter~\citep{zhang2023llama} exemplifies this approach by incorporating an image encoder to extend LLMs into multi-modal systems, enabling image-conditioned instruction tracking. Its successor, LLaMA-Adapter V2~\citep{gao2023llama}, refines this process by embedding visual tags into early LLM layers, fostering improved assimilation of visual knowledge. FROMAGe~\citep{koh2023grounding} employs fine-tuning of input and output layers within a frozen LLM and visual encoder framework to enable cross-modal interactions, while LLaVA-1.5~\citep{liu2024improved} utilizes a bilinear multilayer perceptron (MLP) to bolster robustness in multi-modal processing. Recent developments, such as Shikra~\citep{chen2306shikra}, integrate spatial coordinates to enhance natural language dialogues, and VILA~\citep{lin2024vila} optimizes visual-language pre-training for superior zero-shot capabilities. DetGPT~\citep{pi2023detgpt} further advances this paradigm by merging reasoning-driven object detection with natural language interaction, leveraging projection techniques to facilitate effective multi-modal communication. SOLO~\cite{singletransformer2024} employs a single Transformer architecture for unified and end-to-end vision-language modeling, by accepting both raw image patches (in pixels) and texts as inputs, without using a separate pre-trained vision encoder. 
Meanwhile, MiniGPT-4~\citep{zhu2023minigpt} aligns a frozen visual encoder with Vicuna using a single projection layer, achieving GPT-4-like abilities with a two-stage training process. Idefics~\citep{laurenccon2025matters} excels with an autoregressive design and multi-stage pretraining for efficient inference. LaVIT~\citep{jin2024unified} unifies vision and language with a discrete visual tokenizer for seamless generation. DeepSeek-VL2~\citep{wu2024deepseek} enhances high-resolution image understanding with dynamic tiling and multi-head latent attention. Finally, Qwen2.5-VL~\citep{bai2025qwen25vltechnicalreport} advances multi-modal tasks with a redesigned Vision Transformer, excelling in perception and video comprehension.

\noindent\textbf{Query-based Modal Connection.}~~Query-based methods enhance multi-modal integration by employing learnable query tokens to extract structured information from diverse modalities, bridging the gap between textual and non-textual data. BLIP-2~\citep{li2023blip} pioneered this approach with query transformers, integrating text and visual inputs efficiently. Video-LLaMA~\citep{zhang2023video} extends this technique to video comprehension through combined visual encoders, while InstructBLIP~\citep{dai2023instruct} refines query mechanisms to ensure precise adherence to instructions. X-LLM~\citep{chen2023x} aligns multi-modal inputs via specialized interfaces, and subsequent innovations like mPLUG-Owl~\citep{ye2023mplug} and Qwen-VL~\citep{bai2023qwenvl} optimize the Q-Former architecture for computational efficiency. LION~\citep{chen2024lion} further demonstrates the efficacy of query-based methods by advancing visual knowledge integration, underscoring their utility in enhancing LMM performance across varied tasks.
Qwen-VL~\citep{bai2023qwenvl} is a series of large-scale vision-language models built upon Qwen-7B, incorporating a visual receptor, a position-aware adapter, and a three-stage training pipeline to enable multilingual, fine-grained vision-language understanding.
Lyrics~\citep{lu2023lyrics} is a fine-grained vision-language pre-training and instruction fine-tuning framework that enhances large vision-language models (LVLMs) by integrating semantic-aware visual objects through a visual refiner (image tagging, object detection, and semantic segmentation) and a Multi-scale Querying Transformer (MQ-Former).

\noindent\textbf{Fusion-based Modal Connection.}~~Fusion-based techniques deepen cross-modal interactions by embedding multi-modal features directly into the LLM architecture, fostering richer integration at the inference level. Flamingo~\citep{alayrac2022flamingo} employs cross-attention layers to fuse visual features during token prediction, enabling dynamic multi-modal processing. OpenFlamingo~\citep{awadalla2023openflamingo} builds on this by allowing frozen LLMs to attend to vision encoder outputs, enhancing flexibility. Otter~\citep{li2023otter} introduces instruction tuning to improve multi-modal instruction-following, while CogVLM~\citep{wang2023cogvlm} integrates visual expert modules within Transformer layers for seamless feature synthesis. Obelics~\citep{laurenccon2024obelics} leverages interleaved image-text training data, highlighting the robustness of fusion-based methods in achieving cohesive multi-modal performance.
InternVL~\citep{chen2023internvl} is a large-scale vision-language foundation model that scales up the vision encoder to 6 billion parameters and progressively aligns it with LLMs using a language middleware (QLLaMA).
Llama 3~\citep{dubey2024llama} is a new family of multilingual, tool-using foundation models developed by Meta, scaling up to 405B parameters with a 128K token context window, optimized through improved data quality, larger-scale training, and a structured post-training strategy.

\begin{table*}[ht]
\renewcommand{\arraystretch}{1.2}
\caption{Overview of Encoders and Large Multi-modal Models Across Modalities (2022–2025). This table summarizes key multi-modal models, detailing their encoder categories, sizes, input projectors, LLM backbones, and release timelines across vision, audio, and other modalities. Metrics include \textbf{C-a} (Cross-attention), \textbf{Q-F} (Q-Former), \textbf{MQ-F} (Multi-Query Q-Former), and \textbf{LP} (Linear Projector), representing input projection mechanisms.}
\scriptsize
\centering
\resizebox{1\textwidth}{!}{
\begin{tabular}{ccccccr}
\toprule
\multirow{2}{*}{\textbf{Modal}}  & \multirow{2}{*}{\textbf{Model}} & \multicolumn{2}{c}{\textbf{Encoder}}  & \multirow{2}{*}{\textbf{Input Projector}} & \multirow{2}{*}{\textbf{LLM Backbone}} & \multirow{2}{*}{\textbf{Release Time}} \\ \cline{3-4}  
 &  & \textbf{Category}   & \textbf{Size} &   \\ \midrule
 
\rowcolor{Gray}& \textbf{Flamingo}~\citep{alayrac2022flamingo} & NFNet & 0.3B & C-a & Chinchilla-1.4B/7B/70B & Apr-2022\\
\rowcolor{LightGray}& \textbf{BLIP-2}~\citep{li2023blip} & CLIP/Eva & 0.3B/1.5B & Q-F/LP & Flan-T5/OPT & Jan-2023\\
\rowcolor{Gray}& \textbf{MiniGPT-4}~\citep{zhu2023minigpt} & Eva & 1.5B & Q-F/LP & Vicuna-13B & Apr-2023\\
\rowcolor{LightGray}& \textbf{VideoChat}~\citep{li2023videochat}    & Eva   & 1.5B  & Q-F/LP        & StableVicuna-13B & May-2023\\
\rowcolor{Gray}& \textbf{InstructBLIP}~\citep{dai2023instruct} & ViT & 1.5B & Q-F/LP & Flan-T5/Vicuna & May-2023\\
\rowcolor{LightGray}& \textbf{Video-ChatGPT}~\citep{maaz2023video} & CLIP & 0.3B & LP & Vicuna-v1.1 & May-2023\\
\rowcolor{Gray}& \textbf{IDEFICS}~\citep{laurenccon2025matters} & OpenCLIP & 0.3B  & C-a & LLaMA & May-2023\\
\rowcolor{LightGray}& \textbf{Qwen-VL-(Chat)}~\citep{bai2023qwenvl} & OpenCLIP & 1.8B  & C-a & Qwen-7B & Aug-2023\\
\rowcolor{Gray}& \textbf{LLaVA}~\citep{Liu2023VisualIT}          & CLIP    & 0.3B    & LP         & Vicuna-13B   & Oct-2023\\
\rowcolor{LightGray}& \textbf{Lyrics}~\citep{lu2023lyrics} & CLIP & 0.1B$\sim$0.3B & MQ-F/LP & Vicuna-13B & Oct-2023\\
\rowcolor{Gray}& \textbf{CogVLM}~\citep{wang2023cogvlm} & Eva & 1.5B & MLP & Vicuna-v1.5-7B & Nov-2023\\
\rowcolor{LightGray}& \textbf{BT-Adapter}~\citep{liu2023one}   & CLIP & 0.3B   & LP        & Vicuna-7B     & Nov-2023\\
\rowcolor{Gray}& \textbf{SPHINX-Tiny}~\citep{gao2024sphinx}    & DINOv2        & 1.0B    & LP       & TinyLlama-1.1B  & Jan-2024\\           
\rowcolor{LightGray}& \textbf{VL-Mamba}~\citep{qiao2024vl}          & SigLIP       & 0.4B & MLP    & Mamba-2.8B-Slimpj   & Feb-2024\\
\rowcolor{Gray}& \textbf{Mipha}~\citep{zhu2024comprehensive}    & SigLIP     & 0.4B    & MLP    & Phi-2-2.7B      & Mar-2024\\
\rowcolor{LightGray}& \textbf{IntrenVL}~\citep{chen2023internvl} & InternImage & 6.0B & C-a/MLP & Qwen-Llama-8B & Mar-2024\\
\rowcolor{Gray}& \textbf{Cobra}~\citep{zhao2024cobra}          & SigLIP/DINOv2 & 0.4B  & MLP & Mamba-2.8B-Zephyr & Apr-2024\\
\rowcolor{LightGray}& \textbf{LaVIT}~\citep{jin2024unified} & ViT & 0.1B$\sim$0.3B  & C-a & LLaMA-7B & May-2024\\
\rowcolor{Gray}& \textbf{DeepSeekVL2}~\citep{wu2024deepseek} & Dynamic Tiling & 2.8B &MLP& DeepSeek-MoE-16B & May-2024 \\
\rowcolor{LightGray}& \textbf{VILA}~\citep{lin2024vila} & ViT & 0.1B$\sim$0.3B  & LP & LLaMA-2-7B/13B & Jun-2024\\
\rowcolor{Gray}& \textbf{Llama 3}~\citep{dubey2024llama} &SigLIP& 1.5B & LP & LLaMA-3 & Jul-2024 \\
\rowcolor{LightGray}& \textbf{QVQ}~\citep{qvq} &ViT & 0.6B  &C-a & Qwen2-VL-72B & Dec-2024 \\
\rowcolor{Gray}& \textbf{Qwen2.5-VL}~\citep{bai2025qwen25vltechnicalreport} &ViT & 0.8B  &MLP& Qwen-MoE-A2.7B & Feb-2025 \\
\rowcolor{LightGray}& \textbf{Claude3.7}~\citep{claude37} &- & - &-& Claude3.7 & Feb-2025 \\
\rowcolor{Gray}& \textbf{GPT-4.5}~\citep{openai2025gpt45} &- &-  &- &GPT-4.5 & Feb-2025 \\
\rowcolor{LightGray}& \textbf{Kimi-VL}~\citep{team2025kimi} & ViT & 0.4B & MLP & Moonlight-MoE-A2.8B & Apr-2025 \\
\multirow{-24}{*}{\begin{tabular}[c]{@{}c@{}}\textbf{Vision}\end{tabular}}   \\
\noalign{\vskip-1.20em}
\midrule
 
\rowcolor{Gray}& \textbf{AudioPaLM}~\citep{rubenstein2023audiopalm} & AudioMAE  & 0.25B  & LP    & PaLM-2-8B & Sep-2023\\
\rowcolor{LightGray}& \textbf{Qwen-Audio}~\citep{bai2023qwenvl} & Whisper-L-v2 & 1.5B & LP & Qwen-7B & Nov-2023\\
\rowcolor{Gray}& \textbf{SpeechGPT}~\citep{zhang2023speechgpt}  & Transformer  & 0.1B$\sim$0.3B  & LP & LLaMA-13B & Dec-2023\\
\rowcolor{LightGray}& \textbf{VoiceCraft}~\citep{peng2024voicecraft} & Neural-Codec & 0.33B/0.83B & LP & Mamba-2.8B & Mar-2024 \\
\multirow{-6}{*}{\begin{tabular}[c]{@{}c@{}}\textbf{Audio}\end{tabular}}   \\
\noalign{\vskip-1.20em}            
\midrule

\rowcolor{Gray}& \textbf{X-LLM}~\citep{chen2023x}              & Transformer      & 0.1B$\sim$0.3B  & Q-F/LP  & ChatGLM-6B  & May-2023\\
\rowcolor{LightGray}& \textbf{Video-LLaMA}~\citep{zhang2023video} & Eva-CLIP & 1.5B/0.17B & Q-F/LP & Vicuna/LLaMA & Jun-2023\\
\rowcolor{Gray}& \textbf{ImageBind-LLM}~\citep{han2023imagebind}   & ImageBind  & 0.17B  & LP  & LLaMA-7B & Jun-2023\\
\rowcolor{LightGray}& \textbf{AnyMAL}~\citep{moon2024anymal} & CLIP/CLAP & 0.3B & C-a/LP  & LLaMA-2 & Sep-2023\\
\rowcolor{Gray}& \textbf{NEXT-GPT}~\citep{wu2023next}          & ImageBind  & 0.17B      & LP    & Vicuna-7B & Oct-2023\\ 
\rowcolor{LightGray}& \textbf{CoDi-2}~\citep{tang2024codi} & ImageBind & 0.17B & MLP & LLaMA-2-Chat-7B & Jan-2024\\
\rowcolor{Gray}& \textbf{LL3DA}~\citep{chen2024ll3da}           & Transformer     & 0.1B$\sim$0.3B  & LP  & OPT-1.3B  & Feb-2024\\
\rowcolor{LightGray}& \textbf{X-InstructBLIP}~\citep{panagopoulou2023x} & Eva/BEATs/ULIP-2 & 1.5B & Q-F/LP & Vicuna-v1.1-7B/13B & Dec-2024\\
\multirow{-10}{*}{\begin{tabular}[c]{@{}c@{}}\textbf{Others}\end{tabular}}   \\
\noalign{\vskip-1.20em}
\bottomrule
\end{tabular}}
\label{tab:multi}
\end{table*}

\subsubsection{Modal Encoder}\label{sub:encoder}
Modal encoders compress raw multi-modal inputs into compact, semantically rich representations, enabling efficient processing across diverse tasks and modalities. These components are essential for translating heterogeneous data into a format compatible with LLM backbones, supporting applications from visual reasoning to audio comprehension. \textbf{Table \ref{tab:multi}} provides a comprehensive summary of prevalent encoders utilized in vision, audio, and other modalities, detailing their characteristics and contributions to multi-modal integration.

\noindent\textbf{Vision Encoder.}~~Vision encoders are foundational to multi-modal learning, facilitating the interpretation and generation of visual data within LMMs. CLIP~\citep{radford2021learning} establishes joint image-text representations through contrastive learning, enhancing cross-modal alignment. EVA~\citep{fang2023eva} refines visual attention mechanisms to improve efficiency, while ImageBind~\citep{Girdhar_2023_CVPR} creates a unified embedding space across multiple modalities, elevating zero-shot recognition capabilities. SigLIP~\citep{zhai2023sigmoid} introduces a paired sigmoid loss to optimize image-text pre-training, and DINOv2~\citep{oquab2023dinov2} employs unsupervised learning to derive robust visual features from diverse sources. LLaVA~\citep{Liu2023VisualIT} adapts self-instruct strategies to convert images into textual descriptions, generating novel datasets with advanced LLMs. Video-ChatGPT~\citep{maaz2023video} supports conversational video understanding with large-scale instruction datasets, while BT-Adapter~\citep{liu2023one} optimizes video comprehension through efficient temporal modeling. VideoChat~\citep{li2023videochat} focuses on spatiotemporal reasoning, leveraging specialized datasets, and models like CoDi-2~\citep{tang2024codi} and Mipha~\citep{zhu2024comprehensive} achieve efficiency gains in multi-modal processing. VL-Mamba~\citep{qiao2024vl} and Cobra~\citep{zhao2024cobra} introduce state-space models for optimized inference, and SPHINX-Tiny~\citep{gao2024sphinx} emphasizes data diversity and training efficiency.

\noindent\textbf{Audio Encoder.}~~Audio encoders enhance LMMs’ capacity to process and interpret auditory inputs, broadening their multi-modal scope. SpeechGPT~\citep{zhang2023speechgpt} integrates large-scale speech datasets with convolutional and transformer architectures~\citep{dosovitskiy2020image} to achieve robust instruction-following capabilities. AudioPaLM~\citep{rubenstein2023audiopalm} combines text and speech processing using the Universal Speech Model (USM) encoder~\citep{zhang2023google}, excelling in tasks like zero-shot language translation. WavCaps~\citep{10572302} employs CNN14~\citep{kong2020panns} and HTSAT~\citep{chen2022hts} to mitigate audio-language data scarcity, utilizing advanced LLMs to refine dataset quality and bolster learning outcomes, underscoring the critical role of audio modalities in multi-modal systems.

\noindent\textbf{Other Encoder.}~~Beyond vision and audio, encoders for additional modalities, such as 3D understanding and multi-modal fusion, are integral to comprehensive LMMs. NEXT-GPT~\citep{wu2023next} facilitates cross-modal content generation across text, images, video, and audio, advancing human-like AI capabilities with minimal parameter adjustments. ImageBind-LLM~\citep{han2023imagebind} aligns visual and linguistic embeddings to improve instruction-following across modalities. LL3DA~\citep{chen2024ll3da} processes point cloud data for 3D reasoning and planning, introducing novel approaches to spatial understanding. X-LLM~\citep{chen2023x} employs Q-Former~\citep{li2023blip} for image and video inputs and C-Former~\citep{chen2023x} for speech, compressing audio features into token-level embeddings to enhance multi-modal learning efficiency.

\subsection{Domain Adaptation}\label{Section 7.2}
\definecolor{hidden-draw}{RGB}{106,142,189} 
\definecolor{hidden-blue}{RGB}{194,230,247} 
\definecolor{hidden-orange}{RGB}{217, 230, 252} 

Domain Adaptation (DA) constitutes a pivotal post-training strategy for refining LLMs to excel within specialized domains, ensuring their efficacy in targeted applications. Rooted in the principles of transfer learning~\citep{kouw2018introduction,DASurvey}, DA transforms an initial model, denoted \( M_{\text{source}} \), through an adaptation function \( F_{\text{adapt}} \) to produce a domain-specific model \( M_{\text{target}} \), as depicted:

\begin{equation}
\label{eq:07-merge-overview}
\resizebox{0.4\textwidth}{!}{
    \includegraphics[width=\columnwidth]{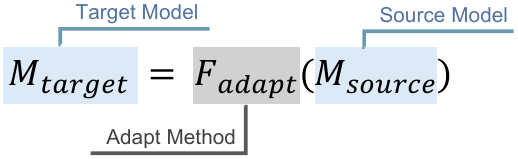}
}
\end{equation}
This process tailors \( M_{\text{target}} \) to address the unique demands and intricacies of a designated domain, thereby optimizing its performance and relevance. By enhancing LLMs’ proficiency in fields such as programming~\citep{codellama,codegen} and mathematical reasoning~\citep{imani2023mathprompter}, DA not only elevates domain-specific capabilities but also improves computational efficiency, mitigating the limitations of general-purpose models that often struggle with domain-specific terminologies and reasoning paradigms. Furthermore, DA substantially reduces the reliance on extensive labeled datasets and computational resources typically required for training domain-specialized models from scratch~\citep{wu2024continual}, rendering it a cornerstone of post-training methodologies.

\subsubsection{Knowledge Editing}
Knowledge Editing represents a sophisticated post-training approach aimed at modifying LLMs to meet domain-specific requirements without compromising their foundational capabilities. This technique facilitates targeted parameter adjustments, preserving the model’s pre-existing performance while integrating new or updated domain knowledge~\citep{zhang2024comprehensive}. By enabling rapid adaptation to evolving knowledge landscapes, Knowledge Editing emerges as an indispensable component of post-training pipelines. An overview of principal methods (e.g., encompassing external knowledge utilization, integration, and intrinsic editing) is presented in \textbf{Table~\ref{tab:conceptual_analysis}}.

\begin{table*}[h]
    \caption{A comparative analysis of representative approaches for knowledge editing in LLMs. 
    \textbf{Edit Area} specifies the components of the model that are targeted for modification; 
    \textbf{Editor \#Params} indicates the parameters that require updating during the editing process. 
    $L$ represents the number of layers subjected to modification, 
    $d_h$ denotes the dimensionality of the hidden layers within the transformer architecture, 
    $d_m$ refers to the intermediate dimension that exists between the up-projection and down-projection phases, 
    and $N$ symbolizes the total number of neurons that undergo updates within each individual layer.
    }
    \label{tab:conceptual_analysis}
    \scriptsize
    \centering
    \renewcommand{\arraystretch}{1.1}
    \resizebox{1\textwidth}{!}{
    \begin{tabular}{ccccc}
    \toprule 
    \textbf{Category} & \textbf{Method} & \textbf{Edit Area} & \textbf{Edit Function} & \textbf{\makecell{Edited\\ \#Params}} \\
    \midrule
    
    \rowcolor{Gray}& SERAC \citep{mitchell2022memory}  & memory+classifier+auxiliary model & $\text{Output} \rightarrow \text{Model}_{cf}(\boldsymbol{x})$ & - \\
    \rowcolor{LightGray}& PokeMQA~\cite{gu2023pokemqa} &  memory+retriever & $\text{Input} \rightarrow [\text{Mem}:\text{Input}]$ & - \\
    \multirow{-4}{*}{\begin{tabular}[c]{@{}c@{}}Identification\end{tabular}}   \\
    \noalign{\vskip-1.20em}  
    \midrule
      
    \rowcolor{Gray}& CaliNET \citep{dong2022calibrating}  & FFN+params & $\boldsymbol{h} \rightarrow \boldsymbol{h} + \text{FFN}_{\text{add}}(\boldsymbol{x})$ & $ N \times d_h$\\ 
    \rowcolor{LightGray}& Transformer-Patcher\citep{huang2023transformer}  & FFN+params & $\boldsymbol{h} \rightarrow \boldsymbol{h} + \text{FFN}_{\text{add}}(\boldsymbol{x})$  & $ N \times d_h$\\
    \rowcolor{Gray}& REMEDI \citep{hernandez2023inspecting}& auxiliary model & $\boldsymbol{h} \rightarrow \text{REMEDI}(\boldsymbol{x})$  & $ d_h \times d_h$\\
    \rowcolor{LightGray}& GRACE \citep{hartvigsen2024aging} & FFN+codebook & $\boldsymbol{h} \rightarrow \text{GRACE}(\boldsymbol{x})$ & $ N \times 2d_h$ \\
    \rowcolor{Gray}& Eva-KELLM \citep{wu2023eva} & Attn or FFN &$\boldsymbol{h} \rightarrow \boldsymbol{h} + s \cdot \text{LoRA}(\boldsymbol{x})$  & $2L \times 2d_{am}d_h$\\
    \rowcolor{LightGray}& MELO \citep{yu2024melo} & Attn or FFN &$\boldsymbol{h} \rightarrow \boldsymbol{h} + s \cdot \text{LoRA}(\boldsymbol{x})$  & $2L \times 2d_{am}d_h$\\
    \multirow{-8}{*}{\begin{tabular}[c]{@{}c@{}}Association\end{tabular}}   \\
    \noalign{\vskip-1.20em} 
    \midrule
        
    \rowcolor{Gray}& Constrained Fine-tuning \citep{zhu2020modifying} & Any & $\boldsymbol{W} \rightarrow \boldsymbol{W}^{'}$ & $2\times L \times d_md_h$ \\
    \rowcolor{LightGray}& Editable Training \citep{sinitsin2020editable} & Any & $\boldsymbol{W} \rightarrow \boldsymbol{W}^{'}$ & $2\times L \times d_md_h$ \\
    \rowcolor{Gray}& KnowledgeEditor\cite{de2021editing}  & Attn or FFN + auxiliary model  &  $\boldsymbol{W} \rightarrow \boldsymbol{W}^{'}$& $2\times L \times d_md_h$ \\
    \rowcolor{LightGray}& SLAG \citep{hase2023methods}  & Attn or FFN + auxiliary model & $\boldsymbol{W} \rightarrow \boldsymbol{W}^{'}$ & $2\times L \times d_md_h$  \\
    \rowcolor{Gray} & MEND \citep{mitchell2021fast}  & FFN + auxiliary model  & $\boldsymbol{W} \rightarrow \boldsymbol{W}^{'}$ & $2\times L \times d_md_h$ \\
    \rowcolor{LightGray}& MALMEN \citep{tan2023massive} & FFN & $\boldsymbol{W}_{\text{down}} \rightarrow \boldsymbol{W}_{\text{down}}^{'}$ & $L \times d_md_h$\\
    \rowcolor{Gray} &  LLM Surgery \citep{veldanda2024llm} & Any & $\boldsymbol{W} \rightarrow \boldsymbol{W}^{'}$ & $2\times L \times d_md_h$\\
    \rowcolor{LightGray} & KNE~\citep{li2024knowledge} & subset of parameters & $\boldsymbol{W} \rightarrow \boldsymbol{W}^{'}$ & $ L \times d_{m} d_{h}$\\
    \rowcolor{Gray} & OVERTONE~\citep{liu2025mitigating} & Any & $\boldsymbol{W} \rightarrow \boldsymbol{W}^{'}$ & $2 \times L \times d_m d_h$\\
    \multirow{-13}{*}{\begin{tabular}[c]{@{}c@{}}Editing\end{tabular}}   \\
    \noalign{\vskip-1.20em} 
    \bottomrule
    \end{tabular}
    }
\end{table*}

\noindent\textbf{Formal Definition of Knowledge Editing.}~~Consider an original LLM parameterized by \(\theta\), pre-trained on a dataset \(\mathcal{D}_{\text{old}}\). Let \(\mathcal{D}_{\text{new}}\) denote a dataset containing novel or updated information \(\Delta K\). The objective of Knowledge Editing is to derive a revised parameter set \(\theta'\) by applying an adjustment \(\Delta \theta\), effectively assimilating \(\Delta K\) while minimizing degradation on \(\mathcal{D}_{\text{old}}\). Formally, this is framed as a constrained optimization problem, where the updated parameters are defined as:

\begin{equation}
    \theta' = \theta + \Delta \theta,
    \text{where} ~ \mathcal{L}\bigl(\theta'; \mathcal{D}_{\text{new}}\bigr) \;\rightarrow\; \min,
\end{equation}
with \(\mathcal{L}\) representing a loss function (e.g., cross-entropy) assessing model quality on \(\mathcal{D}_{\text{new}}\). To safeguard performance on the original dataset, a constraint is imposed:

\begin{equation}
    \mathcal{L}\bigl(\theta'; \mathcal{D}_{\text{old}}\bigr) 
    \;\leq\;
    \mathcal{L}\bigl(\theta; \mathcal{D}_{\text{old}}\bigr)
    \;+\;
    \epsilon,
\end{equation}
where \(\epsilon\) is a small positive constant limiting performance loss on \(\mathcal{D}_{\text{old}}\). This formulation ensures that \(\theta'\) incorporates \(\Delta K\) while retaining the model’s prior knowledge base. Practically, \(\Delta \theta\) may be constrained to specific architectural components (e.g., attention layers (\(\mathrm{Attn}\)) or feed-forward networks (\(\mathrm{FFN}\))), reducing computational overhead and preserving core functionality by avoiding comprehensive retraining.

\noindent\textbf{Knowledge Identification.}~~The initial phase of Knowledge Editing focuses on detecting and assimilating new information into the model. PokeMQA~\citep{gu2023pokemqa} employs a programmable scope detector and knowledge prompts to dissect queries, retrieving pertinent facts efficiently. Conversely, SERAC~\citep{mitchell2022memory} integrates a counterfact model with a classifier to determine the applicability of new knowledge sources, offering a minimally invasive approach that preserves the base model’s integrity without necessitating extensive structural modifications. ~\cite{ripple2024} analyzes the reasons why LLM knowledge updating creates messy ripple effect. Real world edits usually originate from emerging events that encompass logical connections between new facts and past facts, based on this observation, EvEdit~\cite{eventedit2024} proposes an event-based knowledge editing method to determine knowledge anchor and knowledge updating boundary.

\noindent\textbf{Knowledge Association.}~~Following identification, this phase associates newly acquired information with the model’s existing knowledge framework. Transformer-Patcher~\citep{huang2023transformer} adapts transformer architectures to integrate updated facts, while CaliNET~\citep{dong2022calibrating} recalibrates parameters to align with factual content. Methods such as Eva-KELLM~\citep{wu2023eva}, MELO~\citep{yu2024melo}, and REMEDI~\citep{hernandez2023inspecting} refine specific behaviors for precise updates, and GRACE~\citep{hartvigsen2024aging} enhances predictive accuracy post-knowledge insertion, ensuring seamless integration with prior representations.

\noindent\textbf{Intrinsic Knowledge Editing.}~~The final stage embeds associated facts into the model’s internal structure, ensuring comprehensive assimilation. While traditional fine-tuning can be resource-intensive, advanced techniques mitigate this burden. Constrained Fine-tuning~\citep{zhu2020modifying} and Meta-learning~\citep{de2021editing} minimize knowledge loss and overfitting risks. Editable Training~\citep{sinitsin2020editable} and KnowledgeEditor~\citep{de2021editing} enable swift parameter adjustments with minimal performance impact, whereas SLAG~\citep{hase2023methods}, MEND~\citep{mitchell2021fast}, and MALMEN~\citep{tan2023massive} resolve edit conflicts and support large-scale updates, maintaining foundational competencies while incorporating new domain insights.
LLM Surgery \citep{veldanda2024llm} unifies unlearning and editing by applying reverse gradients to remove outdated data, gradient descent to integrate new facts, and a KL divergence term to preserve existing knowledge, achieving significant computational efficiency. 
KNE \citep{li2024knowledge} introduces a Knowledge Neuronal Ensemble method that pinpoints and updates only those neurons strongly associated with newly inserted facts, achieving more accurate edits while preserving unrelated knowledge.
OVERTONE \citep{liu2025mitigating} tackles heterogeneous token overfitting in knowledge editing by introducing a token-level smoothing technique that adaptively refines the training objective, thereby preserving pre-trained knowledge and improving the model’s ability to reason about newly inserted facts.
These targeted techniques ensure that the model retains its foundational competencies while integrating newly acquired information.

\subsubsection{Retrieval-Augmented Generation}
Retrieval-Augmented Generation (RAG) integrates traditional information retrieval with contemporary LLMs to enhance the relevance and factual accuracy of generated outputs~\citep{lewis2020retrieval,gao2023retrieval,zhao2024retrieval}. By dynamically retrieving pertinent information from external sources and embedding it into the generation process, RAG addresses deficiencies in LLMs’ domain-specific knowledge and reduces the propensity for hallucinated content. This approach proves particularly effective in domains requiring precise, up-to-date information, such as question-answering systems~\citep{lewis2020retrieval}, scientific research~\citep{zhang2023optimizing}, and healthcare~\citep{xiong2024benchmarking}, where it adeptly handles complex queries and knowledge-intensive tasks. Moreover, RAG mitigates the prevalence of misleading responses in conversational systems, advancing the fidelity of knowledge-driven natural language generation~\citep{xiong2024benchmarking,zhang2023enhancing}.

\begin{figure}[h]
    \centering
    \includegraphics[width=0.8\linewidth]{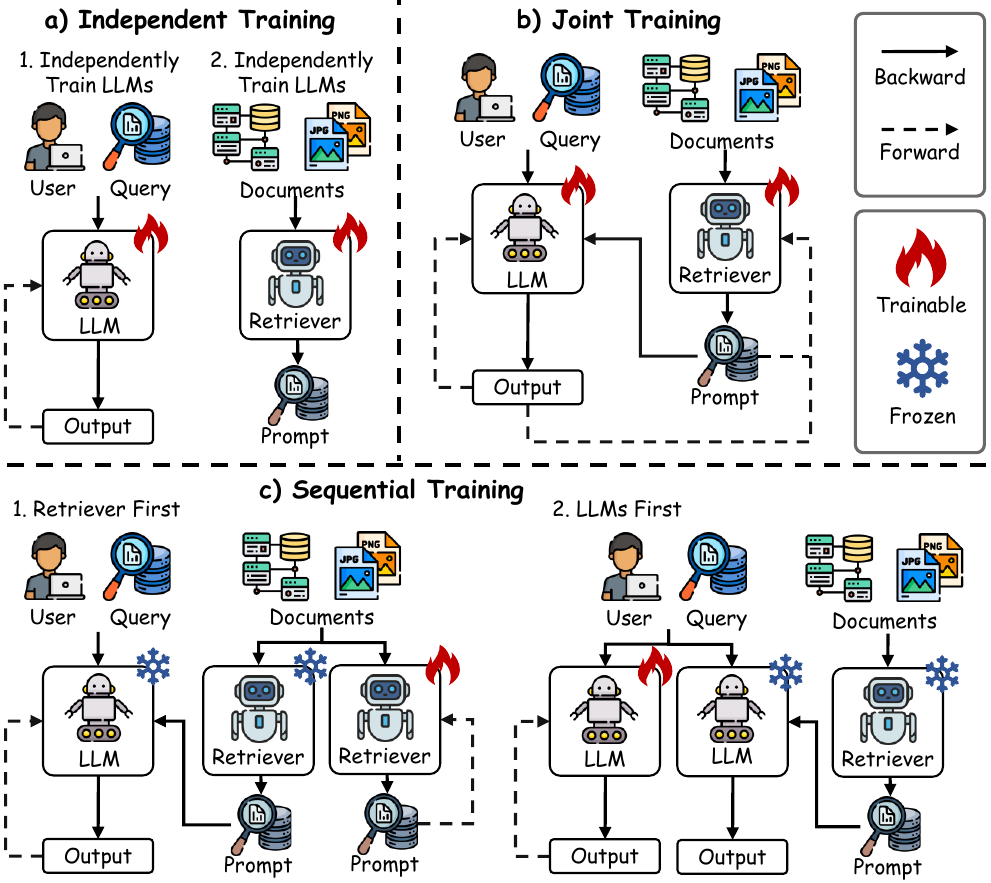}
    \caption{Taxonomy of Retrieval-Augmented Generation (RAG) training methods, categorizing Independent Training, Sequential Training, and Joint Training strategies.}
    \label{fig:07-rag-type}
\end{figure}

This subsection focuses on training-based RAG methodologies~\citep{fan2024survey}, recognizing that training-free RAG approaches~\citep{izacard2020leveraging,jiang2023active,gutierrez2024hipporag} may compromise knowledge utilization efficiency due to the absence of task-specific optimization. Three predominant training strategies—Independent Training, Sequential Training, and Joint Training—enhance model adaptability and integration proficiency, as illustrated in \textbf{Fig. \ref{fig:07-rag-type}}.

\noindent\textbf{Independent Training.}~~This strategy trains the retriever and generator as distinct modules, enabling flexibility in employing sparse or dense retrievers tailored to task demands. DPR~\citep{karpukhin2020dense}, for instance, utilizes dual BERT networks to encode queries and paragraphs separately, applying contrastive learning to optimize retrieval without generator interaction. Likewise, \citep{nguyen2024reward} propose Reward-RAG, which leverages a reward model to fine-tune only the retriever according to GPT-based feedback, leaving the generator untouched.

\noindent\textbf{Sequential Training.}~~Sequential Training enhances efficiency by optimizing one module at a time, promoting synergy between the retriever and generator. It includes Retriever-First approaches~\citep{asai2023self,sarto2022retrieval,schick2024toolformer,wang2023shall,zhu2024realm}, like RETRO~\citep{borgeaud2022improving}, which pre-trains a BERT-based retriever before training an encoder-decoder to seamlessly integrate retrieved content for improved performance. Alternatively, LLM-First methods~\citep{shao2023enhancing,shi2023replug,wang2023learning}, such as RA-DIT~\citep{lin2023ra}, first fine-tune the language model to effectively utilize retrieved knowledge, then refine the retriever for better alignment and coherence~\citep{asai2023self,shao2023enhancing}.

\noindent\textbf{Joint Training.}~~Joint Training synchronizes retriever and generator optimization in an end-to-end framework. RAG~\citep{lewis2020retrieval} minimizes negative log-likelihood to co-train both components, while REALM~\citep{guu2020retrieval} enhances retrieval precision with Maximum Inner Product Search (MIPS)~\citep{shen2015learning}. These methods adapt to task-specific needs, maximizing external knowledge benefits and minimizing generation errors.

\subsection{Model Merging} \label{Section 7.3}

Model merging has emerged as a vital post-training strategy for enhancing the performance and efficiency of LLMs across both training and inference phases~\citep{yang2024model,goddard2024arcee}. This approach consolidates specialized models into a unified architecture, circumventing the need for extensive retraining and addressing challenges posed by large model sizes and computational demands. Unlike training on amalgamated datasets, model merging integrates single-task models into a cohesive entity capable of multi-task proficiency, offering a resource-efficient paradigm for multi-task learning. By streamlining the training pipeline and fostering the development of versatile models with robust generalization across applications, this technique optimizes LLM deployment in diverse contexts. Given a set of candidate models \( M = \{M_1, M_2, \dots, M_n\} \), the objective is to devise a merging function \( F_{\text{merge}} \) that yields a unified model \( M' \), potentially anchored by a base model \( M_1 \), as depicted:

\begin{equation}
\label{eq:08-merge-overview}
\resizebox{0.5\textwidth}{!}{
    \includegraphics[width=\columnwidth]{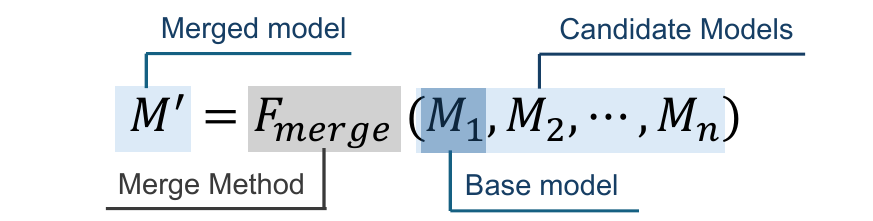}
}
\end{equation}

\subsubsection{Model Merging at Hierarchical Levels}
\begin{figure}[h]
    \centering
    \includegraphics[width=0.8\textwidth]{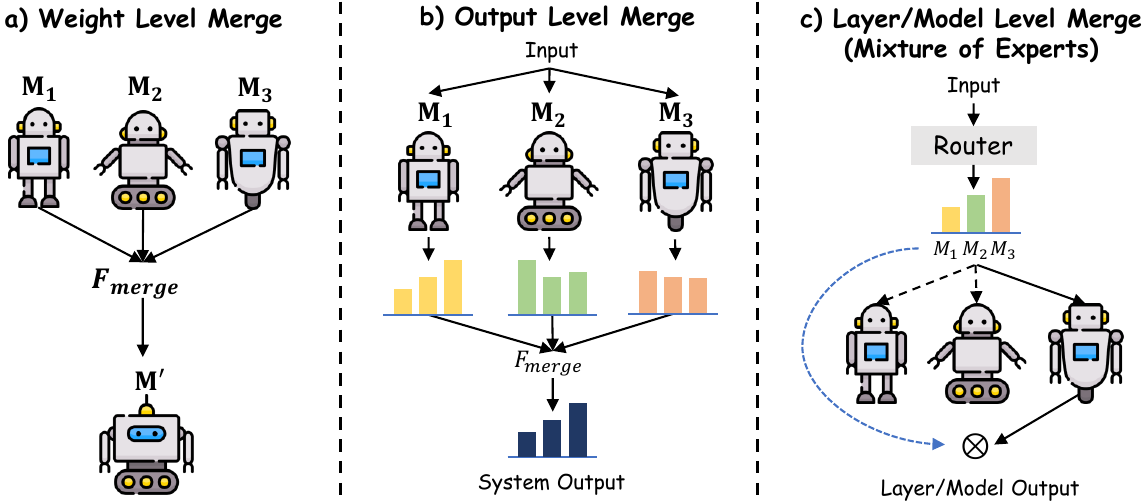}
    \caption{Taxonomy of model merging techniques, delineating hierarchical levels including weight-level, output-level, and model-level approaches for Large Language Models.}
\label{fig:08-merge-overview}
\end{figure}

Model merging techniques are systematically categorized into three hierarchical levels—weight-level, output-level, and model-level merging—as illustrated in \textbf{Fig.~\ref{fig:08-merge-overview}}.

\noindent\textbf{Weight-Level Model Merging.}~~Weight-level merging directly manipulates the parameter space, making it particularly effective for models sharing architectural similarities or trained on related tasks. Formally, given parameter sets \(\theta_1, \theta_2, \dots, \theta_n \in \mathbb{R}^d\), a linear merging scheme aggregates these into a unified set \(\theta'\) as:
\begin{equation}
    \theta' \;=\; \alpha_1\,\theta_1 \;+\; \alpha_2\,\theta_2 \;+\; \dots \;+\; \alpha_n\,\theta_n,
\quad
\text{subject to}
\quad
\alpha_k \,\ge\, 0,\,
\sum_{k=1}^n \alpha_k \;=\; 1.
\end{equation}

Model Soup~\citep{wortsman2022model,entezari2020all} exemplifies this by linearly combining weights from models fine-tuned on diverse tasks, yielding a single, efficient model. Task Arithmetic (TA)~\citep{ilharco2022editing} extends this flexibility through arithmetic operations on parameters, enhancing performance adaptability. To mitigate alignment issues, TIES-merging~\citep{yadav2024ties} ensures parameter congruence, while DARE~\citep{yu2024language} minimizes interference by probabilistically adjusting parameter deltas, optimizing the merging process for coherence and efficiency.

\noindent\textbf{Output-Level Model Merging.}~~Output-level merging becomes advantageous when models diverge in architecture or initialization, rendering weight-level methods impractical. This approach aggregates output distributions rather than internal parameters, formulated as:

\begin{equation}
y' 
\;=\; 
\alpha\,y_1 
\;+\; 
(1-\alpha)\,y_2,
\quad
\alpha \in [0,1],
\end{equation}
where \( y_1 \) and \( y_2 \) represent probability distributions from models \( M_1 \) and \( M_2 \), respectively.akin to ensemble strategies, this method synthesizes model predictions into a unified output. LLMBlender~\citep{jiang2023llm} implements this by generating independent outputs and fusing them via ranking and generative processes, while FuseLLM~\citep{wan2024knowledge} distills combined output probabilities into a single network for distributional fidelity. FuseChat~\citep{wan2024fusechat} bridges weight- and output-level merging by transferring knowledge from multiple LLMs into a consolidated target, enhancing cross-model synergy.

\noindent\textbf{Model-Level Model Merging.}~~Model-level merging integrates submodels or layers through routing mechanisms, often within mixture-of-experts (MoE) frameworks, expressed as:

\begin{equation}
M' 
\;=\; 
\mathrm{Merge}\bigl(M_1,\, M_2\bigr),
\end{equation}
where \(\mathrm{Merge}\) denotes either hard or soft routing functions. The Switch Transformer~\citep{fedus2022switch} employs discrete gating to activate expert layers selectively, reducing computational load albeit with potential performance trade-offs due to rigid routing. SoftMoE~\citep{puigcerver2023sparse} and SMEAR~\citep{muqeeth2023soft} utilize continuous gating to facilitate smoother transitions between experts, enhancing component integration and model cohesion.

\subsubsection{Pre-Merging Methods}
Pre-merging methods establish a compatibility foundation for model merging by optimizing the weight space, architectural coherence, and parameter alignment of independent models, thereby minimizing conflicts and interference in subsequent fusion stages. These techniques enhance the efficacy of merging processes, ensuring that the resulting unified model retains the strengths of its constituents while mitigating potential degradation.

\noindent\textbf{Linearization Fine-tuning.}~~This approach refines models within the tangent space of a pre-trained model, eschewing the original nonlinear parameter space to achieve weight disentanglement, which reduces interference during merging. Techniques such as partial linearization of adapters (e.g., TAFT~\citep{Liu2023TangentTF}) or attention layers~\citep{DBLP:journals/corr/abs-2407-07089} align weight updates to disjoint input regions, preserving independent functionalities in the merged model~\citep{OrtizJimnez2023TaskAI}. By constraining updates to a linearized framework, this method facilitates seamless integration across diverse models.

\noindent\textbf{Architecture Transformation.}~~This strategy converts heterogeneous models with divergent architectures into a homogeneous form amenable to direct parameter merging. Approaches include knowledge distillation, as exemplified by FuseChat~\citep{wan2024fusechat}, and identity layer insertion, such as CLAFusion~\citep{Nguyen2021OnCA}. GAN Cocktail~\citep{Avrahami2021GANCM} initializes target models to assimilate outputs from varied architectures, enabling a unified merging process that bridges structural disparities effectively.

\noindent\textbf{Weight Alignment.}~~This method aligns models to a shared weight basin through permutation, capitalizing on the Linear Mode Connectivity (LMC) property to enhance compatibility. Techniques encompass optimal transport (OTFusion~\citep{Singh2019ModelFV}), heuristic matching (Git re-basin~\citep{Ainsworth2022GitRM}), and learning-based alignment (Deep-Align~\citep{Shamsian2021PersonalizedFL}). REPAIR~\citep{Jordan2022REPAIRRP} mitigates alignment failures in models lacking normalization layers, ensuring robust parameter convergence prior to fusion.

\subsubsection{During-Merging Methods}
During-merging methods focus on dynamically optimizing parameter fusion strategies to resolve task conflicts, mitigate interference, and elevate the performance and generalization capacity of the resultant merged model. These approaches address the challenges of integrating disparate models in real-time, enhancing the adaptability and robustness of the unified architecture.

\noindent\textbf{Basic Merging.}~~This method leverages straightforward parameter averaging or task vector arithmetic, defining the task vector \(\tau_t\) as the deviation between fine-tuned parameters \(\Theta^{(t)}\) for the \(t\)-th task and the initial pre-trained parameters \(\Theta^{(0)}\):
\begin{equation}
\tau_t = \Theta^{(t)} - \Theta^{(0)},
\end{equation}
and facilitating multi-task learning through the formulation \(\Theta^{(\text{merge})} = \Theta^{(0)} + \lambda \sum_{t=1}^T \tau_t\)~\citep{ilharco2022editing}. While computationally efficient and conceptually elegant, this approach often encounters task interference arising from unmitigated parameter interactions, constraining its utility in scenarios demanding intricate task reconciliation.

\noindent\textbf{Weighted Merging.}~~This strategy dynamically allocates merging coefficients based on the significance of individual models, tailoring contributions to optimize fusion outcomes. MetaGPT~\citep{Zhou2024MetaGPTML} computes optimal weights by normalizing the squared L2 norm of each task vector:
\begin{equation}
\lambda_t^* = \frac{\|\tau_t\|^2}{\sum_{k=1}^T \|\tau_k\|^2},
\end{equation}  
thereby assigning greater influence to tasks with more substantial parameter shifts, as indicated by higher \(\|\tau_t\|^2\). SLERP~\citep{goddard2024arcee} employs spherical interpolation to ensure smooth parameter transitions, preserving model continuity, while Layer-wise AdaMerging~\citep{yang2024adamerging} refines this process by optimizing coefficients at a per-layer granularity, enhancing task-specific precision within the merged architecture.

\noindent\textbf{Subspace Merging.}~~This approach projects model parameters into sparse subspaces to minimize interference while upholding computational efficiency, addressing overlap in parameter contributions. TIES-Merging~\citep{yadav2024ties} retains the top 20\% of parameters by magnitude, resolving sign conflicts to maintain coherence, DARE~\citep{yu2024language} scales sparse weights to curtail redundancy, and Concrete~\citep{Tang2023ConcreteSL} utilizes bi-level optimization to craft adaptive masks, ensuring meticulous integration of model components with reduced interference across tasks.

\noindent\textbf{Routing-based Merging.}~~This technique dynamically fuses models based on input-specific attributes, enabling a context-responsive integration process. SMEAR~\citep{muqeeth2023soft} calculates sample-dependent expert weights to prioritize pertinent features, Weight-Ensembling MoE~\citep{DBLP:journals/corr/abs-2402-00433} employs input-driven routing of linear layers for selective activation, and Twin-Merging~\citep{Lu2024TwinMergingDI} melds task-shared and task-private knowledge, fostering a flexible merging framework that adapts to diverse input demands and enhances multi-task robustness.

\noindent\textbf{Post-calibration.}~~This technique corrects representation bias post-merging by aligning hidden representations of the unified model with those of its independent constituents, mitigating performance degradation. Representation Surgery~\citep{Yang2024RepresentationSF} exemplifies this by refining representational consistency, bolstering the merged model’s robustness and accuracy.

\section{Datasets}\label{Section 8}
Post-training techniques are meticulously engineered to refine the adaptability of LLMs to specialized domains or tasks, leveraging datasets as the cornerstone of this optimization process. A thorough examination of prior research~\citep{gunter2024apple, young2024yi} underscores that the quality, diversity, and relevance of data profoundly influence model efficacy, often determining the success of post-training endeavors. To elucidate the critical role of datasets in this context, we present a comprehensive review and in-depth analysis of those employed in post-training phases, categorizing them into three principal types based on their collection methodologies: human-labeled data, distilled data, and synthetic data. These categories reflect distinct strategies in data curation, with models adopting either a singular approach or a hybrid methodology integrating multiple types to balance scalability, cost, and performance. \textbf{Table~\ref{Dataset_categories}} provides a detailed overview of these dataset types, encompassing their origins, sizes, languages, tasks, and post-training phases (e.g., SFT and RLHF), which we explore in subsequent sections to highlight their contributions and challenges in advancing LLM capabilities.

\begin{table*}[h]
    \caption{Summary of Datasets Utilized in Post-training of Large Language Models (2021–2025). This table outlines key datasets, detailing their sizes, origins, release timelines, and attributes across three metrics: \textbf{Lang} (Language: EN for English, CN for Chinese, ML for Multilingual), \textbf{Task} (Type: MT for Multi-task, TS for Single-task), and \textbf{Phase} (Usage: SFT for Supervised Fine-Tuning, RLHF for Reinforcement Learning from Human Feedback). Datasets span from OpenAI Summarization to Magpie Reasoning V2, categorized by Human-Labeled, Distilled, and Synthetic types.}
    \label{Dataset_categories}
    \scriptsize
    \centering
    \renewcommand{\arraystretch}{1.2}
    \resizebox{1\textwidth}{!}{
    \begin{tabular}{lccccccr}
    \toprule
    \textbf{Datasets} & \textbf{Nums} &\textbf{Origin} &\textbf{Phase} & \textbf{Lang} & \textbf{Task} & \textbf{Categories} & \textbf{Time}\\
    \midrule 
    \rowcolor{Gray}OpenAI Summarization~\citep{hu2022sum} & 93K & Openai & RLHF & EN & MT & Human-Labeled & Jun-2021 \\
    \rowcolor{LightGray}Flan~\citep{wei2021finetuned} & 1.8M & Google & SFT & EN & MT & Human-Labeled & Sep-2021 \\
    \rowcolor{Gray}P3~\citep{sanh2021multitask} & 23M & Bigscience & RLHF & ML & MT & Human-Labeled & Oct-2021 \\
    \rowcolor{LightGray}SHP~\citep{ethayarajh2022understanding} & 349K & Stanfordnlp &RLHF & EN & MT & Human-Labeled & Oct-2021 \\
    \rowcolor{Gray}WebGPT~\citep{nakano2021webgpt} & 18K & OpenAI & RLHF & EN & TS & Human-Labeled & Dec-2021 \\
    \rowcolor{LightGray}Sup-Natinst~\citep{wang2022super} & 15K & Allenai &SFT & ML & MT & Human-Labeled & Apr-2022 \\
    \rowcolor{Gray}HH-RLHF~\citep{bai2022training} & 284K & Anthropic & RLHF & ML & TS & Human-Labeled & Apr-2022 \\
    \rowcolor{LightGray}Self-Instruct-52K~\citep{Wang2022SelfInstructAL} & 52K & UW &SFT & EN & MT & Synthetic Data & Dec-2022 \\
    \rowcolor{Gray}Unnatural Instructions~\citep{Honovich2022UnnaturalIT} & 52K & Orhonovich &SFT & EN & MT & Synthetic Data & Dec-2022 \\
    \rowcolor{LightGray}FlanV2~\citep{longpre2023flan} & - & Google & SFT & EN & MT & Human-Labeled & Oct-2022 \\
    \rowcolor{Gray}xP3~\citep{muennighoff2022crosslingual} & 78M & Bigscience & RLHF & ML & MT & Human-Labeled & Nov-2022 \\
    \rowcolor{LightGray}Alpaca~\citep{taori2023stanford} & 52K & Tatsu-lab &SFT & EN & MT & Synthetic Data & Mar-2023 \\
    \rowcolor{Gray}Vicuna~\citep{vicuna2023} & 53K & UC Berkeley &SFT & EN & MT & Synthetic Data & Mar-2023 \\
    \rowcolor{LightGray}OpenAssistant~\citep{kopf2024openassistant} & 84.4K, 161K & Laion.ai & SFT & ML & MT & Human-Labeled & Jan-2023 \\
    \rowcolor{Gray}HC3~\citep{guo-etal-2023-hc3} & 161K & Hello-SimpleAI & RLHF &  ML & TS & Distilled Data & Jan-2023\\
    \rowcolor{LightGray}Dolly-15K~\citep{conover2023free} & 15K & Databricks & SFT & EN & TS & Human-Labeled & May-2023 \\
    \rowcolor{Gray}ShareGPT~\citep{RyokoAI_ShareGPT} & 90K & RyokoAI &RLHF & ML & MT & Distilled Data & Apr-2023\\
    \rowcolor{LightGray}Alpaca-GPT4~\citep{peng2023instruction} & 52K & Microsoft &SFT & ML & MT & Synthetic Data & Apr-2023 \\
    \rowcolor{Gray}Evol-Instruct~\citep{luo2023wizardcoder} & 70K, 143K & WizardLM &SFT &  EN & MT & Synthetic Data & Apr-2023\\
    \rowcolor{LightGray}Belle~\citep{ji2023belle} & 0.5M, 1.1M & BelleGroup &SFT & CN & MT & Synthetic Data & Apr-2023 \\
    \rowcolor{Gray}Openorca~\citep{lian2023openorca} & 4.5M & Together & SFT & EN & MT & Human-Labeled & Jun-2023 \\
    \rowcolor{LightGray}StackExchange~\citep{h4stackexchange} & 4.5M & HuggingFace &RLHF & EN & MT & Human-Labeled & 2023 \\
    \rowcolor{Gray}OpenHermes-1~\citep{OpenHermes1} & 243K & Teknium & SFT & EN & MT & Synthetic Data & 2023 \\
    \rowcolor{LightGray}OpenHermes-2.5~\citep{OpenHermes2.5} & 1M & Teknium & SFT & EN & MT & Synthetic Data & 2023 \\
    \rowcolor{Gray}UltraChat~\citep{ding2023enhancing} & 200K, 28M & Thnlp & SFT & EN & MT & Synthetic Data & Oct-2023\\
    \rowcolor{LightGray}Instinwild~\citep{instructionwild}  & 52K & NUS &SFT & ML & MT & Synthetic Data & Oct-2023\\
    \rowcolor{Gray}\rowcolor{Gray}Baize~\citep{xu2023baize} & 653K & Project-Baize & SFT & EN & MT & Synthetic Data & Dec-2023 \\
    \rowcolor{LightGray}WildChat~\citep{zhao2024wildchat} & 1M & Allenai &SFT & EN & MT & Synthetic Data & May-2024\\
    \rowcolor{Gray}GenQA~\citep{chen2024genqa} & 513K & UMD & SFT & EN & MT & Synthetic Data & Jun-2024 \\
    \rowcolor{LightGray}Magpie Llama 3~\citep{xu2024magpie} & 300k, 1M & Allenai & SFT/DPO & EN & MT & Synthetic Data & Jun-2024 \\
    \rowcolor{Gray}Magpie Phi-3~\citep{xu2024magpie} & 300k, 1M & Allenai & SFT & EN & TS & Synthetic Data & Jun-2024 \\
    \rowcolor{LightGray}Magpie Qwen-2~\citep{xu2024magpie} & 200k, 300k, 1M & Allenai & SFT & EN & TS & Synthetic Data & Jul-2024 \\
    \rowcolor{Gray}Magpie Llama-3.1~\citep{xu2024magpie} & 300k, 500k, 1M & Allenai & SFT/DPO & EN & MT & Synthetic Data & Jul-2024 \\
    \rowcolor{LightGray}Magpie Gemma-2~\citep{xu2024magpie} & 200k, 534k & Allenai & SFT & EN & TS & Synthetic Data & Jul-2024 \\
    \rowcolor{Gray}Magpie Qwen-2.5~\citep{xu2024magpie} & 300k, 1M & Allenai & SFT & EN & TS & Synthetic Data & Oct-2024 \\
    \rowcolor{LightGray}Magpie Llama-3.3~\citep{xu2024magpie} & 500k, 1M & Allenai & SFT & EN & TS & Synthetic Data & Jan-2025 \\
    \rowcolor{Gray}Magpie Reasoning V2~\citep{xu2024magpie} & 150k, 250k & Allenai & SFT & EN & TS & Synthetic Data & Jan-2025 \\
    \bottomrule
    \end{tabular}
    }
\end{table*}

\subsection{Human-Labeled Datasets}
\label{Section 8.1}
Human-labeled datasets are distinguished by their exceptional accuracy and contextual fidelity, attributes derived from annotators' nuanced understanding of task intricacies and their ability to make precise, context-sensitive adjustments. These datasets serve as a cornerstone for refining instruction fine-tuning, significantly enhancing LLM performance across a diverse array of tasks by providing high-quality, expertly curated training signals. Within this category, prominent exemplars such as Flan~\citep{wei2021finetuned}, P3 (Public Pool of Prompts)~\citep{sanh2021multitask}, Sup-Natinst (Super-Natural Instructions)~\citep{wang2022super}, and Dolly-15K~\citep{conover2023free} stand out as widely adopted resources in LLM post-training, each contributing unique strengths to the optimization of model capabilities through human expertise.

\noindent \textbf{Human-Labeled Data for SFT.}~~In the SFT phase, human-labeled datasets play an indispensable role, as demonstrated by the contributions of Flan, Sup-Natinst, and Dolly-15K, which deliver meticulously crafted prompt-response pairs and task-specific instructions to elevate LLM efficacy across diverse NLP benchmarks.

$\bullet$~\textbf{Flan.}~~The Flan dataset~\citep{wei2021finetuned} constitutes a foundational resource, originally encompassing 62 widely recognized NLP benchmarks—such as HellaSwag~\citep{zellers2019hellaswag}, MRPC~\citep{dolan2005automatically}, and ANLI~\citep{nie2019adversarial}—to facilitate robust multi-task learning in English with its 1.8 million examples. Recently, FlanV2~\citep{longpre2023flan} has emerged as an advanced iteration, expanding its predecessor by integrating Flan~\citep{wei2021finetuned}, P3~\citep{sanh2021multitask}, Sup-Natinst~\citep{wang2022super}, and a plethora of additional datasets into a cohesive, comprehensive corpus, thereby amplifying its utility for SFT across diverse linguistic and task domains.

$\bullet$~\textbf{Sup-Natinst.}~~Super-Natural Instructions (Sup-Natinst)~\citep{wang2022super} offers an expansive and diverse array of 76 task types across 55 languages, establishing itself as a versatile resource for multilingual LLM post-training. Each task is meticulously paired with an instruction comprising a clear task definition—outlining the mapping from input text to desired output—and a set of examples that illustrate both correct and incorrect responses, providing a robust framework for guiding models toward precise task execution and enhancing cross-linguistic adaptability.

$\bullet$~\textbf{Dolly-15k.}~~Developed by Databricks employees, Dolly-15K~\citep{conover2023free} represents a curated corpus of 15,000 high-quality, human-generated prompt-response pairs, explicitly designed for instruction fine-tuning of LLMs. Encompassing a broad spectrum of topics and scenarios—including brainstorming, content generation, information extraction, open-ended question answering, and summarization—this dataset reflects a rich diversity of task types, enabling models to adapt flexibly to varied instructional contexts with enhanced contextual relevance.

The potency of human-labeled datasets in SFT stems from their extensive coverage of tasks and scenarios, a feature exemplified by the aforementioned corpora. Complementing these, OpenAssistant~\citep{kopf2024openassistant} delivers a substantial multilingual dialogue corpus derived from global crowdsourcing efforts, freely available to advance research pursuits, while OpenOrca~\citep{lian2023openorca} extends FlanV2~\citep{longpre2023flan} with millions of GPT-3.5 and GPT-4 completions, constituting a dynamic, expanding resource for fine-tuning and task alignment. However, despite their significant contributions to model generalization, the challenge of ensuring consistent annotation quality and diversity persists, necessitating rigorous quality control to maximize their impact.

\noindent\textbf{Human-Labeled Data for RLHF.}~~For RLHF, human-labeled datasets such as P3, its multilingual extension xP3~\citep{muennighoff2022crosslingual}, and SHP~\citep{ethayarajh2022understanding} provide essential human-annotated evaluations that refine LLM alignment with user preferences, offering a nuanced feedback mechanism for reward modeling. 

$\bullet$~\textbf{P3.}~~The P3 dataset~\citep{sanh2021multitask} is a meticulously curated instruction-tuning resource, aggregating 23 million multi-task prompts from the Hugging Face Hub, each accompanied by manually crafted instructions to span a diverse suite of NLP tasks, thereby providing a rich foundation for RLHF to enhance LLM adaptability and precision across varied applications.

$\bullet$~\textbf{xP3.}~~xP3 (Crosslingual Public Pool of Prompts)~\citep{muennighoff2022crosslingual} extends P3 into a multilingual framework, encompassing prompts and supervised data across 46 languages and 16 NLP tasks, designed to support multitask prompted fine-tuning for models like BLOOMZ and mT0. Its content integrates the English P3 dataset, four novel English tasks (e.g., translation, program synthesis), and 30 multilingual NLP datasets, offering a comprehensive resource for cross-linguistic RLHF optimization.

$\bullet$~\textbf{SHP.}~~SHP~\citep{ethayarajh2022understanding} comprises 349,000 human preference annotations for responses to questions and instructions across 18 subject areas, evaluating response helpfulness to train RLHF reward models and assess natural language generation (NLG) quality, distinguished by its exclusive reliance on human-authored data, setting it apart from hybrid datasets like HH-RLHF.

These datasets enhance RLHF by providing diverse human-annotated evaluations that refine model alignment with user preferences. OpenAI Summarization~\citep{hu2022sum} and Webgpt~\citep{nakano2021webgpt} offer structured, comparison-based feedback and Likert scale ratings, which help align model outputs more closely with human expectations. HH-RLHF~\citep{bai2022training} further strengthens this framework by including assessments of helpfulness and harmlessness, laying a solid foundation for models aimed at ensuring safety and ethical responses. Meanwhile, StackExchange~\citep{h4stackexchange} contributes domain-specific, user-generated content that enriches training data, particularly benefiting models that require expertise in technical fields. However, these datasets encounter challenges such as scalability, potential biases from human annotations, and limited applicability beyond their specific domains. Thus, while they are valuable, these resources may need to be supplemented with broader datasets to achieve comprehensive model alignment across diverse real-world tasks.

\subsection{Distilled Dataset}\label{Section 8.2}
Distilled data arises from a sophisticated process of refining expansive raw datasets into compact, optimized subsets that preserve critical information for LLM training, balancing performance retention with enhanced training efficiency and reduced computational demands. This methodology yields datasets that often rival or surpass their unrefined counterparts in efficacy, accelerating model convergence and minimizing resource consumption, particularly within the RLHF phase. Key examples, ShareGPT~\citep{RyokoAI_ShareGPT} and HC3 (Human-ChatGPT Comparison Corpus)~\citep{guo-etal-2023-hc3}, exemplify this approach, serving as widely adopted resources for fine-tuning LLMs by distilling real-world interactions and comparative insights into actionable training signals.

$\bullet$~\textbf{ShareGPT.}~~ShareGPT~\citep{RyokoAI_ShareGPT} functions as a dynamic data collection platform, aggregating approximately 90,000 conversations uploaded via its API from authentic user interactions with ChatGPT or GPT-4. Comprising genuine human instructions and queries paired with corresponding AI responses, this dataset distills naturalistic dialogue patterns into a concentrated resource, enabling RLHF to refine LLMs’ conversational fluency and contextual responsiveness with high relevance and quality.

$\bullet$~\textbf{HC3.}~~The HC3 dataset~\citep{guo-etal-2023-hc3} is purposefully engineered to juxtapose AI-generated responses from ChatGPT with human-authored answers, featuring 161,000 question-answer pairs across domains including open-ended topics, finance, medicine, law, and psychology. This distilled corpus facilitates comparative analysis of response characteristics and quality, empowering researchers to enhance LLMs’ output authenticity and domain-specific accuracy during RLHF, while highlighting distinctions between human and AI-generated content.  

\subsection{Synthetic Datasets}\label{Section 8.3}
Synthetic data constitutes a transformative asset in the SFT phase of LLM post-training, generated through AI models to deliver cost-effective, scalable, and privacy-preserving alternatives to human-labeled datasets. By automating the creation of instruction-response pairs and dialogues, synthetic data enables expansive training corpora that bolster model adaptability, with Self-Instruct-52K~\citep{Wang2022SelfInstructAL}, Vicuna~\citep{vicuna2023}, and Baize~\citep{xu2023baize} standing as principal examples widely utilized to enhance LLM instruction-following and dialogue generation capabilities.

\noindent\textbf{Datasets Based on the Self-Instruct Method.}~~Synthetic datasets employing the Self-Instruct method initiate with a modest set of manually crafted seed examples, leveraging LLMs to produce extensive instruction-following data that amplifies models’ responsiveness to diverse directives, exemplified by Self-Instruct-52K, Alpaca, and the Magpie series, which collectively advance instruction tuning through scalable automation.

$\bullet$~\textbf{Self-Instruct-52K.}~~Self-Instruct-52K~\citep{Wang2022SelfInstructAL} establishes a foundational benchmark for instruction-following models, generating 52,000 examples from manually crafted seeds using a variety of prompt templates to guide LLMs, thereby enhancing their ability to interpret and execute task-specific instructions with precision and consistency.

$\bullet$ \textbf{Alpaca.}~~Alpaca~\citep{taori2023stanford} and Alpaca-GPT4~\citep{peng2023instruction} expand an initial set of 175 seed pairs into 52,000 high-quality instruction-response pairs using GPT-3 and GPT-4, respectively, improving instruction-following proficiency, while InstInWild~\citep{instructionwild} adapts this approach for multilingual contexts, generating English and Chinese datasets to bolster cross-linguistic adaptability.

$\bullet$ \textbf{Magpie Datasets.}~~The Magpie datasets~\citep{xu2024magpie} harness aligned LLMs to generate instruction-response pairs from predefined templates, yielding specialized families like Magpie Reasoning V2 (emphasizing chain-of-thought reasoning), Magpie Llama-3 and Qwen-2 Series (tailored to popular models), Magpie Gemma-2 (for the Gemma architecture), and variants like Magpie-Air-DPO incorporating preference optimization signals, collectively enhancing SFT and instruction tuning across conversational and reasoning tasks.

Beyond these, datasets like Unnatural Instructions~\citep{Honovich2022UnnaturalIT} (240K examples), Evol-Instruct~\citep{luo2023wizardcoder} (70K-143K refined entries via iterative complexity enhancement), and Belle~\citep{ji2023belle} (0.5M-1.1M Chinese dialogues from ChatGPT) significantly scale instruction generation, though challenges in quality assurance, complexity calibration, and bias mitigation persist, necessitating ongoing refinement to ensure reliability in intricate applications.

\noindent\textbf{Datasets Based on Self-Chat Methods.}~~Self-Chat datasets employ a technique where models simulate multi-turn conversations internally or with peers, enhancing dialogue generation capabilities and addressing deficiencies in existing corpora, with Baize, UltraChat, and OpenHermes exemplifying this approach through automated interaction strategies.

$\bullet$~\textbf{Baize.}~~Baize~\citep{xu2023baize} utilizes ChatGPT’s Self-Chat technique to produce 653,000 multi-turn dialogues, integrating seed data from Quora, Stack Overflow, and Alpaca to enrich instruction-following quality, thereby refining LLMs’ conversational coherence and task adherence for SFT.

$\bullet$~\textbf{UltraChat.}~~UltraChat~\citep{ding2023enhancing} employs multiple ChatGPT APIs to generate over 12 million high-quality conversation records across diverse topics, overcoming prevalent issues in multi-turn datasets like poor quality and inaccurate annotations, providing a robust SFT resource for dialogue enhancement.

$\bullet$~\textbf{Openhermes.}~~OpenHermes, developed by Teknium, includes OpenHermes-1~\citep{OpenHermes1} (243K entries) and its expanded successor OpenHermes-2.5~\citep{OpenHermes2.5} (1M entries), offering high-quality SFT datasets with increased volume and diversity, spanning a wide array of topics and task types to bolster dialogue and instruction-following proficiency.

These Self-Chat datasets enable models to craft multi-turn dialogues through self-interaction, as seen in Baize’s use of ChatGPT with diverse seeds and UltraChat’s extensive API-driven conversations, significantly improving dialogue quality and filling critical gaps in training data availability.

\noindent\textbf{Datasets Based on Real User Interactions.}~~Datasets derived from real user interactions harness authentic conversational exchanges with LLMs, capturing diverse and genuine inputs to enhance models’ capacity to address real-world scenarios, with Vicuna, WildChat, and GenQA serving as key examples of this approach.

$\bullet$~\textbf{Vicuna.}~~Vicuna~\citep{vicuna2023} is fine-tuned on approximately 70,000 user-shared conversations from ShareGPT’s public API, processed by converting HTML to markdown, filtering low-quality samples, and segmenting lengthy dialogues to fit model context lengths, ensuring high-quality SFT data for realistic interaction modeling.

$\bullet$~\textbf{WildChat.}~~WildChat~\citep{zhao2024wildchat} comprises 1 million real-world user-ChatGPT interactions across multiple languages and prompt types, featuring unique exchanges like ambiguous requests and code-switching, serving dual purposes as an SFT resource and a tool for analyzing user behavior.

$\bullet$~\textbf{GenQA.}~~GenQA~\citep{chen2024genqa} offers a vast SFT dataset of over 10 million cleaned and filtered instruction samples, generated entirely by LLMs without human input or complex pipelines, complementing existing corpora by rapidly producing synthetic data to address coverage gaps.

Synthetic data’s advantages in cost, scalability, and privacy are tempered by potential deficits in depth and authenticity compared to human-labeled counterparts, risking bias propagation and oversimplification. Reliance on AI-generated content may perpetuate model-inherent errors, underscoring the necessity of integrating synthetic and human-generated data to bolster LLM robustness and applicability across diverse contexts.

\section{Applications}\label{Section 9}
Despite the robust foundational capabilities imparted by pre-training, Large Language Models (LLMs) frequently encounter persistent limitations when deployed in specialized domains, including constrained context lengths, tendencies toward hallucination, suboptimal reasoning proficiency, and ingrained biases. These shortcomings assume critical significance in real-world applications, where precision, reliability, and ethical alignment are paramount. Such challenges prompt fundamental inquiries: (1) How can LLM performance be systematically enhanced to meet domain-specific demands? (2) What strategies can effectively mitigate the practical obstacles inherent in applied settings? Post-training emerges as a pivotal solution, augmenting LLMs’ adaptability by refining their recognition of domain-specific terminology and reasoning patterns while preserving their broad-spectrum competencies. This chapter delineates the transformative applications of post-trained LLMs across professional, technical, and interactive domains, elucidating how tailored post-training methodologies address these challenges and elevate model utility in diverse contexts. 


\subsection{Professional Domains}

\noindent\textbf{Legal Assistant.}~~The legal domain exemplifies a compelling arena for leveraging post-training to imbue LLMs with specialized expertise, enabling them to navigate the intricate landscape of legal knowledge and address multifaceted challenges inherent in jurisprudence. A burgeoning body of research~\citep{savelka2023explaininglegalconceptsaugmented,westermann2023llmediatorgpt4assistedonline,liu2024catastrophicforgettingintegratinggeneral} has investigated LLM applications in this field, spanning legal question answering~\citep{mansouri2023falqufindinganswerslegal,louis2023interpretablelongformlegalquestion}, judgment prediction~\citep{chalkidis-etal-2021-paragraph,trautmann2022legalpromptengineeringmultilingual}, document summarization~\citep{zongyue2023leeclegalelementextraction,jiang-etal-2024-leveraging}, and broader tasks like retrieval enhancement and judicial reasoning~\citep{nguyen2024enhancinglegaldocumentretrieval,hussain2024largelanguagemodelsjudicial,deroy2024applicabilitylargelanguagemodels}. Post-trained legal assistants, such as those exemplified by LawGPT~\citep{lawgpt} and Lawyer-LLaMA~\citep{Lawyer-LLama}, have demonstrated remarkable proficiency, not only offering dependable guidance across diverse legal matters but also achieving success in professional qualification exams, a testament to their advanced interpretive and analytical capabilities. Multilingual support, as seen in models like LexiLaw~\citep{LexiLaw} and SAUL~\citep{colombo2024saullm54bsaullm141bscaling}, extends this utility to languages including English and Chinese, broadening accessibility. Central to these advancements is post-training on curated legal corpora, such as ChatLaw~\citep{ChatLaw}, which integrates extensive legal texts into conversational datasets, enabling models to refine their reasoning and terminology recognition.

\noindent\textbf{Healthcare and Medical.}~~Post-training substantially elevates LLM performance across a spectrum of healthcare and medical applications, harnessing domain-specific data to address clinical and academic needs with precision. In clinical settings, LLMs facilitate tasks such as drug discovery~\cite{MolT5}, drug synergy prediction~\cite{SynerGPT2024} and catalyst design~\cite{catalyst2023}, diagnostic support, medical record generation, and patient interaction, while in academia, they excel in medical report synthesis~\citep{hou2023organo} and question answering~\citep{kim2024medex}, driven by performance gains from tailored post-training. For instance, ChatMed~\citep{zhu2023ChatMed}, honed on 500,000 medical consultation records, exemplifies enhanced diagnostic and consultative accuracy, while PULSE~\citep{pulse2023}, fine-tuned with 4 million instructions spanning Chinese medical and general domains, showcases superior multi-task proficiency. These models outperform their general-purpose counterparts by leveraging post-trained adaptations that embed nuanced medical knowledge, highlighting the indispensability of customized datasets in achieving practical utility. Such advancements not only improve task-specific outcomes but also pave the way for integrating LLMs into healthcare workflows, where precision and contextual relevance are non-negotiable, underscoring post-training’s transformative impact on real-world medical applications.

\noindent\textbf{Finance and Economics.}~~In the domains of finance and economics, LLMs exhibit considerable potential for tasks including sentiment analysis~\citep{luo2024pretrained}, information extraction~\citep{Kaur_2023}, and question answering~\citep{islam2023finance}, with post-training amplifying their efficacy through domain-specific refinements. While general-purpose LLMs provide a solid foundation, specialized models like FinGPT~\citep{yang2023fingpt} and DISC-FinLLM~\citep{chen2023disc} demonstrate marked improvements when post-trained on financial corpora, excelling in tasks requiring nuanced understanding of market dynamics and terminology. Similarly, XuanYuan~\citep{zhang2023xuanyuan} employs extensive financial datasets and advanced post-training techniques to enhance accuracy in economic modeling and prediction, outperforming untuned benchmarks. These developments illustrate post-training’s critical role in adapting LLMs to the intricate demands of financial applications, where precision in interpreting quantitative data and qualitative insights is paramount, ensuring models deliver reliable, domain-informed outputs that align with industry standards and expectations.

\noindent\textbf{Mobile Agents.}~~The evolution of large multi-modal models (LMMs) has catalyzed a burgeoning domain of agentic research focused on LMM-based graphical user interface (GUI) agents~\citep{wang2024guisurvey}. This field aims to develop AI assistants capable of executing tasks across diverse GUI environments, encompassing web interfaces~\citep{deng2023mindweb,zheng2024seeact,he2024webvoyager,yoran2024assistantbenchwebagentssolve,reddy2024infogent}, personal computing platforms~\citep{hong2023cogagent,ufo,liu2024visualagentbench,xie2024osworld,cradle}, and mobile devices~\citep{wang2024mobile,yang2023appagent,li2024appagentv2,wang2024mobile2,liu2024autoglm}. Within the mobile context, one research trajectory enhances the perceptual and reasoning capacities of individual agents through tool integration~\citep{wang2024mobile} and an additional exploration phase~\citep{yang2023appagent,li2024appagentv2}. Recent advancements exhibit considerable potential by employing multi-agent systems for decision-making and reflection~\citep{rawles2024androidworld,wang2024mobile2}, thereby improving task efficacy. Notably, MobileAgent-E~\citep{Mobile-Agent-E2025} introduces a hierarchical structure among agents, facilitating robust long-horizon planning and elevating the precision of low-level actions. These developments underscore the transformative role of multi-modal post-training strategies in fostering adaptive, efficient agents for complex mobile environments.

\subsection{Technical and Logical Reasoning}

\noindent\textbf{Mathematical Reasoning.}~~LLMs demonstrate significant promise in mathematical reasoning, spanning algebraic manipulations, calculus, and statistical analysis, with post-training pivotal in bridging the gap between computational and human-like proficiency. GPT-4~\citep{achiam2023gpt} achieves high scores on standardized math assessments, a feat attributed to its diverse pre-training corpus, yet post-training further refines this capability. DeepSeekMath~\citep{shao2024deepseekmath}, for instance, leverages specialized mathematical datasets and techniques like Supervised Fine-Tuning (SFT) and Group Relative Policy Optimization (GRPO)~\citep{shao2024deepseekmath} to enhance its reasoning precision, tackling complex problems with structured chains of thought (CoT). OpenAI’s o1~\citep{jaech2024openai} advances this frontier through reinforcement learning (RL), iteratively optimizing reasoning strategies to achieve superior performance in multi-step derivations and proofs. This continuous refinement via post-training not only elevates accuracy but also aligns LLM outputs with rigorous mathematical logic, positioning them as valuable tools in educational and research contexts where advanced reasoning is essential.

\noindent\textbf{Code Generation.}~~Post-training has revolutionized code generation, empowering LLMs to excel in automated coding, debugging, and documentation, thereby transforming software development workflows. Codex~\citep{openaicodex}, trained on a vast, diverse codebase, underpins GitHub Copilot~\footnote{https://github.com/features/copilot}, delivering real-time coding assistance with remarkable accuracy. Specialized models like Code Llama~\citep{codellama} further refine this capability, leveraging post-training on programming-specific datasets to assist developers across languages and frameworks. OpenAI’s o1~\citep{jaech2024openai} extends its mathematical reasoning prowess to code generation, producing high-quality, context-aware code snippets that rival human outputs. Current research focuses on enhancing personalization, deepening contextual understanding, and embedding ethical safeguards to mitigate risks like code misuse, ensuring LLMs maximize productivity while adhering to responsible development principles in technical domains.

\subsection{Understanding and Interaction}

\noindent\textbf{Recommendation System.}~~LLMs have emerged as transformative agents in recommendation systems, analyzing user interactions, product descriptions, and reviews to deliver personalized suggestions with unprecedented granularity~\citep{Zhang_2023,friedman2023recllm,wei2024llmrec}. Post-training enhances their capacity to integrate sentiment analysis, enabling nuanced comprehension of content and emotional undertones, as evidenced in models like GPT-4~\citep{achiam2023gpt} and specialized systems like LLaRA~\citep{liao2024llara} and AgentRec~\citep{zhang2024agentrec}. E-commerce giants such as Amazon and Taobao harness these capabilities to process review sentiments, search queries, and purchase histories, refining customer preference models and predicting interests with high fidelity~\citep{friedman2023recllm}. Beyond ranking items, post-trained LLMs engage in conversational recommendation, planning, and content generation, elevating user experience by offering dynamic, context-sensitive interactions that adapt to evolving preferences, a testament to post-training’s role in bridging data analysis with practical utility.

\noindent\textbf{Speech Conversation.}~~Post-trained LLMs have redefined speech processing, advancing recognition, synthesis, and translation to unprecedented levels of naturalness and accuracy~\citep{wang2017tacotros}. These models tackle tasks like text-to-speech~\citep{li2023styletts}, text-to-audio generation~\citep{huang2023makeanaudio}, and speech recognition~\citep{warden2018speech}, powering ubiquitous tools such as Amazon’s Alexa, Apple’s Siri, and Alibaba’s Tmall Genie. Whisper~\citep{whisper} exemplifies this progress with its high-fidelity transcription, while GPT-4o~\citep{openai2024gpt4o} introduces real-time voice interaction, merging multimodal inputs seamlessly. Future trajectories include multilingual translation and personalized voice synthesis, where post-training refines LLMs to break language barriers and tailor responses to individual user profiles, enhancing accessibility and engagement in human-computer interactions across global contexts.

\noindent\textbf{Video Understanding.}~~The extension of LLMs into video understanding marks a significant frontier, with post-training enabling models like Video-LLaMA~\citep{zhang2023video} to perform captioning, summarization, and content analysis, streamlining multimedia creation and comprehension. Sora~\citep{sora} further revolutionizes this domain by generating complex videos from textual prompts, democratizing content production by reducing technical barriers and fostering innovative storytelling. These advancements leverage post-training to adapt LLMs to visual-temporal data, enhancing their interpretative depth and utility in applications ranging from education to entertainment. However, they introduce challenges in computational scalability, privacy protection, and ethical governance, particularly concerning generated content misuse. As post-training methodologies evolve, addressing these issues will be imperative to ensure sustainable, responsible deployment in video-related applications, balancing innovation with societal considerations.

\section{Open Problems and Future Directions}\label{Section 10}

In this section, we critically evaluate the unresolved challenges and prospective trajectories in post-training methodologies for Large Language Models (LLMs), situating our analysis within the transformative advancements heralded by the releases of OpenAI’s o1~\citep{jaech2024openai} and DeepSeek-R1~\citep{DeepSeekAI2025DeepSeekR1IR}. These models, leveraging large-scale reinforcement learning (RL), have redefined reasoning benchmarks, yet their emergence amplifies the urgency of addressing persistent limitations in post-training techniques. The following subsections delineate senven pivotal open problems, each underscored by its critical importance to the field’s progression and the pressing need for resolution, alongside feasible strategies to propel future research and ensure the responsible evolution of LLMs across diverse applications.

\noindent\textbf{Reasoning Enhancement Beyond Large-Scale RL.}~~The introduction of o1 and DeepSeek-R1 has marked a paradigm shift in LLM reasoning capabilities, harnessing extensive RL frameworks like RLHF and Group Relative Policy Optimization (GRPO) to achieve unprecedented accuracy in multi-step problem-solving, such as mathematical proofs and logical derivations. However, the reliance on binary reward signals and extensive human feedback exposes a critical limitation: their inability to generalize effectively across complex, open-ended tasks, such as scientific hypothesis generation or strategic decision-making in dynamic environments. This gap is urgent, as the demand for LLMs to emulate human-like reasoning in real-world contexts grows, and its importance lies in unlocking their potential as autonomous intellectual agents beyond current benchmarks. Current RL approaches struggle with reward sparsity and lack adaptability to task complexity, necessitating innovative frameworks. Feasible solutions include developing multi-objective RL systems that integrate self-supervised consistency checks (e.g., verifying logical coherence across reasoning steps) and domain-specific priors, such as mathematical axioms or scientific principles, to guide inference without exhaustive human annotations~\citep{holliday2024conditional,cheng2025self}. Such advancements could reduce dependency on costly feedback loops, enhance scalability, and enable LLMs to tackle uncharted reasoning domains, a prospect made tangible by DeepSeek-R1’s cold-start RL innovations.

\noindent\textbf{Scalability of Post-Training for Next-Generation LLMs.}~~As LLMs escalate in size and complexity, exemplified by the parameter-heavy architectures of next-generation models, the scalability of post-training emerges as a formidable and pressing challenge. The resource-intensive nature of RL-based methods, such as DeepSeek-R1’s cold-start approach requiring extensive computational infrastructure, restricts accessibility to well-funded entities and raises significant sustainability concerns, particularly for multi-modal applications (e.g., video analysis) and real-time systems (e.g., conversational agents). This issue is critically important, as it threatens to widen the gap between resource-rich and resource-constrained research communities, impeding equitable progress in LLM development. While Parameter-Efficient Fine-Tuning (PEFT)~\citep{Hu2021LoRALA} mitigates some overhead, its performance often degrades on large-scale datasets, underscoring the need for scalable alternatives. Viable future directions~\citep{xu2024scaling,ibrahim2024simple,jiang2024megascale} include the design of lightweight RL algorithms—potentially adapting GRPO for reduced memory footprints—federated post-training frameworks that distribute computational loads across decentralized networks, and advanced distillation techniques that preserve reasoning and adaptability while minimizing resource demands. These solutions, if realized, could democratize post-training, aligning with the field’s urgent need for sustainable and inclusive innovation.

\noindent\textbf{Ethical Alignment and Bias Mitigation in RL-Driven Models.}~~Post-training via RL, as demonstrated in o1’s cautious alignment strategies, amplifies ethical risks by potentially reinforcing biases embedded in training datasets like HH-RLHF~\citep{bai2022training} or synthetic corpora, a challenge of paramount urgency given LLMs’ deployment in sensitive domains such as healthcare diagnostics and judicial decision-making. The dynamic variability of ethical alignment—where fairness in one cultural context may constitute bias in another—poses a significant barrier to achieving universally trustworthy LLMs, making this issue critically important for ensuring equitable and safe AI systems. Current methods risk over-censorship, compromising utility (e.g., stifling creative outputs), or under-correction, perpetuating harmful biases (e.g., racial or gender disparities). Addressing this demands the development of fairness-aware RL objectives, incorporating multi-stakeholder preference models (e.g., aggregating diverse human judgments) and adversarial debiasing techniques to neutralize dataset biases during training. The feasibility of these approaches~\citep{lindstrom2024ai} is bolstered by recent advances in interpretability tools and multi-objective optimization, enabling a balanced trade-off between ethical robustness and practical functionality, a necessity underscored by o1’s real-world deployment challenges.

\noindent\textbf{Seamless Multi-Modal Integration for Holistic Reasoning.}~~The trajectory toward multi-modal LLMs, foreshadowed by o1’s reasoning enhancements and GPT-4o’s synthesis capabilities~\citep{openai2024gpt4o}, accentuates an urgent need for post-training methods that seamlessly integrate text, images, audio, and other data types to enable holistic reasoning—a capability critical for applications like real-time video analysis, augmented reality, and cross-modal scientific inquiry. Current approaches falter in achieving robust cross-modal alignment due to data heterogeneity and the scarcity of comprehensive multi-modal training corpora, limiting LLMs’ ability to reason across diverse inputs cohesively. This challenge’s importance lies in its potential to unlock transformative applications, yet its resolution remains elusive without scalable frameworks. DeepSeek-R1’s cold-start RL offers a promising starting point, suggesting that unified modal encoders (e.g., capable of encoding heterogeneous data into a shared latent space) and dynamic RL policies that adaptively weight modal contributions could bridge this gap. Future research should prioritize the creation of multi-modal benchmarks and synthetic datasets, building on efforts like Magpie~\citep{xu2024magpie}, to drive progress, a feasible endeavor given recent strides in multi-modal pre-training and RL optimization.

\noindent\textbf{Context-Adaptive Trustworthiness Frameworks.}~~Trustworthiness in post-trained LLMs is increasingly recognized as a dynamic, context-dependent attribute rather than a static quality, as evidenced by o1’s cautious outputs in sensitive domains like education versus its freer responses in creative tasks. This variability—where safety imperatives (e.g., avoiding misinformation in educational settings) may conflict with utility demands (e.g., fostering creativity in writing)—poses an urgent challenge, given its critical importance to user trust and LLM applicability across diverse real-world scenarios. Current post-training methods often over-prioritize safety, yielding utility trade-offs that diminish practical value, or fail to adapt to context-specific needs, undermining reliability. Resolving this requires context-sensitive RL models that dynamically adjust safety-utility trade-offs, leveraging real-time user feedback and interpretable safety metrics (e.g., transparency scores for generated outputs) to ensure adaptability. The feasibility of this approach~\citep{sun2024trustllm} is supported by advances in adaptive learning systems and real-time monitoring, offering a pathway to balance trustworthiness with functionality, a pressing need as LLMs like o1 expand into high-stakes applications.

\noindent\textbf{Accessibility and Democratization of Post-Training Innovations.}~~The computational intensity of advanced post-training methods, epitomized by DeepSeek-R1’s RL-driven approach, confines their application to resource-rich entities, presenting an urgent barrier to accessibility that stifles innovation within smaller research communities and industry sectors (i.e., an issue of paramount importance for fostering equitable progress in AI). This exclusivity not only limits the diversity of contributions but also hampers the field’s ability to address global challenges collaboratively. Democratizing these innovations demands the development of efficient, open-source tools and frameworks that lower entry barriers without compromising quality, a goal rendered feasible through scalable PEFT adaptations for RL~\citep{Hu2021LoRALA}, collaborative platforms for sharing post-trained models (e.g., Hugging Face hubs), and streamlined synthetic data generation pipelines akin to Magpie~\citep{xu2024magpie}. Future efforts should focus on optimizing these solutions to enable widespread adoption, ensuring that the transformative potential of post-training—exemplified by o1 and DeepSeek-R1—extends beyond elite institutions to enrich the broader AI ecosystem.

\noindent\textbf{Creative Intelligence \& System 2 Thinking.}~~The integration of creative intelligence into System 2 reasoning represents an emergent frontier in the evolution of LLMs, as highlighted in \citep{li2025system}. While reasoning LLMs like OpenAI’s o1 and DeepSeek’s R1 excel in deliberate, step-by-step logical analysis—mimicking System 2 thinking—their capacity for creative intelligence, which involves generating novel ideas, synthesizing disparate concepts, and adapting flexibly to unstructured problems, remains underexplored. This gap is critical, as creative intelligence underpins human-like problem-solving in domains such as artistic creation, scientific discovery, and strategic innovation, where rigid logical frameworks alone are insufficient. The urgency of this challenge lies in its potential to elevate LLMs from analytical tools to autonomous creative agents, a transformative leap toward Artificial General Intelligence (AGI). Below, we outline this open problem and propose future directions, drawing on insights from the survey.

\newpage
\section{Conclusion}\label{Section 11}

This paper offers the first exhaustive survey of Post-training Language Models (PoLMs), systematically tracing their trajectory from ChatGPT’s alignment origins in 2018 to DeepSeek-R1’s reasoning milestone in 2025, and affirming their transformative influence on reasoning precision, domain adaptability, and ethical integrity. We have evaluated a broad spectrum of techniques (i.e., Fine-tuning, Alignment, Reasoning, Efficiency, and Integration and Adaptation), synthesizing their contributions across professional, technical, and interactive domains, from legal analysis to multi-modal comprehension. Our analysis underscores that PoLMs have markedly advanced LLM capabilities, evolving from initial alignment innovations to sophisticated reasoning frameworks; nonetheless, it reveals persistent challenges, including bias persistence, computational scalability, and context-variable ethical alignment. These findings, encapsulated within a novel taxonomy, emphasize the necessity of an integrative approach that aligns reasoning advancements with efficiency and ethical imperatives. We conclude that sustained interdisciplinary collaboration, rigorous methodological assessment, and the development of adaptive, scalable frameworks are critical to realizing LLMs’ potential as reliable, responsible tools across diverse applications. As the pioneering survey of its kind, this work consolidates PoLMs progress over recent years and lays a robust intellectual foundation, inspiring future research to cultivate LLMs that adeptly integrate precision, ethical robustness, and versatility to meet the evolving demands of scientific and societal contexts.


\bibliographystyle{unsrt}
\bibliography{main}


\end{document}